\documentclass{ieeetj}
\usepackage{indentfirst}

\usepackage{cite}
\usepackage{amsmath,amssymb,amsfonts}
\usepackage{algorithmic}
\usepackage{graphicx,color}
\usepackage{textcomp}
\usepackage{xcolor}
\usepackage{hyperref}
\hypersetup{hidelinks=true}
\usepackage{algorithm,algorithmic}
\usepackage{bm}
\usepackage{array}
\usepackage[caption=false,font=normalsize,labelfont=sf,textfont=sf]{subfig}
\usepackage{stfloats}
\usepackage{url}
\usepackage{verbatim}
\usepackage{tabularx}
\usepackage{booktabs}
\usepackage{subcaption}
\usepackage{multirow}
\usepackage{threeparttable}

\usepackage[export]{adjustbox}

\def\BibTeX{{\rm B\kern-.05em{\sc i\kern-.025em b}\kern-.08em
    T\kern-.1667em\lower.7ex\hbox{E}\kern-.125emX}}
\AtBeginDocument{\definecolor{tmlcncolor}{cmyk}{0.93,0.59,0.15,0.02}\definecolor{NavyBlue}{RGB}{0,86,125}}

\def\OJlogo{\vspace{-4pt}$<$Society logo(s) and publication title will appear here.$>$}
\def\seclogo{\vspace{10pt}$<$Society logo(s) and publication title will appear here.$>$}

\def\authorrefmark#1{\ensuremath{^{\textbf{#1}}}}

\begin{document}
\receiveddate{XX Month, XXXX}
\reviseddate{XX Month, XXXX}
\accepteddate{XX Month, XXXX}
\publisheddate{XX Month, XXXX}
\currentdate{XX Month, XXXX}
\doiinfo{XXXX.2022.1234567}

\markboth{Terrain Perception for Agricultural UAVs in Complex Farmland via Rotating mmWave Radar}{Zhan {et al.}}

\title{
Terrain Perception for Agricultural UAVs in Complex Farmland via Rotating mmWave Radar
}


\author{
Zhihao Zhan\authorrefmark{1,2}, Le Tao\authorrefmark{2}, Shaobin Li\authorrefmark{2}, Chenxin Fang\authorrefmark{1}, Xingrui Yang\authorrefmark{3}, \\ Liang Li\authorrefmark{4}, Rui Fan\authorrefmark{5}, and Yuhang Ming\authorrefmark{1}
}
\affil{School of Computer Science, Hangzhou Dianzi University, Hangzhou, 310018, China}
\affil{TopXGun Robotics, Nanjing, 211100, China}
\affil{CARDC, Mianyang, 621000, China}
\affil{College of Control Science and Engineering, Zhejiang University, Hangzhou, 310027, China}
\affil{College of Electronic and Information Engineering, Shanghai Institute of Intelligent Science and Technology, \\Shanghai Research Institute for Intelligent Autonomous Systems, Shanghai Key Laboratory of Intelligent Autonomous Systems, \\and State Key Laboratory of Autonomous Intelligent Unmanned Systems, Tongji University, Shanghai, 201804, China}
\corresp{Corresponding author: Yuhang Ming (email: yuhang.ming@hdu.edu.cn).}
\authornote{This work was supported in part by Zhejiang Provincial Natural Science Foundation of China under Grant LQN25F030015, in part by the National Natural Science Foundation of China under Grant 62401188 and Grant 62473288, in part by the Fundamental Research Funds for the Central Universities, and in part by the Xiaomi Young Talents Program.
}

\begin{abstract}
Accurate terrain perception is essential for terrain-following flight of agricultural uncrewed aerial vehicles (UAVs), yet remains challenging in real-world farmland due to occlusions, complex terrain geometry, and environmental disturbances. 
Millimeter-wave (mmWave) radar is a promising sensing modality for this task due to its robustness to adverse conditions; however, existing UAV-mounted radar systems rely on fixed field of view (FoV) and terrain extraction methods designed for dense light detection and ranging (LiDAR) data, leading to incomplete and unreliable terrain estimation.
To address these limitations, we present a low-cost rotating mmWave radar–enabled terrain perception framework for agricultural UAVs operating in complex farmland environments. 
Specifically, a mechanically rotating sensing design is introduced to enlarge spatial coverage and improve terrain observability beyond the limitations of fixed-view radar under dynamic low-altitude flight. 
Building upon this sensing capability, we further design a pose-consistent terrain reconstruction pipeline tailored for sparse, noisy, and partially observable radar data, enabling reliable ground extraction and continuous terrain surface estimation in challenging agricultural scenarios.
The complete system is deployed on a real agricultural UAV platform and evaluated through extensive field experiments. 
Experimental results demonstrate improved terrain coverage and estimation accuracy, achieving an F1 score of \textit{94.42\%} for ground segmentation, while the closest rival only achieves \textit{90.48\%}. 
\end{abstract}



\maketitle

\section{INTRODUCTION}

\begin{figure}[t]
    \centering
    \includegraphics[width=0.49\textwidth]{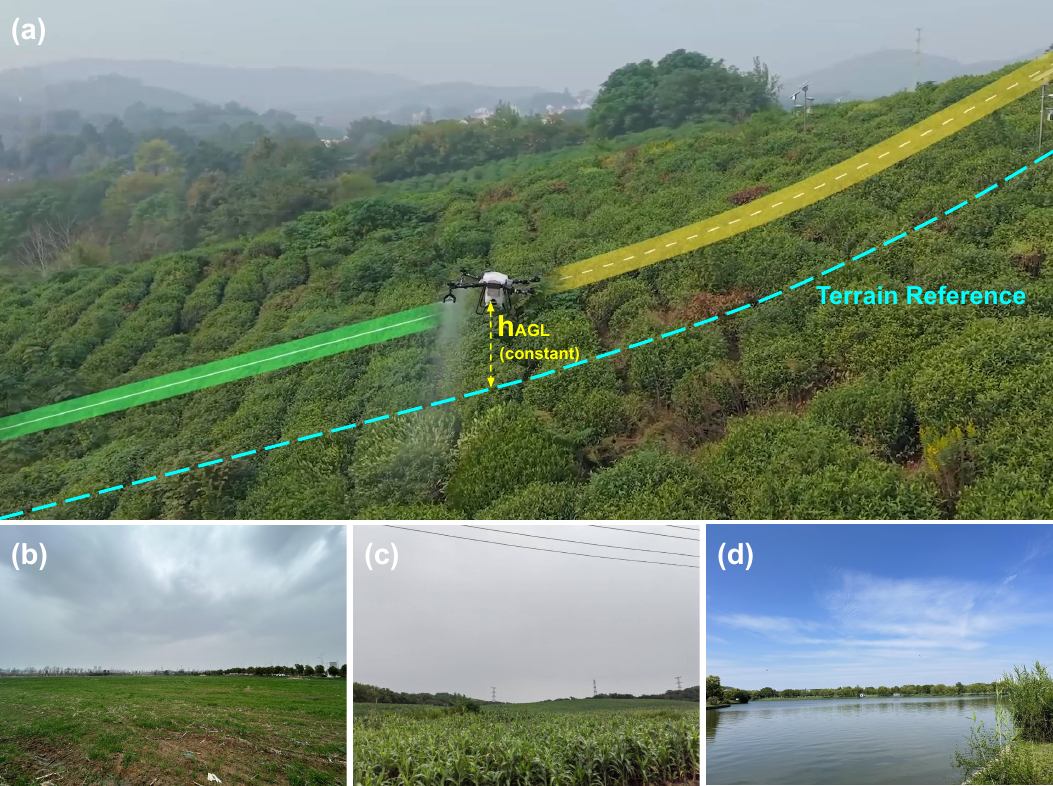}
    \caption{\textbf{Terrain-following task and representative operating scenarios for agricultural UAVs.} 
    (a) Overall terrain-following process in a \textit{tea plantation hill} environment, where the planned flight path adapts to terrain elevation variations to maintain a nearly constant above-ground level. 
    (b)–(d) Three additional agricultural scenarios covered by the terrain-following task: \textit{flat field}, \textit{sloped terrain}, and \textit{water area}, respectively. 
    Green solid curve: executed UAV trajectory; yellow dashed curve: target terrain-following path.}

    \vspace{-0.3cm}
    \label{fig:teaser}
\end{figure}

\IEEEPARstart{W}{ith} the rapid development of agricultural mobile robotics, autonomous robotic platforms have become increasingly important for improving operational efficiency and reducing labor requirements in modern precision agriculture~\cite{nguyen2025semi,wembe2026predictive}. 
Among them, uncrewed aerial vehicles (UAVs) have become important platforms in precision agriculture owing to their flexibility and high operational efficiency in tasks such as spraying, fertilization, inspection and surveying~\cite{cheng2023precision,radoglou2020compilation,he2025multi}.
They are especially valuable in hilly, mountainous, and terraced farmlands, where conventional agricultural machinery is often difficult to deploy. 
In such environments, agricultural UAVs can support effective low-altitude agricultural operations through terrain-following flight, which allows the vehicle to adapt to varying ground elevation while maintaining a relatively stable above-ground flight height, as illustrated in Fig.~\ref{fig:teaser}.
Such height stability directly affects spray quality and operational safety: excessive flight height may lead to droplet drift and reduced deposition efficiency, whereas insufficient flight height can result in poor coverage or even collision risks.
However, achieving such terrain-following capability, especially in complex field environments, fundamentally depends on accurate terrain perception.

\begin{figure*}[t]
    \centering
    \setlength{\tabcolsep}{2pt}
    \renewcommand{\arraystretch}{1.1}
    \begin{tabular}{c c c c}
        \toprule
        & \textbf{Camera} & \textbf{LiDAR} & \textbf{mmWave Radar} \\
        
        \midrule
        \rotatebox{90}{\parbox{4cm}{\centering\textbf{Low}\\\textbf{Illumination}}}
        & \includegraphics[width=0.3\textwidth]{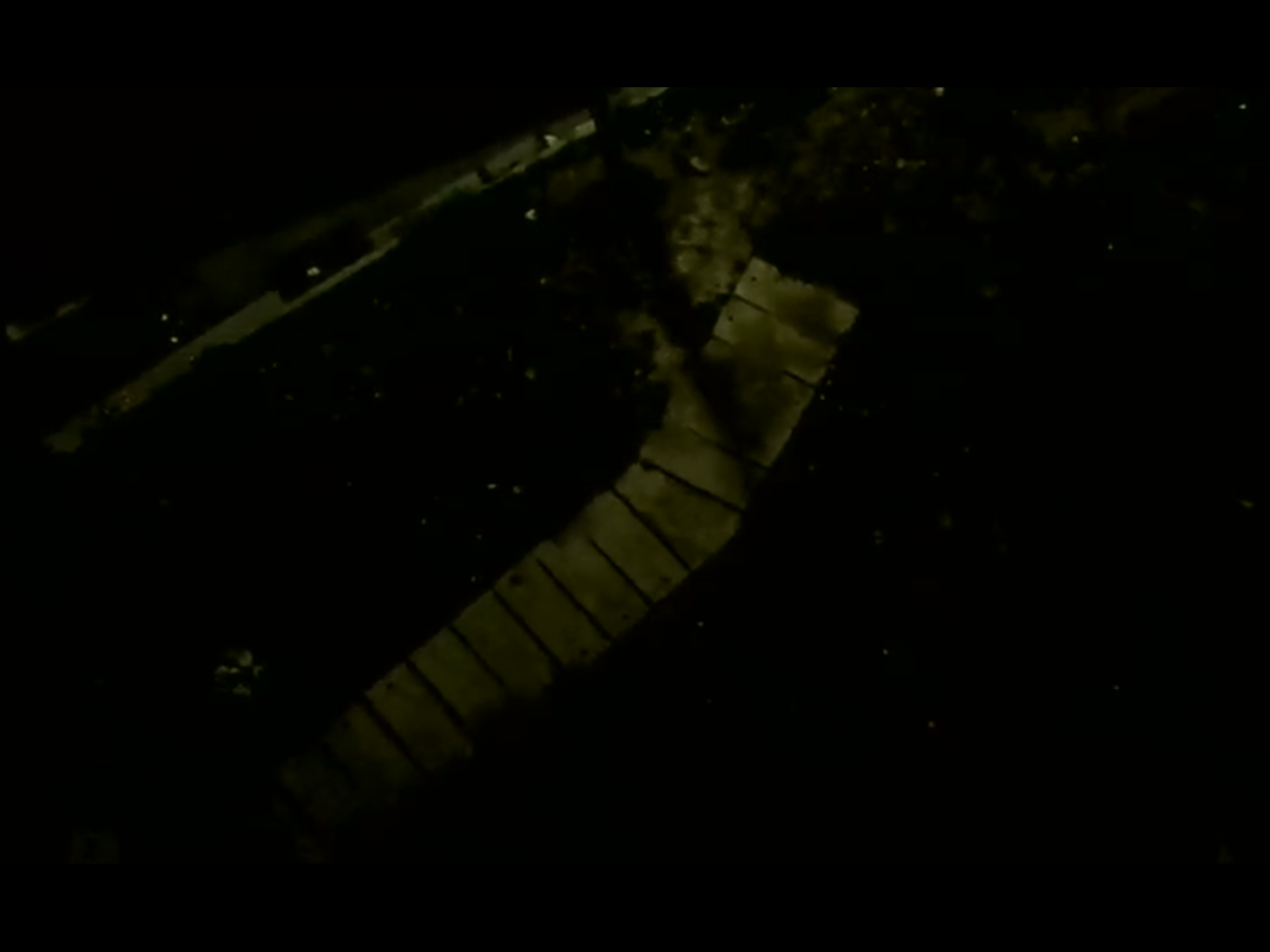}
        & \includegraphics[width=0.3\textwidth]{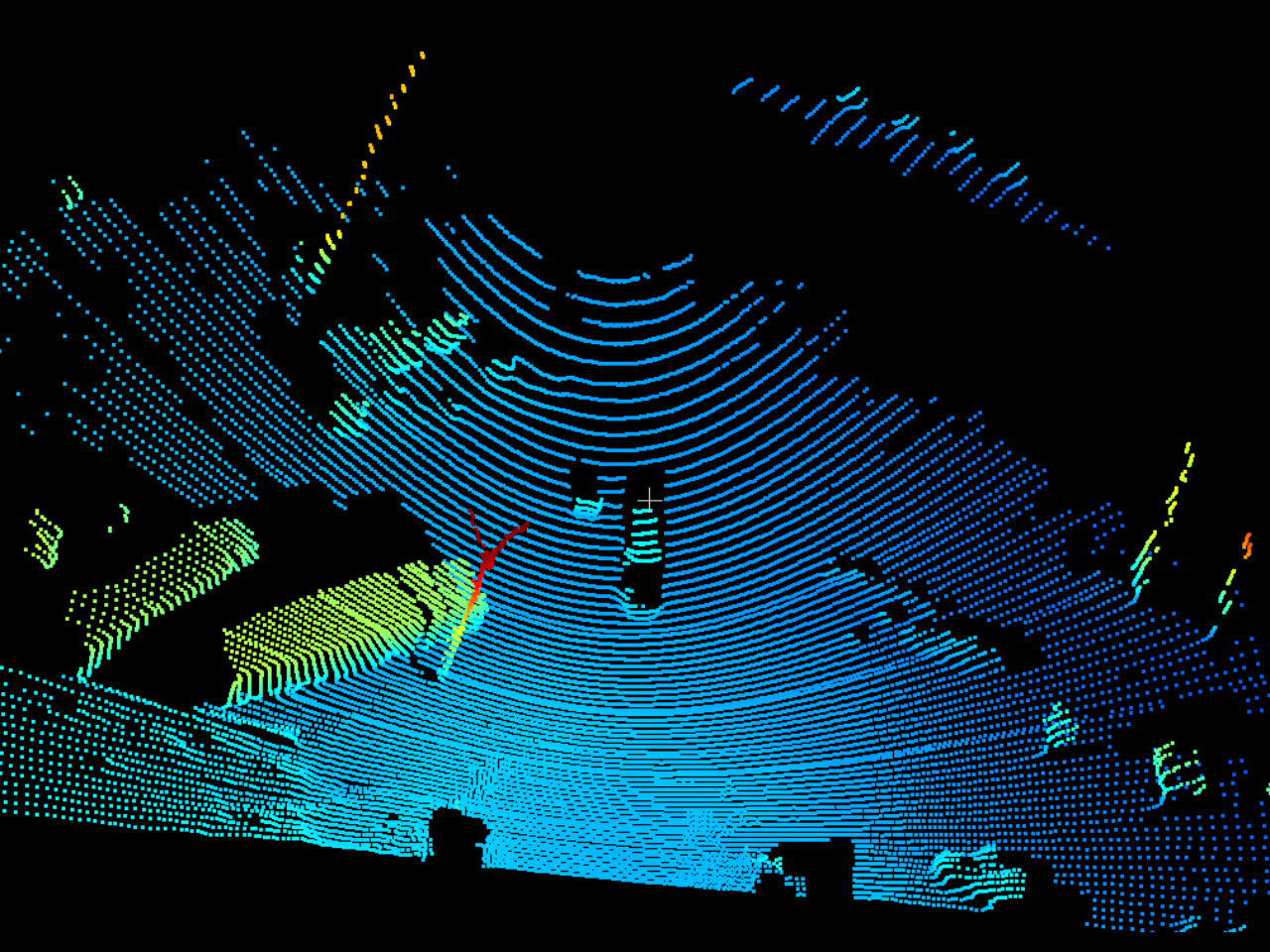}
        & \includegraphics[width=0.3\textwidth]{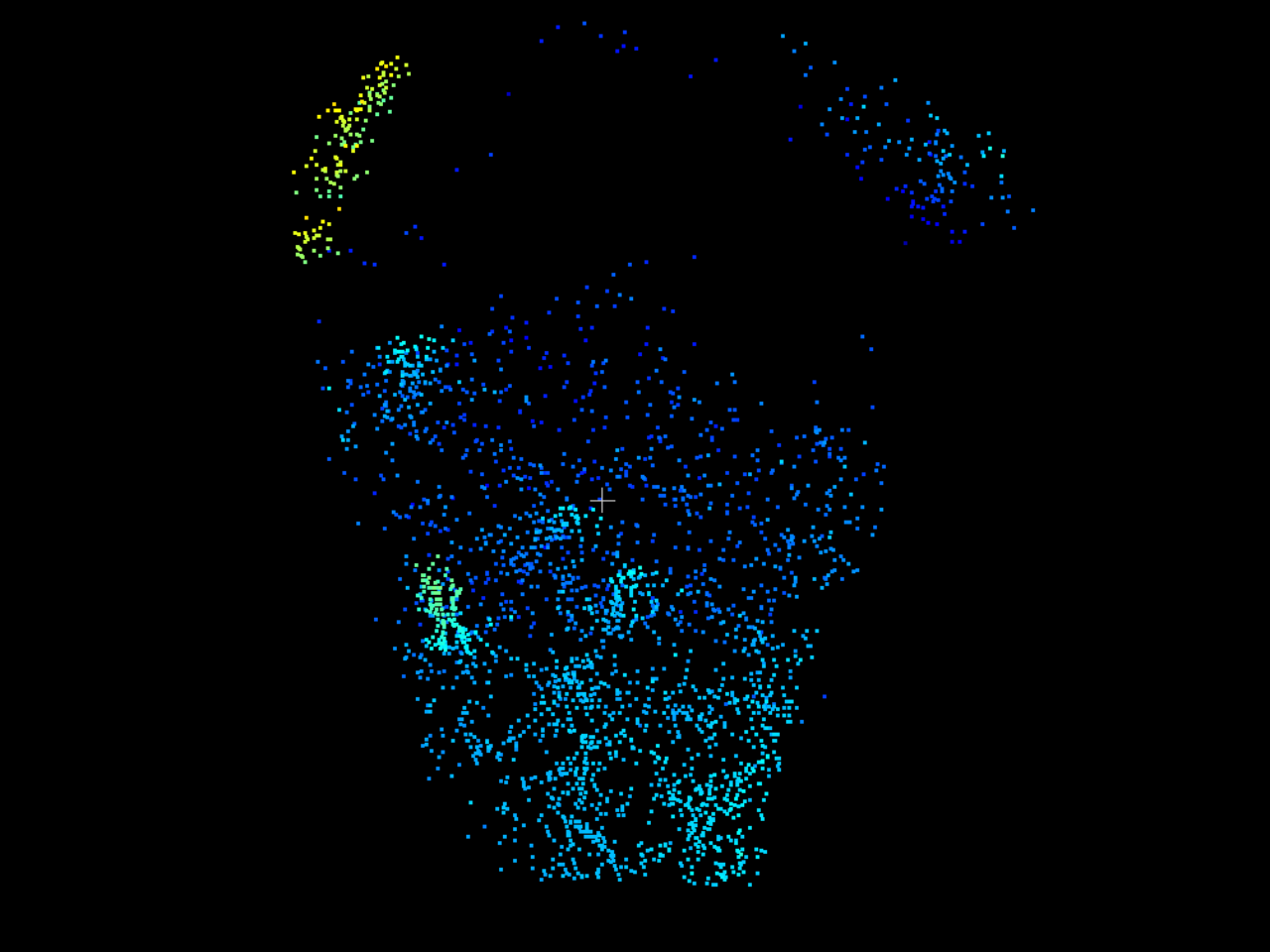} \\

        \rotatebox{90}{\parbox{4cm}{\centering\textbf{Spraying}\\\textbf{Disturbance}}}
        & \includegraphics[width=0.3\textwidth]{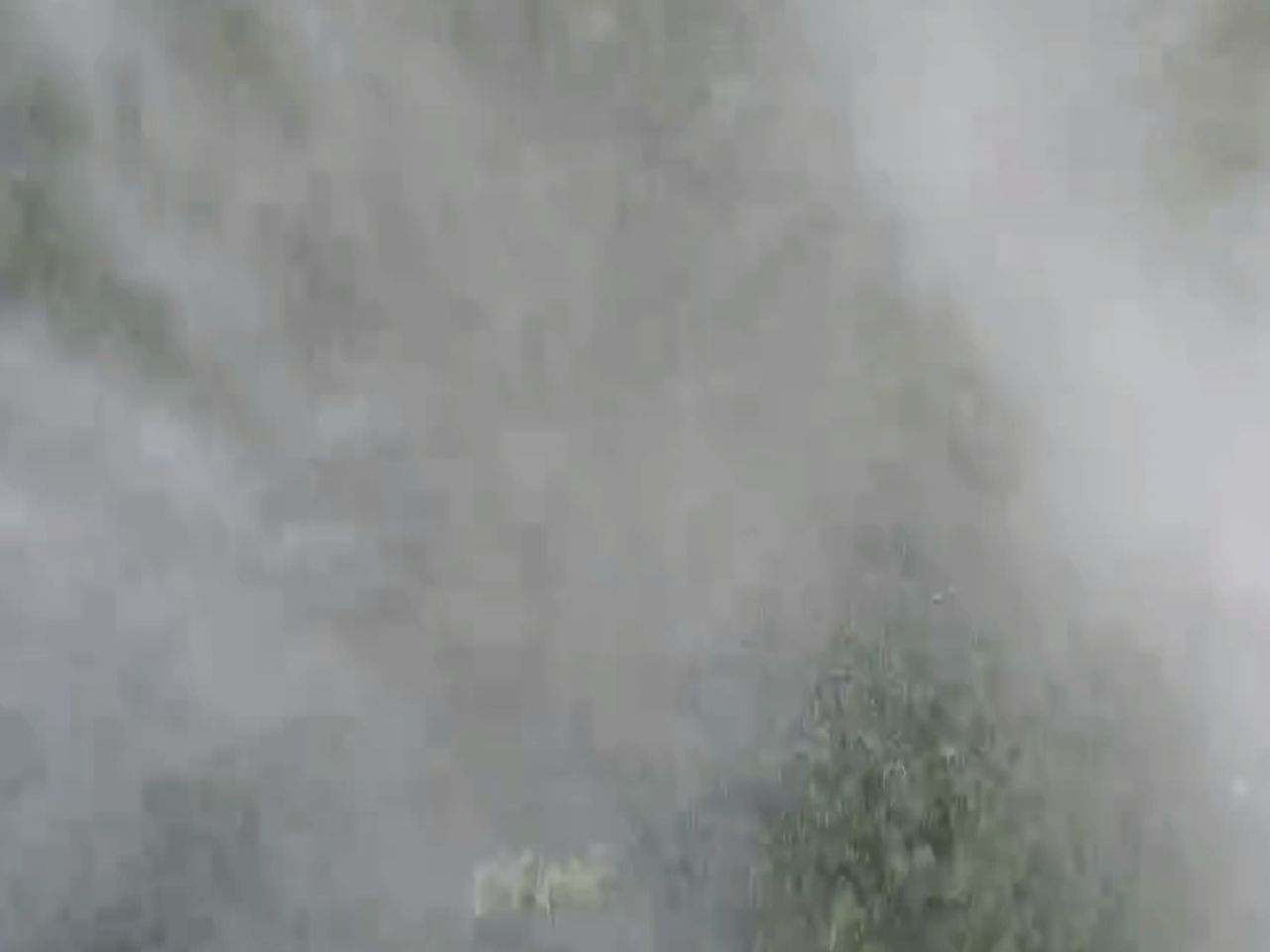}
        & \includegraphics[width=0.3\textwidth]{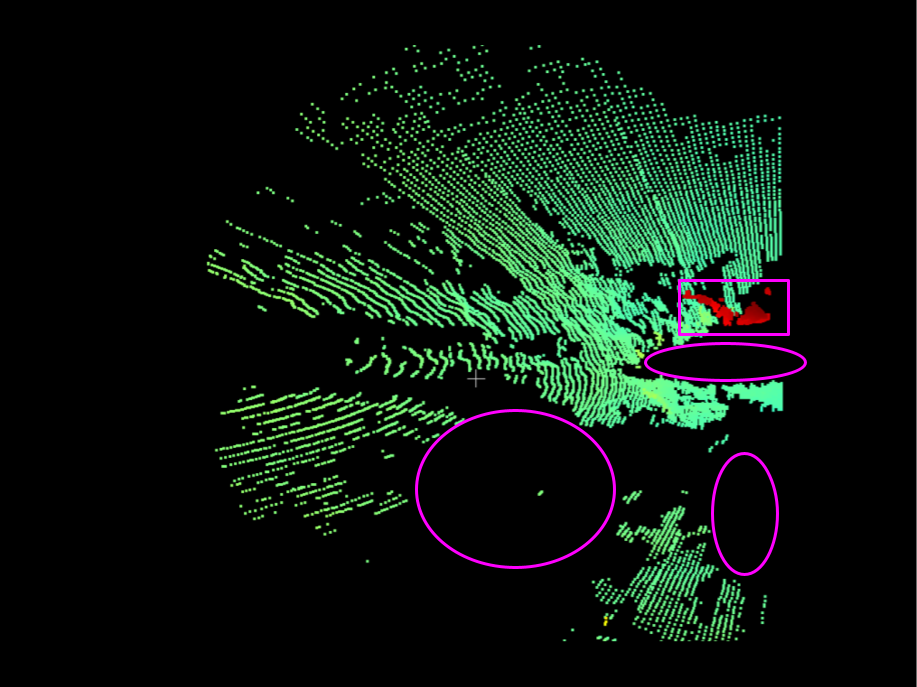}
        & \includegraphics[width=0.3\textwidth]{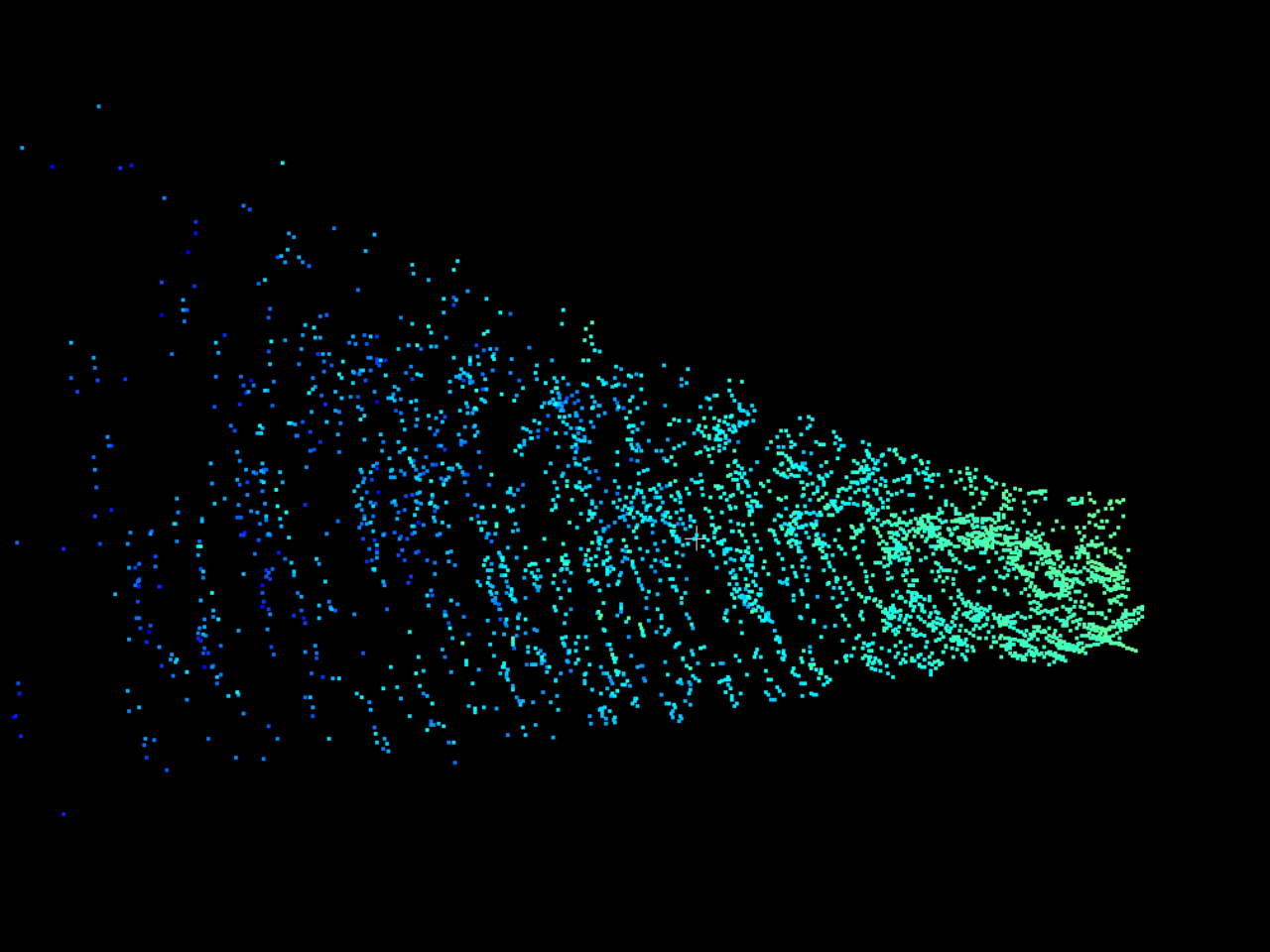} \\

        \bottomrule
    \end{tabular}
    \caption{\textbf{Qualitative comparison of camera, LiDAR, and the proposed rotating radar sensing observations under representative agricultural field conditions.}
    The point clouds are color-coded by height. In challenging scenarios such as low illumination and spraying disturbance, the camera exhibits degraded visual observations, while LiDAR suffers from partial terrain occlusion under spraying interference. 
    In contrast, the proposed rotating radar sensing system provides more robust terrain observations, highlighting its suitability for agricultural UAV terrain perception.}

    \vspace{-0.3cm}
    \label{fig:sensor_comparison}
\end{figure*}

To enable accurate terrain perception, various onboard sensing modalities have been explored, with common choices being cameras, light detection and ranging (LiDAR), and millimeter-wave (mmWave) radar.
Cameras provide rich visual information at low cost, but their performance is sensitive to illumination changes, shadows, weak texture, and adverse weather conditions~\cite{wang2024uav,zang2019impact,bijelic2020seeing}.
LiDAR provides accurate 3D geometry and strong terrain modeling capability, but its higher cost and reported degradation in rain, fog, or dust may limit deployment in agricultural environments~\cite{farhan2024comprehensive,zang2019impact,bijelic2020seeing,qian2021robust}.
In contrast, mmWave radar is relatively less sensitive to lighting variations and certain adverse environmental conditions while remaining cost-effective, which makes it a promising sensing modality for agricultural UAV terrain perception~\cite{corradi2022radar,doer2022gnss,zhan2025agrilira4d,qian2021robust,yang2025ralibev}.
A qualitative comparison of camera, LiDAR, and mmWave radar observations under representative agricultural field conditions is shown in Fig.~\ref{fig:sensor_comparison}.

Despite the availability of various sensing modalities, achieving robust and accurate terrain perception in real agricultural environments remains challenging.
Agricultural fields often feature rolling hills, terraces, gullies, and uneven ground surfaces, which complicate terrain estimation~\cite{tarolli2020agriculture}.
In addition, the ground surface is often partially occluded by crops, weeds, or tree canopies, making direct observation of the underlying terrain difficult.
Low-altitude operation further subjects onboard sensing to illumination changes, dust, moisture, and other environmental disturbances, thereby degrading measurement stability and reliability.
These factors jointly lead to incomplete, noisy, or unstable observations of the ground surface, making accurate terrain estimation difficult in real-world UAV deployments.

Among the available sensing modalities, mmWave radar is particularly promising for agricultural UAV applications due to its robustness to lighting variations and certain environmental disturbances.
However, several key limitations hinder its effectiveness for terrain perception in practice.
During low-altitude UAV flight, terrain slope and platform motion continuously change the relative geometry between the radar sensor and the ground surface, leading to unstable observation conditions.
Existing UAV-mounted radar systems commonly employ a fixed field of view (FoV), which restricts spatial coverage and often results in sparse and incomplete terrain observations~\cite{zhan2025agrilira4d}.
While rotating and scanning sensing mechanisms have been widely explored in LiDAR-based robotic perception and mapping~\cite{zhang2014loam,fu2023rotating,li2025ua,chen2026spinning,chen2023self}, existing scanning-radar systems are more commonly studied on ground vehicles and often provide 2D range-azimuth observations rather than explicit 3D terrain representations~\cite{barnes2020oxford}.
Commercial agricultural UAVs may also employ active phased-array radar for obstacle sensing or terrain following, but such systems are comparatively expensive and not openly documented.
In addition, platform motion and pitch variations further alter the radar viewing geometry, making it difficult for a fixed-FoV radar to maintain consistent observation of the ground surface.
Moreover, terrain extraction from radar observations remains challenging, since existing ground segmentation methods are largely designed for dense LiDAR-style point clouds~\cite{zermas2017fast,lim2021patchwork,lee2022patchwork++}, whereas radar point clouds are typically sparse, noisy, and limited in angular resolution.
As a result, it remains challenging to obtain reliable and continuous terrain representations for terrain-following control using radar sensing alone.
This motivates the need for a radar sensing configuration with broader spatial support, as well as a terrain extraction pipeline tailored to radar observation characteristics.

To address these challenges, this paper presents a terrain perception system based on a rotating mmWave radar for agricultural UAVs, which provides broader spatial coverage and more stable ground observation under dynamic flight conditions.
The system consists of three key components: rotating radar sensing, ground extraction, and terrain modeling, forming an onboard perception pipeline for terrain-aware UAV operation.
Specifically, the rotating radar sensing module expands the effective FoV and mitigates observation inconsistency caused by platform motion; the ground extraction module identifies ground points from sparse and noisy radar measurements; and the terrain modeling module generates a continuous terrain representation for terrain-following operation.
To the best of our knowledge, this is one of the first works to explore rotating mmWave radar sensing for terrain perception in agricultural UAV applications.

The main contributions of this work are summarized as follows.
\begin{enumerate}
    \item We develop a \textbf{low-cost rotating mmWave radar sensing system for agricultural UAV terrain perception in complex farmland environments}, enabling broader spatial coverage and more stable ground observation under hilly terrain, terraced fields, crop occlusions, and sparse ground visibility, while providing a practical sensing alternative to heavier or more expensive perception solutions for low-altitude agricultural autonomy.
    \item We propose a \textbf{pose-aware ground extraction and continuous terrain modeling pipeline} specifically designed for sparse, noisy, and partially observable rotating radar measurements, integrating registration, FoV-constrained filtering, temporal accumulation, region-wise geometric analysis, and prior-guided refinement to achieve robust terrain reconstruction under conditions where existing generic ground segmentation methods are often unreliable.
    \item We establish a \textbf{complete real-world UAV deployment and validation framework}, including onboard rotating radar hardware integration, online terrain modeling, and terrain-referenced flight support, together with extensive field experiments across representative agricultural scenarios to demonstrate practical deployability, robustness, and system-level effectiveness in real operational conditions.

\end{enumerate}

The remainder of this paper is organized as follows. 
Section~\ref{sec:related} reviews the related work on agricultural UAV terrain perception, sensing modalities, and point-cloud-based ground segmentation and terrain modeling. 
Section~\ref{sec:method} presents the proposed system design and methodology, including rotating radar sensing, ground segmentation and terrain modeling, and UAV system integration and deployment. 
Section~\ref{sec:exp} provides the experimental evaluation, including ground segmentation, terrain modeling accuracy, terrain-following consistency analysis, and a discussion of practical deployment limitations. 
Finally, Section~\ref{sec:conclusion} concludes the paper and outlines future work.

\section{RELATED WORK}
\label{sec:related}

\subsection{Vision-based Terrain Perception}
Vision-based terrain perception has been extensively studied in ground robotic platforms, particularly for off-road traversability analysis~\cite{cai2024riskaware,fan2018road}, terrain classification~\cite{zurn2021terrainclassification}, and elevation prediction~\cite{chung2024elevation}. 
Representative works learn terrain traversability or properties directly from images using self-supervision, contrastive learning, or uncertainty-aware modeling~\cite{chen2024terrainparameter, xue2025traversability,fan2019pothole}. 

However, most of these methods are designed for wheeled or legged robots operating in ground-level off-road environments, rather than for onboard perception from an aerial viewpoint. In the UAV domain, vision-based terrain perception remains relatively limited and is mainly explored for terrain classification, autonomous landing, or monocular height/elevation estimation. 
Early work by Khan \emph{et al.} investigated visual terrain classification for flying robots using a monocular onboard camera~\cite{khan2012visual}. 
Forster \emph{et al.} proposed continuous onboard monocular elevation mapping for micro aerial vehicle landing~\cite{forster2015continuous}, while Campos \emph{et al.} presented a monocular-vision-based height estimation approach for terrain-following flight~\cite{campos2016height}. 
Related UAV-oriented work also includes vision-based terrain following for an uncrewed rotorcraft~\cite{garratt2008vision} and terrain classification from UAV flights using monocular vision~\cite{campos2015terrain}. 

Although these studies demonstrate the feasibility of vision-only terrain understanding from UAVs, their performance is sensitive to illumination variation, weak texture, vegetation occlusion, and adverse weather, which limits robustness in complex agricultural environments.

\subsection{LiDAR-based Terrain Perception}
LiDAR-based terrain perception has been widely studied for both aerial mapping and ground robotic navigation. 
In the UAV domain, existing works primarily focus on high-accuracy topographic surveying, digital terrain model (DTM) generation, and elevation modeling. 
Representative studies include UAV-LiDAR topographic mapping for precision land levelling~\cite{du2022uavlidar}, terrain quantification in vegetated fluvial environments~\cite{macdonell2023consumer}, ultra-high-resolution elevation modeling for urban flood risk assessment~\cite{trepekli2022uav}, and accuracy evaluation of UAV LiDAR in coastal environments~\cite{lin2019evaluation}. 
Recent studies further examine sensor-dependent DTM quality~\cite{bartminski2023effectiveness}, filtering strategies for precision mapping~\cite{oniga2024enhancing}, and field-deployable systems for high-resolution terrain acquisition in complex environments~\cite{choi2023acquisition}. 

In contrast, LiDAR-based terrain perception for ground vehicles more commonly emphasizes online ground segmentation, terrain estimation, and traversability analysis, as demonstrated by methods such as GroundGrid~\cite{steinke2024groundgrid}, probabilistic graph-based real-time ground segmentation~\cite{delpino2023probabilistic}, and LiDAR-based terrain modeling for autonomous driving~\cite{xue2023traversability}. 

Although LiDAR provides accurate 3D geometry and strong terrain modeling capability, existing UAV-oriented works are largely focused on offline mapping rather than real-time onboard terrain perception. 
Moreover, the higher sensor cost and reduced robustness in adverse conditions such as rain, fog, and dust may limit large-scale deployment in complex agricultural environments.

\begin{figure*}[htbp]
    \centering
    \includegraphics[width=1.0\textwidth]{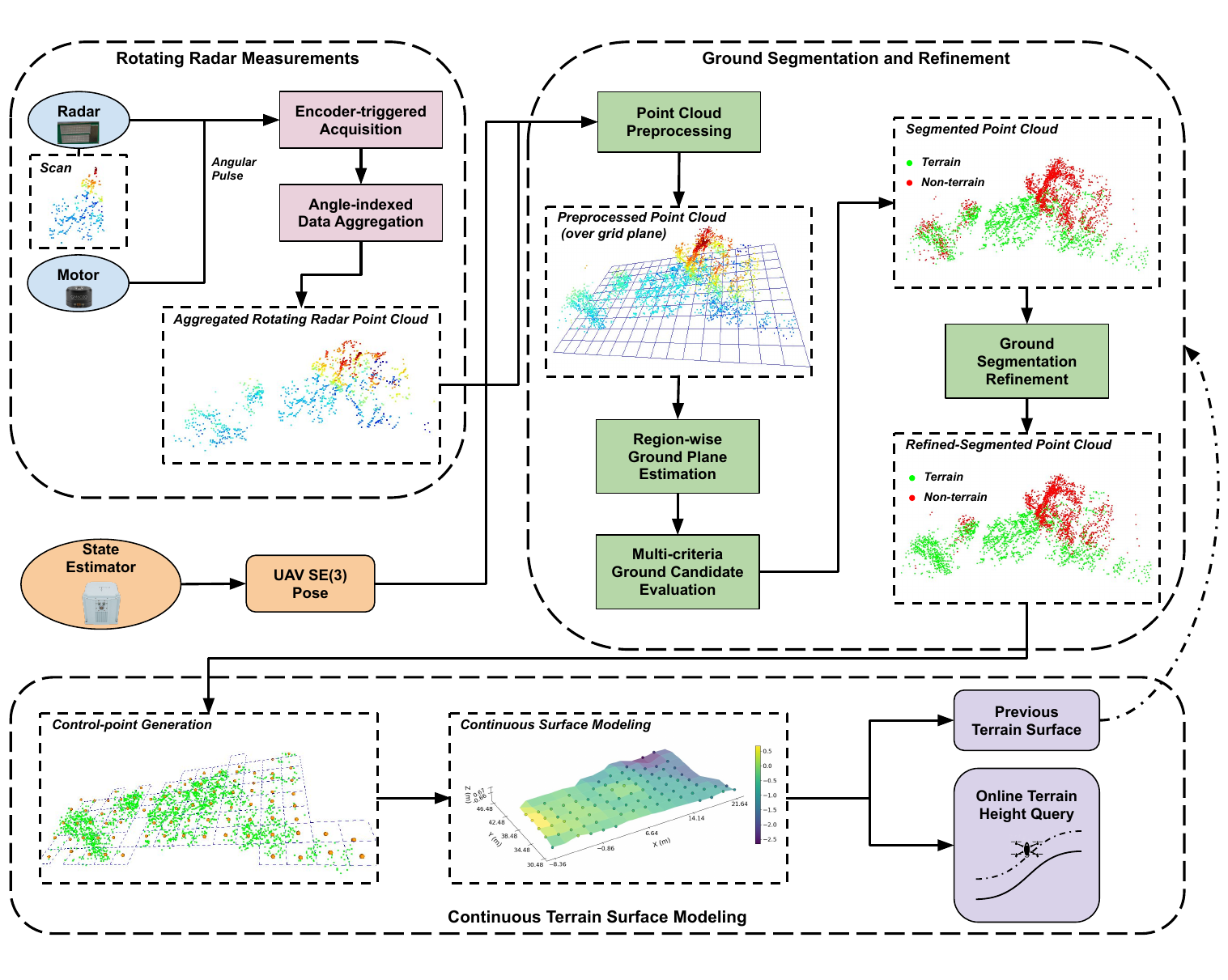}
    \caption{
    \textbf{System Overview.} 
Our proposed system consists of three main components: \textit{rotating radar sensing}, \textit{radar-specific ground extraction}, and \textit{terrain modeling}. 
In \textit{rotating radar sensing}, measurements are generated by encoder-triggered acquisition and angle-indexed aggregation, which are then fused with UAV pose to form a point cloud. 
In \textit{radar-specific ground extraction}, the point cloud is processed via grid-based partitioning and region-wise estimation with multi-criteria evaluation to extract terrain points from sparse radar observations. 
Finally, in \textit{terrain modeling}, the extracted terrain points are used to construct a continuous surface for online terrain height query in terrain-following flight.
}
    \vspace{-0.3cm}
    \label{fig:system_overview}
\end{figure*}

\subsection{Radar-based Terrain Perception}
Radar-based perception has attracted increasing attention in robotics due to its robustness to poor lighting, dust, fog, and other adverse environmental conditions~\cite{harlow2024newwave}. 
In ground robotics and autonomous driving, prior studies have investigated radar-based ground segmentation~\cite{reina2011radar}, ground-aware odometry~\cite{herraez2025groundaware}, and radar-inertial estimation~\cite{yang2025rio}, demonstrating that radar can support terrain-related perception in challenging outdoor scenarios. 
However, these works also highlight inherent challenges, including sparse and noisy measurements, limited angular resolution, and sensitivity to sensor configuration and motion.

In the aerial domain, radar-related studies remain limited and are primarily focused on height estimation, landing assistance, or general perception enhancement rather than explicit terrain perception. 
Representative examples include micro-drone ego-velocity and height estimation using FMCW radar~\cite{barra2023microdrone}, mmWave altimetry for uncrewed aerial systems~\cite{awan2024altimetry}, and radar-assisted drone landing~\cite{wang2025landing}. 
Recent cross-modal approaches further explore densifying sparse radar observations~\cite{zhang2024radardiffusion} or learning robust radar representations for navigation~\cite{huang2021cmclr}. 

Despite these advances, radar-based terrain perception for agricultural UAVs remains underexplored, particularly for low-altitude operation over complex farmland, where both broad spatial coverage and radar-specific terrain extraction are required.

\section{SYSTEM DESIGN AND METHODOLOGY}
\label{sec:method}

\subsection{System Overview}

As illustrated in Fig.~\ref{fig:system_overview}, we present a terrain perception system for agricultural UAVs.
The system takes rotating radar measurements and UAV pose as input, and outputs a continuous terrain representation that supports real-time terrain height query during low-altitude operation.

The proposed system consists of three main modules: rotating radar sensing, ground extraction, and terrain modeling. 
First, a rotating mmWave radar is employed to enlarge the effective FoV and improve terrain observability under dynamic flight conditions, where a rotary encoder associates radar measurements with the corresponding rotation angle. 
The acquired radar points are further transformed into the global coordinate frame using pose information from the UAV state estimation system. 
Second, the ground extraction module identifies ground points from sparse and noisy radar observations for subsequent terrain modeling. 
Third, based on the extracted ground points, terrain control points are generated on a regular grid, from which a continuous 2.5-D terrain model is constructed for real-time height querying and incremental update.

We deploy the proposed system on an agricultural UAV platform, where radar measurements, UAV state information, and terrain perception results are exchanged online via the Robot Operating System (ROS). 
The resulting terrain model provides the perception basis for terrain-following operation in complex agricultural environments.

\begin{figure*}[htbp]
    \centering
    \includegraphics[width=0.95\textwidth]{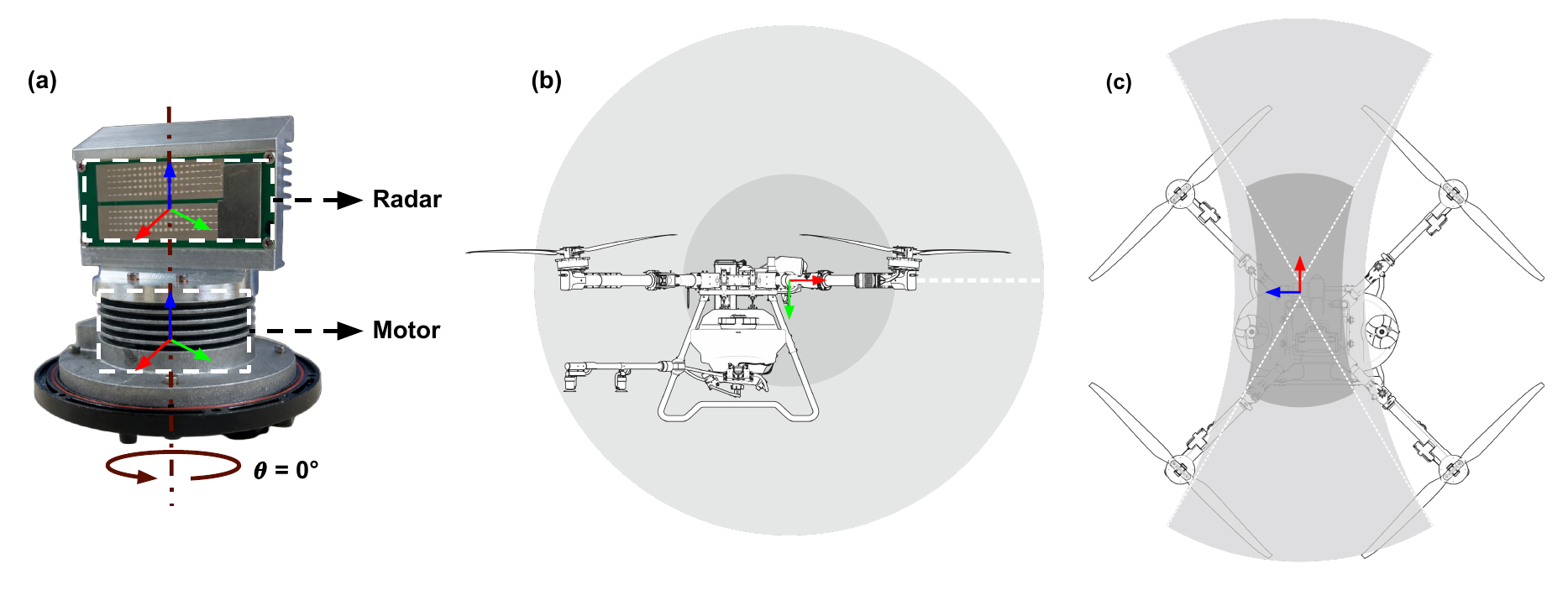}
    \caption{\textbf{Rotating radar sensing configuration used in the proposed terrain perception system.} 
    (a) Internal mechanical structure of the rotating radar unit, showing the radar sensor, motor-driven rotating mechanism, and the corresponding radar and motor frames. 
    (b) Side view of the sensing geometry, illustrating the radar FoV during rotation. 
    (c) Effective ground-projected FoV of the rotating radar. 
    Since part of the instantaneous FoV points toward the sky during rotation, the ground projection is provided to clarify the actual terrain observation coverage.
    }
    \vspace{-0.3cm}
    \label{fig:rotating_radar_sensing}
\end{figure*}

\subsection{Rotating Radar Sensing}

Conventional mmWave radar sensing with a fixed FoV provides limited spatial coverage for terrain observation, especially under UAV motion and terrain variation. 
To address this limitation, we design a rotating mmWave radar sensing unit, as illustrated in Fig.~\ref{fig:rotating_radar_sensing}. 
The unit employs a 60~GHz mmWave radar with an instantaneous FoV of $60^\circ$ in the horizontal direction and $8^\circ$ in the vertical direction.
The radar is mounted on a motor-driven platform to continuously scan the surrounding terrain, thereby expanding the effective FoV during flight.
The complete rotating sensing unit, including the radar module and encoder-equipped motor, has a hardware cost of approximately \$200. 
For comparison, representative lightweight 3D LiDAR sensors and commercial phased-array radar modules for agricultural UAVs are generally in the kilodollar price range. 
This cost difference provides additional practical motivation for extending radar coverage through mechanical rotation.

A rotary encoder is integrated with the rotating mechanism to provide real-time angular feedback. 
Through encoder-triggered acquisition, radar measurements are sampled at predefined angular intervals (with a step of $5^\circ$), such that each measurement is associated with a known rotation angle, enabling spatially consistent reconstruction of angle-aware radar points. 
Using pose information from the UAV state estimation system, the radar points are further transformed into the global coordinate frame.

Radar measurements are aggregated into point cloud frames over a fixed scanning interval, yielding an effective frame rate of 10~Hz. 
Each angle-associated scan is timestamped and temporally aligned with UAV pose estimates (400~Hz) to maintain consistency between sensing and platform motion. 
The resulting synchronized radar point cloud frames serve as input to the subsequent ground extraction and terrain modeling modules.

\subsection{Ground Segmentation and Refinement}

To address the sparse, noisy, and non-uniform observations, we design a four-step radar-specific ground segmentation pipeline.

\subsubsection{Point Cloud Preprocessing}

Due to platform motion and the rotating sensing mechanism, raw radar measurements are not directly suitable for terrain analysis. 
We therefore design a preprocessing pipeline consisting of pose-aware point registration, FoV filtering, short-window temporal accumulation, and grid-based partitioning, to improve spatial consistency and support.

The proposed rotating radar sensing system involves four coordinate frames: the radar frame $\{R\}$, the motor frame $\{M\}$, the UAV body frame $\{B\}$, and the world frame $\{W\}$. 
Each radar point ${}^{R}\mathbf{p}_i$ is sequentially transformed from $\{R\}$ to $\{M\}$ using the encoder-associated rotation angle $\theta_i$, then to $\{B\}$ via the rigid mounting transformation, and finally to $\{W\}$ using the UAV pose estimate:
\begin{equation}
{}^{\Xi}\mathbf{p}_i
=
{}^{\Xi}_{\Theta}\mathbf{R}\,{}^{\Theta}\mathbf{p}_i
+
{}^{\Xi}_{\Theta}\mathbf{t},
\end{equation}
where $\Xi$ and $\Theta$ denote two coordinate frames, and ${}^{\Xi}_{\Theta}\mathbf{R}\in SO(3)$ and ${}^{\Xi}_{\Theta}\mathbf{t}\in\mathbb{R}^{3}$ represent the rigid transformation from $\{\Theta\}$ to $\{\Xi\}$.

After registration, radar measurements are grouped into point cloud frames over a fixed scanning interval. 
Let $\mathcal{P}_{f}=\{{}^{W}\mathbf{p}_i\}$ denote the registered point cloud of the $f$-th frame.

To suppress measurements that are unlikely to contribute to terrain estimation (e.g., airborne clutter or lateral structures), we apply a pose-aware FoV filtering step. 
Since the UAV attitude varies during flight, the valid ground-facing observation region cannot be defined in a fixed local frame. 
Instead, the filtering is performed in the world frame $\{W\}$ according to the current UAV pose.

Let
\begin{equation}
\mathbf{r}_i = {}^{W}\mathbf{p}_i - {}^{W}_{B}\mathbf{t}
\end{equation}
be the vector from the UAV position to the point. 
Only points within a cone of $\pm \phi$ around the downward direction are retained:
\begin{equation}
\mathcal{P}_{f}^{\mathrm{fov}}
=
\left\{
{}^{W}\mathbf{p}_i \in \mathcal{P}_{f}
\;\middle|\;
\frac{-r_{i,z}}{\|\mathbf{r}_i\|}
\ge
\cos(\phi)
\right\}.
\end{equation}

Despite FoV filtering, mmWave radar observations remain sparse. 
To improve spatial support, we adopt a short-window temporal accumulation strategy:
\begin{equation}
\mathcal{P}_{f}^{\mathrm{acc}} = \bigcup_{k=0}^{K-1} \mathcal{P}_{f-k}^{\mathrm{fov}}.
\end{equation}
Since terrain is locally static over a short time interval and all frames are registered in $\{W\}$, this aggregation produces a denser and more complete representation.

To address non-uniform spatial distribution and locally varying terrain geometry, we further partition the accumulated point cloud into regular local regions.

Specifically, a Cartesian grid with resolution $s$ is defined on the X–Y plane of $\{W\}$. 
For ${}^{W}\mathbf{p}_i=[x_i,y_i,z_i]^{\top}$, the grid index is
\begin{equation}
u_i=\left\lfloor \frac{x_i}{s}\right\rfloor,\qquad
v_i=\left\lfloor \frac{y_i}{s}\right\rfloor.
\end{equation}
The corresponding cell point set is
\begin{equation}
\mathcal{P}_{u,v}
=
\left\{
{}^{W}\mathbf{p}_i \in \mathcal{P}_{f}^{\mathrm{acc}}
\;\middle|\;
(u_i,v_i)=(u,v)
\right\}.
\end{equation}

Each grid cell serves as a basic unit for subsequent region-wise ground estimation and terrain modeling.

\subsubsection{Region-wise Ground Plane Estimation}

Given the grid-partitioned point sets, we estimate a local ground plane for each valid grid cell. 
Despite the preprocessing steps that improve spatial density and consistency, the resulting radar observations remain sparse and noisy, making direct plane fitting unstable.
We therefore adopt a three-stage strategy: prior-constrained seed initialization, principal component analysis (PCA)-based plane estimation, and iterative refinement.

For a grid cell $\mathcal{P}_{u,v}$, only cells satisfying
\begin{equation}
|\mathcal{P}_{u,v}| \ge N_{\min}
\end{equation}
are considered valid.

\paragraph{Prior-constrained Seed Initialization}
Direct plane fitting on sparse radar points is highly sensitive to outliers, especially in strong multi-path scenarios such as water surfaces. 
To obtain a stable initialization, we leverage an online historical terrain prior to constrain candidate ground points.

Let $z_{u,v}^{\mathrm{hist}}$ denote the terrain height queried from the historical model reconstructed so far in the current flight. 
If the terrain height cannot be queried, such as during the initial takeoff stage before sufficient terrain observations have been accumulated, we set $z_{u,v}^{\mathrm{hist}}=0$ as the default initialization.
This initialization is consistent with the adopted local world frame, whose origin is anchored at the UAV takeoff position with the $z$-axis aligned vertically, such that the ground surface at the takeoff location is approximately at $z=0$.
In the intended agricultural operating setting, takeoff is performed from an approximately level and open area.
Once a valid terrain height becomes available from the online terrain model, the queried value replaces the default prior.

A point ${}^{W}\mathbf{p}_j=[x_j,y_j,z_j]^{\top}$ is selected as a candidate if
\begin{equation}
z_{u,v}^{\mathrm{hist}}-\delta_{\mathrm{lower}}
\le z_j \le
z_{u,v}^{\mathrm{hist}}+\delta_{\mathrm{upper}}.
\end{equation}
The lower and upper elevation bounds provide tolerance around the terrain prior when selecting seed candidates.

If the number of candidates is greater than or equal to $k$, the $k$ lowest ones are selected as the initial seed set $S_{u,v}$; otherwise, all candidates are used. 
The mean height is computed as
\begin{equation}
z_{\mathrm{init}} = \frac{1}{|S_{u,v}|} \sum_{{}^{W}\mathbf{p}_j \in S_{u,v}} z_j,
\end{equation}
and the initial ground set is defined as
\begin{equation}
\mathcal{G}_{u,v}^{(0)} = \left\{
{}^{W}\mathbf{p}_j \in \mathcal{P}_{u,v}
\;\middle|\;
z_j \le z_{\mathrm{init}}+\delta_{\mathrm{seed}}
\right\}.
\end{equation}

\paragraph{PCA-based Plane Estimation}
Given the current ground set $\mathcal{G}_{u,v}^{(t)}$, we estimate a local plane using PCA~\cite{nurunnabi2014diagnostics}. 
PCA is adopted here because, as discussed in Patchwork~\cite{lim2021patchwork}, it provides a more computationally efficient alternative to iterative random sample consensus (RANSAC)-based fitting for local ground estimation, which is beneficial for real-time onboard operation. 
The centroid is
\begin{equation}
\bar{\mathbf{p}}_{u,v}^{(t)} = 
\frac{1}{|\mathcal{G}_{u,v}^{(t)}|} \sum_{{}^{W}\mathbf{p}_j \in \mathcal{G}_{u,v}^{(t)}} {}^{W}\mathbf{p}_j,
\end{equation}
and the covariance matrix is
\begin{equation}
\mathbf{C}_{u,v}^{(t)}
=
\frac{1}{|\mathcal{G}_{u,v}^{(t)}|}
\sum_{{}^{W}\mathbf{p}_j \in \mathcal{G}_{u,v}^{(t)}}
\left({}^{W}\mathbf{p}_j-\bar{\mathbf{p}}_{u,v}^{(t)}\right)
\left({}^{W}\mathbf{p}_j-\bar{\mathbf{p}}_{u,v}^{(t)}\right)^{\top}.
\end{equation}

The eigenvector corresponding to the smallest eigenvalue is taken as the normal $\mathbf{n}_{u,v}^{(t)}$, and the plane is defined by
\begin{equation}
\left(\mathbf{n}_{u,v}^{(t)}\right)^{\top}\mathbf{p} + d_{u,v}^{(t)} = 0,
\end{equation}
where
\begin{equation}
d_{u,v}^{(t)} = -\left(\mathbf{n}_{u,v}^{(t)}\right)^{\top}\bar{\mathbf{p}}_{u,v}^{(t)}.
\end{equation}

\paragraph{Iterative Refinement}
Due to residual outliers and imperfect initialization, the estimated plane is further refined iteratively. 
Points are re-selected based on the point-to-plane distance:
\begin{equation}
\mathcal{G}_{u,v}^{(t+1)} =
\left\{
{}^{W}\mathbf{p}_j \in \mathcal{P}_{u,v}
\;\middle|\;
\left|
\left(\mathbf{n}_{u,v}^{(t)}\right)^{\top}{}^{W}\mathbf{p}_j + d_{u,v}^{(t)} \right| < \tau_d
\right\}.
\end{equation}

This process progressively removes outliers and stabilizes the plane estimate. 
The final ground set $\mathcal{G}_{u,v}^{(T)}$ and plane parameters are used in the subsequent stage.

\subsubsection{Multi-criteria Ground Candidate Evaluation}

Although a local plane can be estimated for each grid cell, the resulting plane does not necessarily correspond to the true ground surface. 
In complex agricultural environments, locally planar non-ground structures, vegetation, and multi-path effects may lead to false ground hypotheses. 
We therefore introduce a multi-criteria evaluation strategy to determine whether a cell should be accepted as a valid ground candidate.

Following the general idea of progressive ground validation used in Patchwork~\cite{lim2021patchwork} and Patchwork++~\cite{lee2022patchwork++}, the proposed evaluation is organized as a staged filtering process, in which unreliable local plane hypotheses are rejected step by step. 
Specifically, the evaluation is based on three complementary criteria: geometric consistency, prior elevation consistency, and statistical stability.

\paragraph{Geometric Consistency (Uprightness)}
A valid ground surface is expected to be approximately horizontal in the world frame. 
Let $\mathbf{e}_z=[0,0,1]^{\top}$ denote the vertical direction. 
The uprightness criterion is defined as
\begin{equation}
\Phi_{u}(u,v)=
\begin{cases}
1, & \text{if } \left| \left(\mathbf{n}_{u,v}^{(T)}\right)^{\top}\mathbf{e}_z \right| \ge \cos(\theta_{u}),\\
0, & \text{otherwise.}
\end{cases}
\end{equation}
This criterion suppresses strongly inclined structures that are unlikely to be ground.

\paragraph{Prior Elevation Consistency}
However, approximately horizontal non-ground regions may still satisfy the geometric constraint. 
To further eliminate such false positives, we enforce consistency with the historical terrain prior.

At the current frame $F$, we maintain a sliding window of valid ground patches from the previous $N_h$ frames.
For each historical frame $f$, let $\mathcal{B}^{(f)}$ be the set of valid ground patches, and let $\bar{z}_{m}^{(f)}$ denote the mean height of patch $m$. 
The historical height set is defined as
\begin{equation}
\mathcal{H} = \left\{
\bar{z}_{m}^{(f)}
\;\middle|\;
f = F-N_h,\ldots,F-1,\;
m \in \mathcal{B}^{(f)}
\right\}.
\end{equation}
At the initial takeoff stage, the historical set $\mathcal{H}$ may be unavailable or insufficient because only a limited number of previous frames have been accumulated. 
Unlike the prior-constrained seed initialization, which uses a zero-height terrain prior, the prior elevation consistency check is temporarily bypassed until sufficient historical observations are available.

The local sliding window historical terrain prior is computed as
\begin{equation}
z_{\mathrm{local}}^{\mathrm{hist}} = \mu_{\mathrm{hist}} + \theta_{\mathrm{hist}} \sigma_{\mathrm{hist}},
\end{equation}
where $\mu_{\mathrm{hist}}$ and $\sigma_{\mathrm{hist}}$ are the mean and standard deviation of $\mathcal{H}$, respectively, and $\theta_{\mathrm{hist}}$ is a tunable coefficient.

The mean height of the estimated ground set is
\begin{equation}
\bar{z}_{u,v}^{(T)}=
\frac{1}{|\mathcal{G}_{u,v}^{(T)}|}
\sum_{{}^{W}\mathbf{p}_j \in \mathcal{G}_{u,v}^{(T)}} z_j,
\end{equation}
and the consistency criterion is defined as
\begin{equation}
\Phi_{h}(u,v)=
\begin{cases}
1, & \text{if } \bar{z}_{u,v}^{(T)} < z_{\mathrm{local}}^{\mathrm{hist}},\\
0, & \text{otherwise.}
\end{cases}
\end{equation}

\paragraph{Statistical Stability (Height Dispersion)}
Even when both orientation and elevation are plausible, the estimated ground set may still include low-lying vegetation or clutter. 
Such cases typically exhibit higher variation along the vertical direction. 
We therefore introduce the height standard deviation:
\begin{equation}
\sigma_{u,v}^{(T)}=
\sqrt{
\frac{1}{|\mathcal{G}_{u,v}^{(T)}|}
\sum_{{}^{W}\mathbf{p}_j \in \mathcal{G}_{u,v}^{(T)}}
\left(z_j-\bar{z}_{u,v}^{(T)}\right)^2
},
\end{equation}
and define the stability criterion as
\begin{equation}
\Phi_{s}(u,v)=
\begin{cases}
1, & \text{if } \sigma_{u,v}^{(T)} < \tau_s,\\
0, & \text{otherwise.}
\end{cases}
\end{equation}

The final decision is obtained by jointly enforcing all three criteria:
\begin{equation}
\Phi(u,v)=\Phi_u(u,v)\cdot \Phi_h(u,v)\cdot \Phi_s(u,v).
\end{equation}
A cell is accepted as a ground candidate if $\Phi(u,v)=1$. 
The resulting candidates are passed to the subsequent refinement stage.

\subsubsection{Ground Segmentation Refinement}

While the multi-criteria evaluation ensures high precision in ground candidate selection, it may also lead to missed ground points. 
Such false negatives typically arise from local multi-slope terrain structures within a cell or from complete cell rejection due to unreliable plane estimation. 
To improve segmentation completeness, we introduce two complementary refinement strategies: local re-segmentation and global recall.

\paragraph{Local Refinement (Re-segmentation)}
The initial plane estimation assumes a single dominant ground surface within each grid cell. 
However, real terrain may contain multiple locally connected slope components, causing part of the true ground points to be excluded.

For an accepted cell $(u,v)$, let
\begin{equation}
\mathcal{N}_{u,v}^{(T)} = \mathcal{P}_{u,v} \setminus \mathcal{G}_{u,v}^{(T)}
\end{equation}
denote the residual non-ground set. 
Its mean height is
\begin{equation}
\bar{z}_{u,v}^{\mathcal{N}}=
\frac{1}{|\mathcal{N}_{u,v}^{(T)}|}
\sum_{{}^{W}\mathbf{p}_j \in \mathcal{N}_{u,v}^{(T)}} z_j.
\end{equation}

Re-segmentation is triggered when
\begin{equation}
\left|\bar{z}_{u,v}^{\,\mathcal{N}} - \bar{z}_{u,v}^{(T)}\right| < \delta_z,
\qquad
|\mathcal{N}_{u,v}^{(T)}| > N_{\mathrm{re}},
\end{equation}
which indicates that the residual points are likely to belong to an alternative ground surface with similar elevation. 

Under this condition, the plane estimation and candidate evaluation procedure is re-applied to $\mathcal{N}_{u,v}^{(T)}$, yielding an additional ground subset $\Delta\mathcal{G}_{u,v}^{\mathrm{re}}$, which is merged into the current ground set.

\paragraph{Global Refinement (Global Recall)}
Some grid cells may be entirely rejected due to local fitting failure or strong outlier interference. 
To recover such cases, we perform a global recall based on the online terrain model constructed from successfully accepted ground candidates.
Rather than relaxing the first-pass thresholds, which may introduce false positives from vegetation, water surfaces, and other non-ground structures, global recall selectively recovers missed ground points using the accumulated terrain prior.

For a rejected cell $(u,v)$ with $\Phi(u,v)=0$, let $z_{u,v}^{\mathrm{hist}}$ denote the terrain height from the current terrain model. 
If no valid terrain height is available at the queried location, the cell is excluded from global refinement and remains rejected. 
Otherwise, a point ${}^{W}\mathbf{p}_j=[x_j,y_j,z_j]^{\top}\in\mathcal{P}_{u,v}$ is recalled as ground if
\begin{equation}
z_j \le z_{u,v}^{\mathrm{hist}} + \delta_h.
\end{equation}
The recalled points form the set $\Delta\mathcal{G}_{u,v}^{\mathrm{glb}}$.

\paragraph{Final Aggregation}
The final ground segmentation result is obtained by merging the initially accepted ground sets with the additional subsets recovered by local re-segmentation and global recall.

\subsection{Continuous Terrain Surface Modeling}

After ground segmentation and refinement, the extracted ground observations provide a reliable but still discrete representation of the terrain. 
However, terrain-following flight requires a continuous and queryable surface model. 
We therefore construct a continuous 2.5-D terrain representation through four steps: control-point generation, surface modeling, online querying, and incremental update.

\paragraph{Control-point Generation}
The grid structure established during ground segmentation is reused to define terrain control points, avoiding additional discretization. 
Let $\mathcal{G}_{u,v}^{\mathrm{ref}}$ denote the refined ground set of cell $(u,v)$. 
A control point is placed at the grid center:
\begin{equation}
x_{u,v}=\left(u+\frac{1}{2}\right)s,\qquad
y_{u,v}=\left(v+\frac{1}{2}\right)s.
\end{equation}

To obtain a robust height estimate under residual noise and outliers, the representative terrain height is selected as a lower quantile of the sorted height values:
\begin{equation}
h_{u,v}=z_{\left(\lfloor \rho |\mathcal{G}_{u,v}^{\mathrm{ref}}| \rfloor \right)},
\end{equation}
which suppresses elevated outliers while preserving terrain continuity. 
The resulting terrain representation is stored as a 2D grid, with each valid cell containing a height value $h_{u,v}$.
The resulting control point is
\begin{equation}
\mathbf{c}_{u,v}=\left(x_{u,v},\,y_{u,v},\,h_{u,v}\right)^{\top}.
\end{equation}

\paragraph{Continuous Surface Modeling}
Based on the structured control points, we construct a tensor-product B-spline surface:
\begin{equation}
z(x,y)=\sum_{u=0}^{N_x-1}\sum_{v=0}^{N_y-1} B_{u,k_x}(x)\,B_{v,k_y}(y)\,h_{u,v}.
\end{equation}
This piecewise-polynomial formulation provides smoothness together with local support, making it more suitable for spatially varying terrain reconstruction under sparse observations.

\paragraph{Online Terrain Query}
For any horizontal location $(x,y)$ in the world frame, the terrain height is obtained by evaluating the surface:
\begin{equation}
z_{\mathrm{terr}}=z(x,y).
\end{equation}
This allows efficient online querying for terrain-following planning and control.

\paragraph{Incremental Update}
As new observations arrive, control points are updated incrementally. 
Direct replacement may introduce fluctuations due to measurement noise. 
To ensure temporal stability, we adopt a conservative update rule:
\begin{equation}
\label{eq:inre_update}
h_{u,v}\leftarrow
\begin{cases}
h_{u,v}^{(t)}, & \text{if no historical value exists},\\
h_{u,v}^{(t)}, & \text{if } \left|h_{u,v}^{(t)}-h_{u,v}^{(t-1)}\right|>\tau_c,\\
h_{u,v}^{(t-1)}, & \text{otherwise.}
\end{cases}
\end{equation}
This strategy updates the terrain model only when significant changes are observed, with the absolute-difference criterion enabling symmetric updates for both increasing and decreasing terrain heights, reducing noise-induced jitter.

The resulting terrain surface supports continuous representation, efficient querying, and stable incremental refinement, providing a reliable reference for terrain-following flight.

\subsection{UAV System Integration and Deployment}

\begin{figure}[t]
    \centering
    \includegraphics[width=0.49\textwidth]{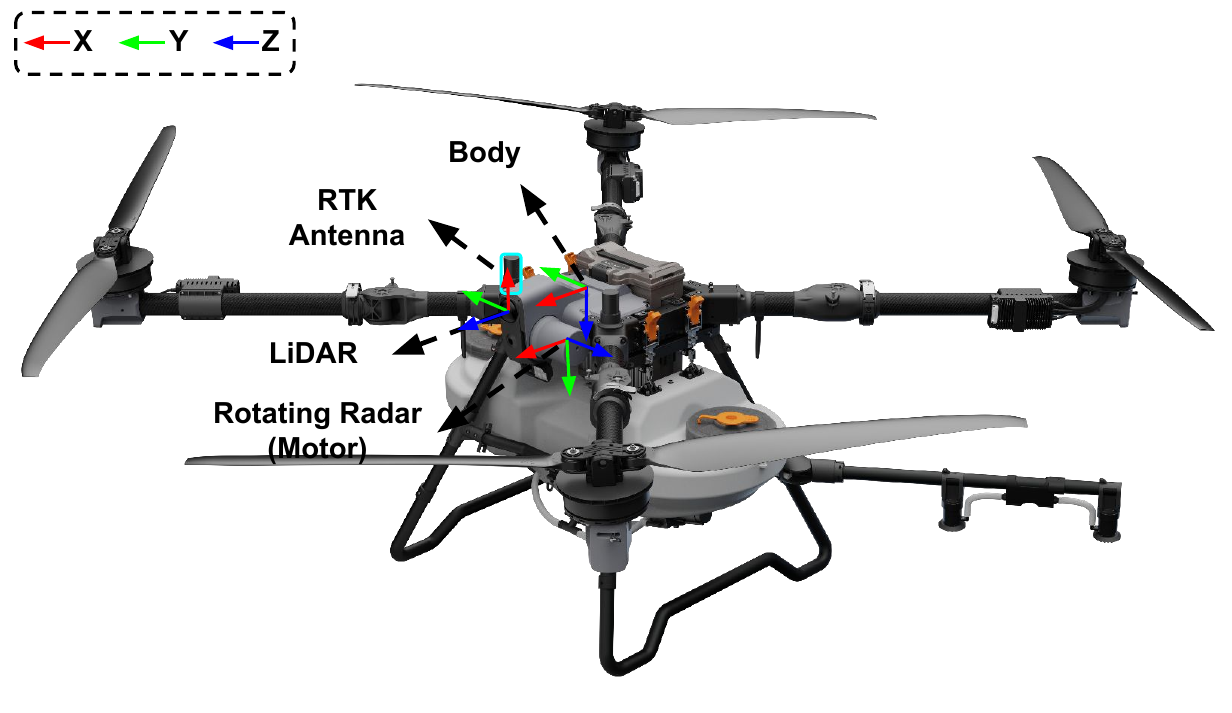}
    \caption{\textbf{Agricultural UAV platform equipped with the rotating radar sensing unit.} 
    The body frame is defined to coincide with the onboard state-estimator frame, and the rotating radar frame is aligned with the built-in motor frame.}

    \vspace{-0.3cm}
    \label{fig:uav_depoly}
\end{figure}

As shown in Fig.~\ref{fig:uav_depoly}, the proposed terrain perception and modeling pipeline is deployed as an onboard module on an agricultural UAV platform equipped with the rotating mmWave radar developed in this work. 
The system integrates radar sensing and state estimation to enable real-time terrain perception in the world frame. 
Specifically, the radar continuously acquires terrain observations, while the onboard state estimation system, implemented as an inertial navigation system / real-time kinematic (INS/RTK) module, provides UAV pose through multi-sensor fusion of inertial measurement unit (IMU) and RTK measurements.
The INS/RTK module achieves a position accuracy better than 2~cm + 1~ppm (50\% circular error probable, CEP) and an orientation accuracy of $0.1^\circ (1\sigma)$.
This allows all radar observations to be consistently transformed, accumulated, and processed for online terrain understanding.

The perception pipeline, including ground segmentation, refinement, and continuous terrain modeling, runs onboard an ARM-based embedded platform with an RK3588 processor. 
All modules are integrated within a ROS framework, enabling real-time data exchange among radar sensing, state estimation, terrain modeling, and flight control interfaces.

At each update cycle, a query point associated with the current UAV state is projected onto the continuous terrain surface model to obtain the terrain height:
\begin{equation}
z_{\mathrm{terr}} = z(x_q,y_q),
\end{equation}
where $(x_q,y_q)$ denotes the horizontal query location. 
Given a desired above-ground flight height $h_{\mathrm{ref}}$, the terrain-referenced altitude command is computed as
\begin{equation}
z_{\mathrm{cmd}} = z_{\mathrm{terr}} + h_{\mathrm{ref}}.
\end{equation}

This terrain-referenced altitude serves as an external input to the flight control system, enabling the UAV to maintain a consistent height relative to the terrain. 
As new observations are incorporated, the terrain model is incrementally updated, ensuring temporal consistency under dynamic sensing conditions.

Overall, the proposed integration forms a closed-loop onboard system that connects radar sensing, terrain perception, continuous modeling, and control input generation, enabling reliable terrain-following flight in complex agricultural environments.

\section{EXPERIMENTS AND EVALUATION}
\label{sec:exp}

\subsection{Experimental Setup}
\subsubsection{Data Collection}
To evaluate the proposed system under realistic agricultural conditions, field flight experiments were conducted across multiple representative terrain scenarios. 
As illustrated in Fig.~\ref{fig:teaser}, the experimental dataset encompasses four typical terrain types: flat field, slope, water, and tea plantation hill. 
These scenarios present distinct terrain characteristics and perception challenges, including relatively smooth ground, locally varying slopes, strong reflection and multi-path interference, and vegetation occlusion.

To facilitate a comprehensive evaluation of altitude effects, data were collected for each terrain scenario at three distinct flight altitudes while maintaining a similar traversal speed range throughout all flights. 
For each scenario, three flights were conducted at different altitudes, resulting in 12 flights in total. 
The UAV traversal speed was maintained within 5--8~m/s, and the total flight duration was 30~min~57~s.
Specifically, the flight altitudes were set to 3, 5, and 8~m for flat field; 5, 8, and 10~m for slope; 3, 6, and 10~m for water; and 18, 20, and 22~m for tea plantation hill. 
These altitude configurations were carefully selected to balance point cloud density, spatial coverage, and operational safety under varying environmental and terrain conditions.

\subsubsection{Ground Truth Annotation}
To quantitatively evaluate both ground segmentation and terrain modeling performance, reference data were prepared for the two tasks.

\paragraph{Ground Segmentation}  


To generate reference labels for radar ground segmentation, we first constructed a LiDAR-based terrain reference map for each experimental area. 
The test area was scanned using the onboard LiDAR sensor, and the corresponding poses were provided by the high-precision INS/RTK module. 
These INS/RTK poses were used as initialization for generalized iterative closest point (GICP)-based scan registration, yielding a spatially consistent LiDAR reference map. 
The reconstructed LiDAR map was then partitioned into local regions and segmented using Patchwork++~\cite{lee2022patchwork++} to obtain initial terrain points. 
Because automatic LiDAR ground segmentation may still produce residual errors in vegetation, sparse-return, or reflective regions, the segmented terrain map was further manually inspected and corrected to remove obvious outliers and recover missed ground regions. 
Finally, radar points from the evaluation flights were transformed into the LiDAR reference-map frame using the INS/RTK poses and the CAD-calibrated LiDAR-radar extrinsic parameters. 
Each radar point was assigned a ground or non-ground label according to its vertical distance to the corrected LiDAR-derived terrain surface, using predefined lower and upper height tolerance thresholds.

We also note that the resulting reference annotations are subject to several sources of uncertainty. 
First, the LiDAR reference map may contain small registration errors after RTK-assisted ICP alignment. 
Second, although Patchwork++ provides an initial ground segmentation and we manually inspect and correct it for large errors, mis-segmentations may still be present in regions with vegetation, sparse returns, or strong reflection. 
Third, the transfer of LiDAR-derived ground labels to radar points depends on the accuracy of the LiDAR-radar extrinsic calibration and the UAV pose estimates. 
For water-covered areas, where LiDAR returns are sparse and partially specular, we assume the local water surface to be approximately planar and assign radar ground labels using a height-tolerance band relative to the calibrated reference surface rather than strict point-to-point correspondence.

\paragraph{Terrain Modeling}  
Ground truth terrain data were captured through on-site measurements. 
Representative sampling points were distributed across each experimental area to capture varying terrain conditions. 
Horizontal coordinates were recorded using the UAV remote controller’s waypoint marking function, while precise elevation data were acquired via the RTK positioning system. 
For water-covered areas, where direct sampling was not feasible, the water surface was assumed to have no significant elevation variation and was therefore modeled as a planar surface based on elevation measurements collected along its boundaries. 
These discrete ground-truth terrain points served as benchmark reference values for quantitative assessment of terrain modeling accuracy.

\subsubsection{Parameter Settings}

All parameters were fixed for all experiments without scenario-specific tuning.
For point cloud preprocessing, the temporal aggregation window and grid size were set to $K=3$ and $s=5.0$~m, respectively.
These settings provide sufficient point density for sparse mmWave radar measurements while preserving real-time efficiency.
For region-wise ground plane estimation, the minimum number of points in each grid cell was set to $N_{\min}=10$ to avoid unreliable plane fitting in sparsely populated regions.
For prior-constrained seed initialization, the lower and upper elevation bounds were set to $\delta_{lower}=\delta_{upper}=2.0$~m.
In the iterative refinement step, the point-to-plane distance threshold was set to $\tau_d=0.8$~m.
These elevation- and distance-related thresholds were selected according to the measurement uncertainty of the radar sensor and the expected local terrain variations in agricultural fields.

For geometric consistency verification, the maximum allowable slope angle was set to $\theta_u=\pi/4$, which constrains the extracted ground regions to physically reasonable terrain surfaces.
For prior elevation consistency, the temporal window size was set to $N_h=3$, and the coefficient of the elevation threshold was set to $\theta_{\mathrm{hist}}=3$.
These settings provide a balance between temporal stability and adaptability while suppressing high-elevation outliers in the historical terrain prior.
For statistical stability verification, the height dispersion threshold was set to $\tau_s=2.0$~m to tolerate natural terrain variations while filtering unstable local regions.

In the local refinement stage, the elevation tolerance and minimum support size were set to $\delta_z=1.5$~m and $N_{re}=20$, respectively.
For global refinement, the elevation threshold was set to $\delta_h=0.8$~m.
These refinement parameters were chosen to recover near-ground false-negative points while limiting the introduction of additional false positives.
Finally, in terrain surface modeling, the update threshold and quantile parameter were set to $\tau_c=0.5$~m and $\rho=0.95$, respectively.
The same parameter set was used for flat fields, slopes, water-covered regions, and tea plantation hills, indicating that the proposed method is robust to different agricultural terrains.

\subsection{Effect of Rotating Radar Sensing}

\begin{figure*}[hbtp]
    \centering
    \subfloat[Fixed Sensing\label{fig:fixed_fov}]{
        \includegraphics[width=0.485\linewidth]{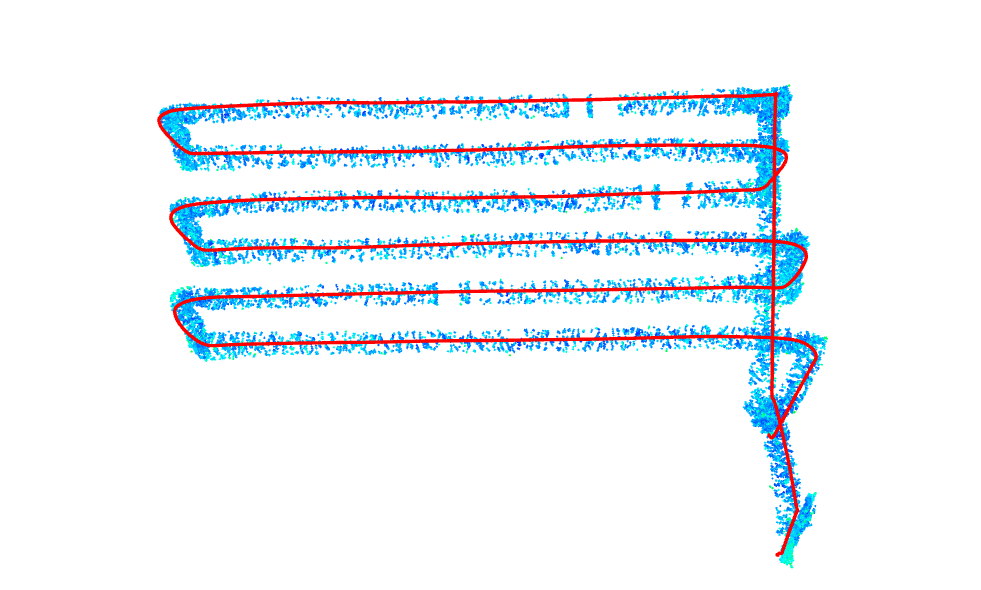}
    }
    \hfill
    \subfloat[Rotating Sensing\label{fig:rotating_fov}]{
        \includegraphics[width=0.485\linewidth]{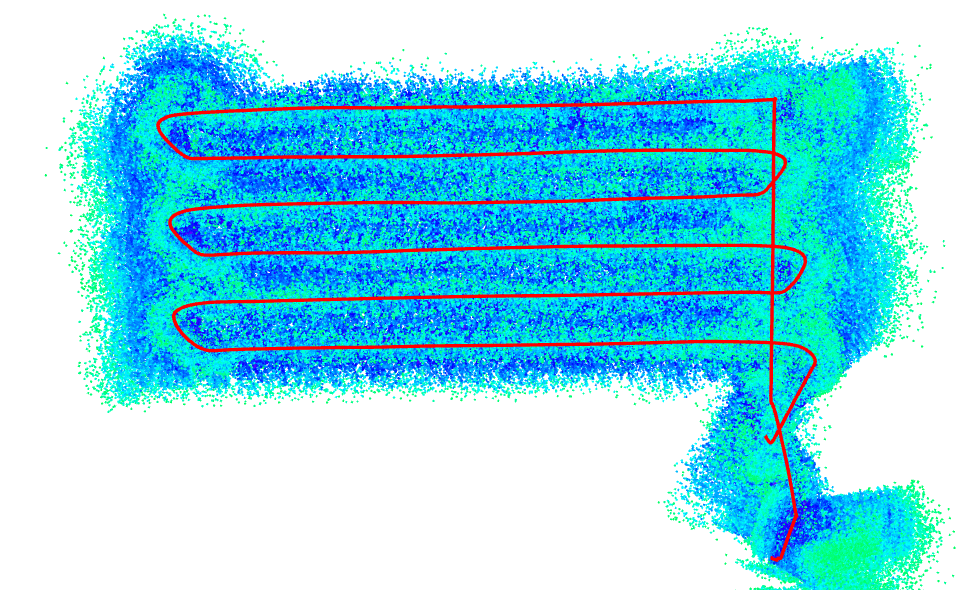}
    }

    \caption{\textbf{Comparison of BEV coverage under fixed and rotating radar sensing configurations.}
    Compared with the fixed configuration, the rotating radar continuously sweeps the sensing field over a wider angular range, thereby providing more complete and stable ground observations under varying UAV attitudes and terrain conditions.
    }

    \vspace{-0.3cm}
    \label{fig:bev_comparison}
\end{figure*}

To evaluate the effectiveness of the rotating radar configuration, we compare the ground observation coverage obtained by static and rotating radar sensing.
The same radar sensor is used in both settings, and the flight trajectory and operating parameters are kept identical.
For the static configuration, the rotation mechanism is disabled, and the radar boresight is oriented downward with an $8^\circ$ forward tilt from the nadir direction.
The accumulated valid ground radar returns along the flight trajectory are projected onto a BEV occupancy grid with a resolution of 0.5~m. 
A grid cell is considered covered if it contains at least one valid ground return, and the covered area is computed as the number of occupied cells multiplied by the cell area.

As shown in Fig.~\ref{fig:bev_comparison}, the static radar configuration covers only 24.19\% of the ground area observed by the rotating configuration.
The static setting also produces discontinuous observations along the flight trajectory, especially during UAV acceleration and deceleration.
In contrast, the rotating radar yields more complete and uniform ground coverage, providing more stable observations for terrain perception and terrain-following flight.

\subsection{Ground Segmentation Evaluation}
Due to limited onboard computational resources, we restrict comparisons to methods that are practical for deployment on our UAV hardware. Specifically, we benchmark against RANSAC-Single, RANSAC-Patch (both based on RANSAC~\cite{fischler1981random}), Patchwork~\cite{lim2021patchwork}, and Patchwork++~\cite{lee2022patchwork++}. These representative baselines are closely aligned with our problem setting and provide a meaningful evaluation of the proposed method under realistic onboard deployment constraints.

RANSAC-Single directly estimates a single plane from the point set within each local region, whereas RANSAC-Patch first partitions the point cloud into local patches and then performs patch-wise robust plane fitting for ground extraction. 
For RANSAC-Patch, the same grid resolution as the proposed method is adopted, and the number of iterations is set to $50$ for fair comparison. 
Since Patchwork and Patchwork++ rely on a concentric-zone partitioning strategy, the point cloud is transformed into a UAV-centered local frame before segmentation, and the resulting labels are then mapped back to the world frame using the corresponding UAV pose. 
In addition, all methods employ the same plane-distance threshold during ground extraction and are evaluated on the mmWave point cloud segmentation dataset constructed in this study.

\subsubsection{Quantitative Comparison}
Quantitative evaluation is conducted using precision (Prec.), Recall, Intersection-over-Union (IoU), and F1 score. 
Table~\ref{tab:overall} reports the overall comparison between the proposed method and the baseline approaches on the mmWave point cloud segmentation dataset constructed in this study.

\begin{table}[t]
\centering
\caption{\textbf{Overall ground segmentation results on the mmWave point cloud dataset.} 
All metrics are reported in percentage. \textbf{Bold} values indicate the best performance.}
\label{tab:overall}

\begin{tabular}{lcccc}
\toprule
\textbf{Method} & \textbf{Prec.} & \textbf{Recall} & \textbf{IoU} & \textbf{F1} \\
\midrule
RANSAC-Single & 99.26 & 80.47 & 80.00 & 88.89 \\
RANSAC-Patch  & 98.78 & 83.47 & 82.62 & 90.48 \\
Patchwork     & 97.70 & 64.79 & 63.81 & 77.91 \\
Patchwork++   & 97.12 & 47.83 & 47.16 & 64.10 \\
\textbf{Proposed} & \textbf{99.27} & \textbf{90.03} & \textbf{89.44} & \textbf{94.42} \\
\bottomrule
\end{tabular}

\end{table}

\begin{table*}[t]
\centering
\caption{\textbf{Scenario-wise comparison of ground segmentation performance in the flat field, slope, water, and tea plantation hill scenarios.} 
All metrics are reported in percentage. \textbf{Bold} values indicate the best performance.
}
\label{tab:scenes}

\renewcommand{\arraystretch}{1.15}
\setlength{\tabcolsep}{4pt}

\begin{tabular}{lcccc cccc cccc cccc}
\toprule

\multirow{2}{*}{\textbf{Method}} 
& \multicolumn{4}{c}{\textbf{Flat Field}} 
& \multicolumn{4}{c}{\textbf{Slope}} 
& \multicolumn{4}{c}{\textbf{Water}} 
& \multicolumn{4}{c}{\textbf{Tea Plantation Hill}} \\

\cmidrule(lr){2-5} 
\cmidrule(lr){6-9} 
\cmidrule(lr){10-13} 
\cmidrule(lr){14-17}

& \textbf{Prec.} & \textbf{Recall} & \textbf{IoU} & \textbf{F1}
& \textbf{Prec.} & \textbf{Recall} & \textbf{IoU} & \textbf{F1}
& \textbf{Prec.} & \textbf{Recall} & \textbf{IoU} & \textbf{F1}
& \textbf{Prec.} & \textbf{Recall} & \textbf{IoU} & \textbf{F1} \\

\midrule

RANSAC-Single 
& 99.82 & 87.93 & 87.79 & 93.50
& 99.01 & 81.82 & 81.16 & 89.60
& 99.83 & 84.76 & 84.65 & 91.68
& 98.84 & 70.22 & 69.65 & 82.11 \\

RANSAC-Patch  
& 99.33 & 89.32 & 88.79 & 94.06
& 98.84 & 85.24 & 84.39 & 91.54
& 99.05 & 73.31 & 72.80 & 84.26
& 97.98 & 75.84 & 74.68 & 85.50 \\

Patchwork     
& 98.62 & 70.13 & 69.45 & 81.97
& 96.86 & 70.95 & 69.36 & 81.91 
& 79.76 & 19.33 & 18.43 & 31.12
& 98.56 & 55.62 & 55.18 & 71.11 \\

Patchwork++   
& 97.95 & 46.27 & 45.82 & 62.85
& 95.89 & 53.55 & 52.34 & 68.72
& 94.03 & 27.17 & 26.71 & 42.16
& 98.28 & 44.34 & 44.00 & 61.11 \\

\textbf{Proposed} 
& \textbf{99.89} & \textbf{90.23} & \textbf{90.14} & \textbf{94.82}
& \textbf{99.21} & \textbf{91.70} & \textbf{91.03} & \textbf{95.31}
& \textbf{99.88} & \textbf{97.73} & \textbf{97.62} & \textbf{98.79}
& \textbf{98.88} & \textbf{86.95} & \textbf{86.10} & \textbf{92.53} \\

\bottomrule
\end{tabular}
\end{table*}

As shown in Table~\ref{tab:overall}, the proposed method achieves the best performance across all evaluation metrics. 
In particular, it achieves an IoU of 89.44\% and an F1-score of 94.42\%, both of which surpass those of all competing methods. These results indicate that the proposed method can more reliably extract ground points from sparse and noisy mmWave radar observations.

Table~\ref{tab:scenes} further presents the scenario-wise segmentation results using the same evaluation metrics for flat field, slope, water, and tea plantation hill. 
The proposed method consistently achieves the best performance across all scenarios, demonstrating strong robustness under varying terrain characteristics and sensing conditions.
The advantage is particularly evident in the tea plantation hill and water scenarios, where terrain variation, vegetation interference, and multi-path effects make ground extraction more challenging.

\subsubsection{Qualitative Comparison}
\begin{figure*}[htbp]
    \centering
    \setlength{\tabcolsep}{1.2pt}
    \renewcommand{\arraystretch}{1.0}

    \newcommand{\segimg}[2][0.13\textwidth]{%
        \includegraphics[width=#1,valign=m]{#2}%
    }

    \newcommand{\seglabel}[1]{%
        \adjustbox{valign=m}{%
            \rotatebox[origin=c]{90}{%
                \makebox[2.5cm][c]{\textbf{#1}}%
            }%
        }%
    }
    
    \begin{tabular}{c c c c c c c c}
        \toprule
        & \textbf{Scene}
        & \textbf{Ground-Truth}
        & \textbf{RANSAC-Single} 
        & \textbf{RANSAC-Patch} 
        & \textbf{Patchwork} 
        & \textbf{Patchwork++} 
        & \textbf{Proposed}  \\
        
        \midrule
        \seglabel{Flat Field}
        & \segimg{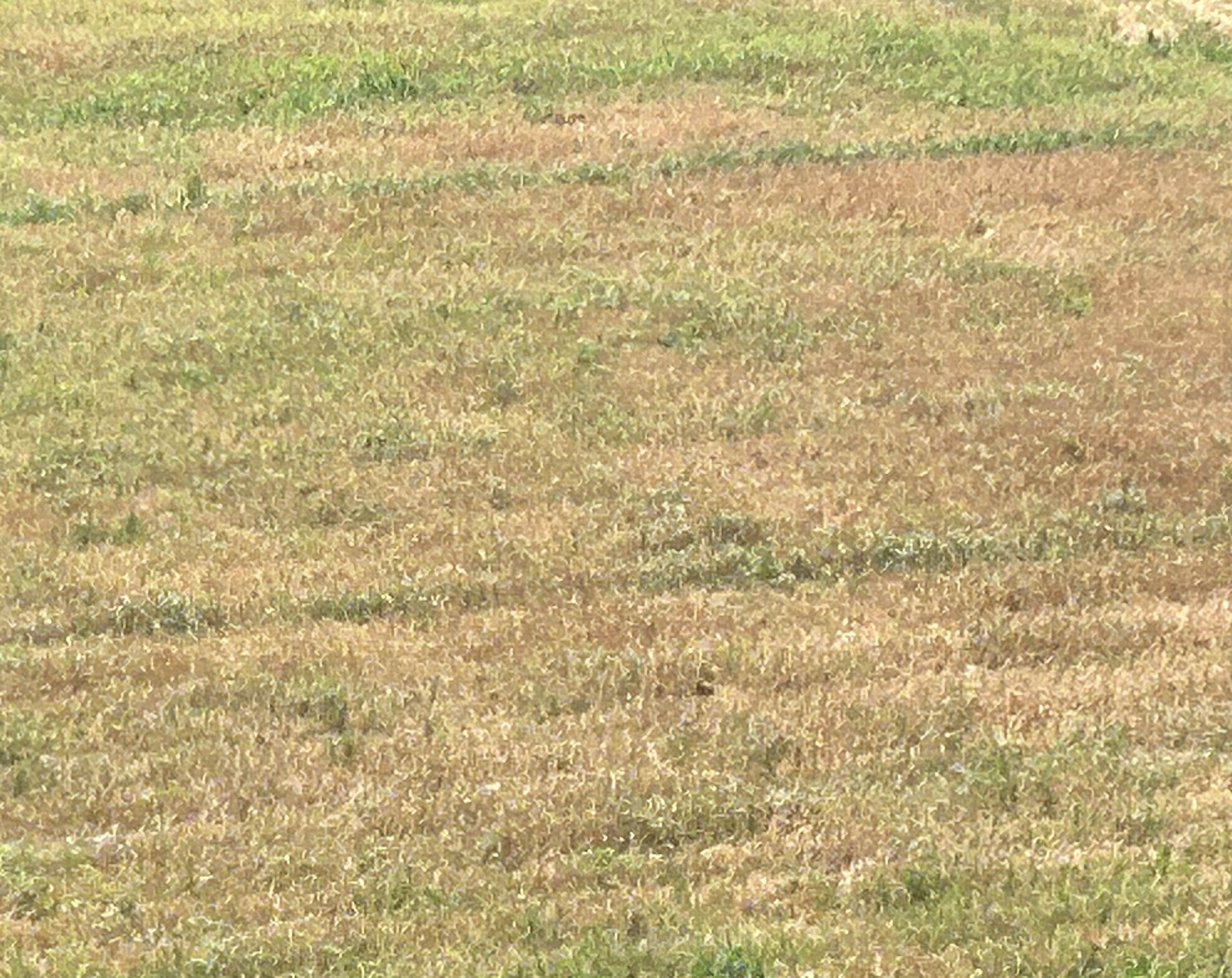}
        & \segimg{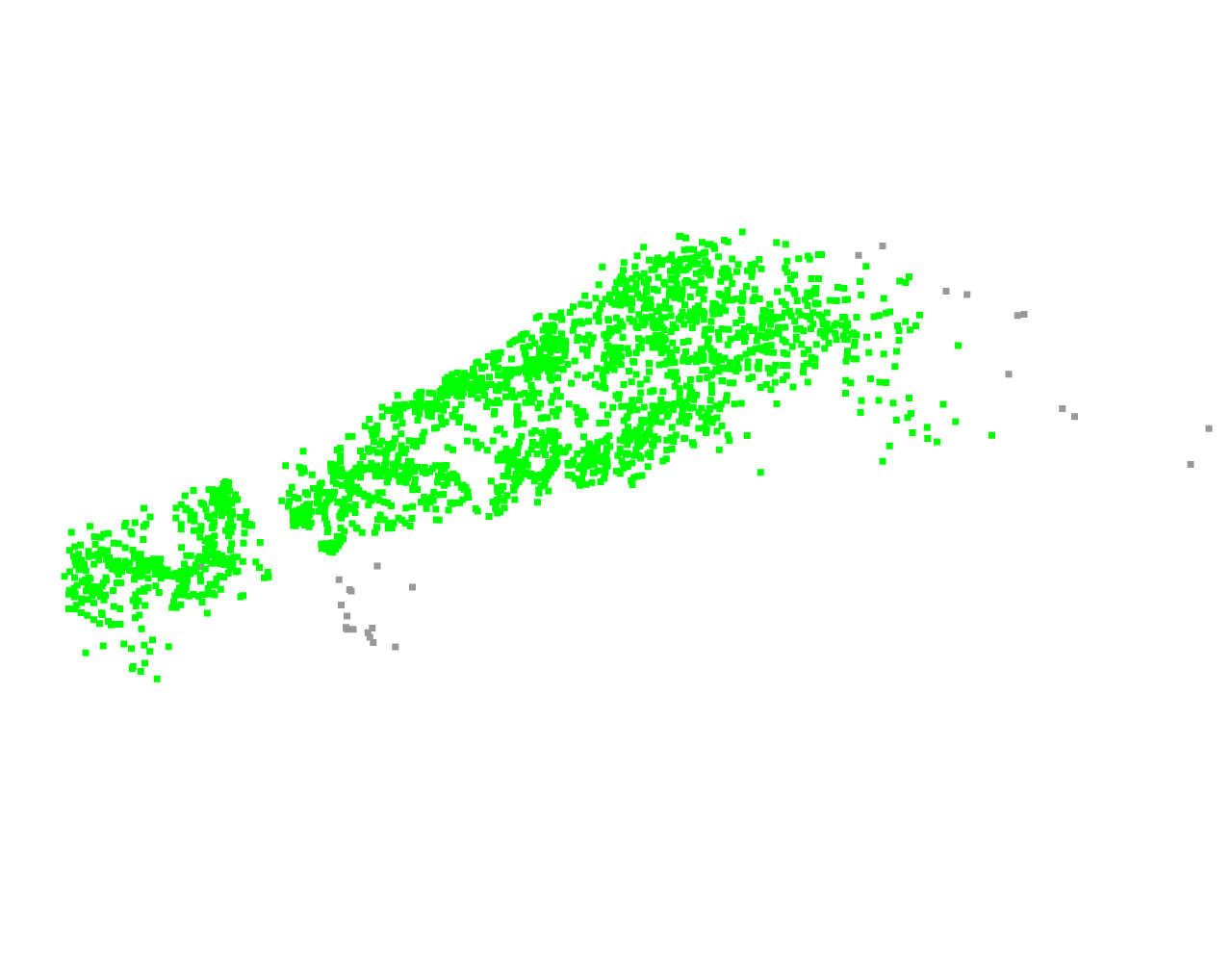}
        & \segimg{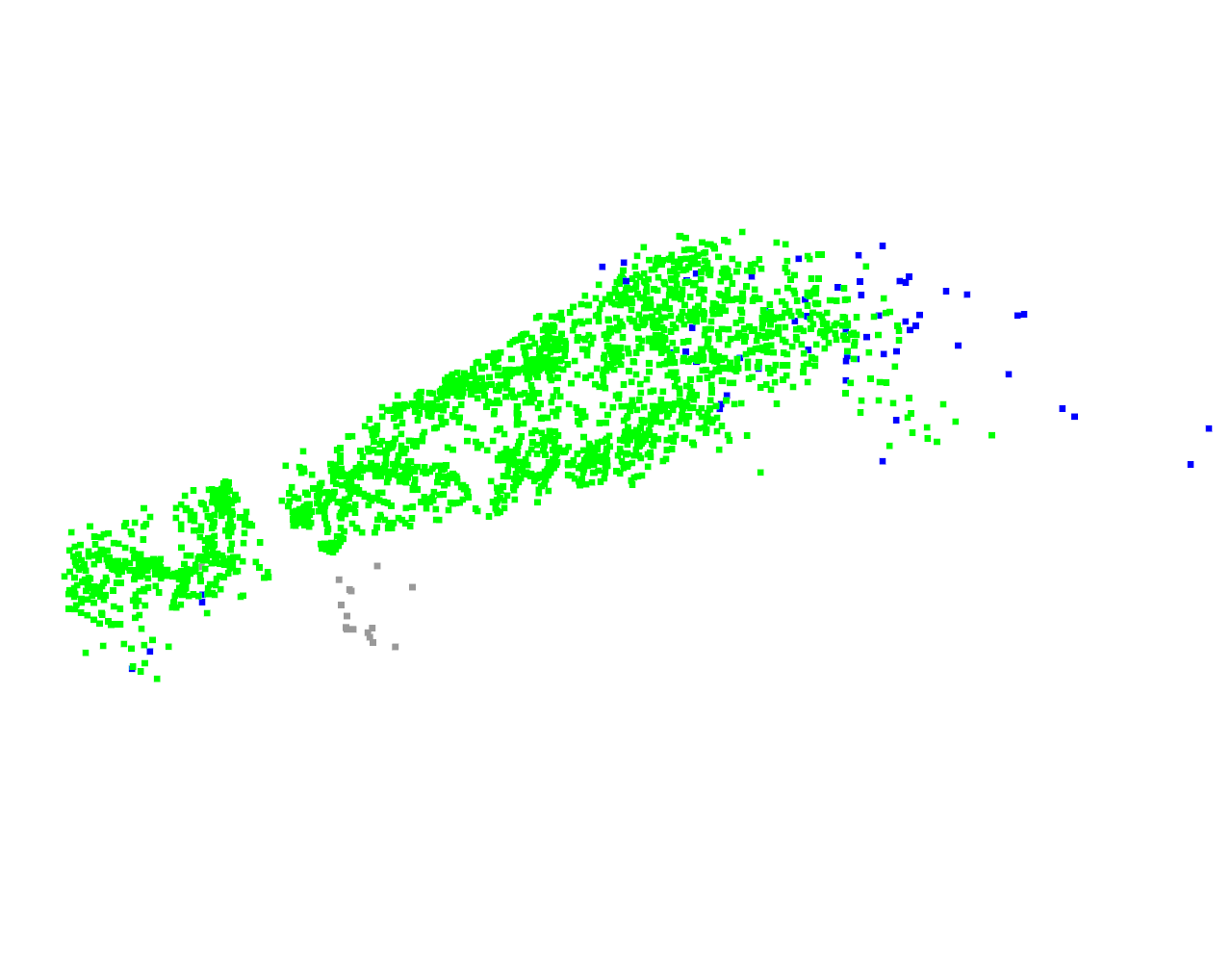}
        & \segimg{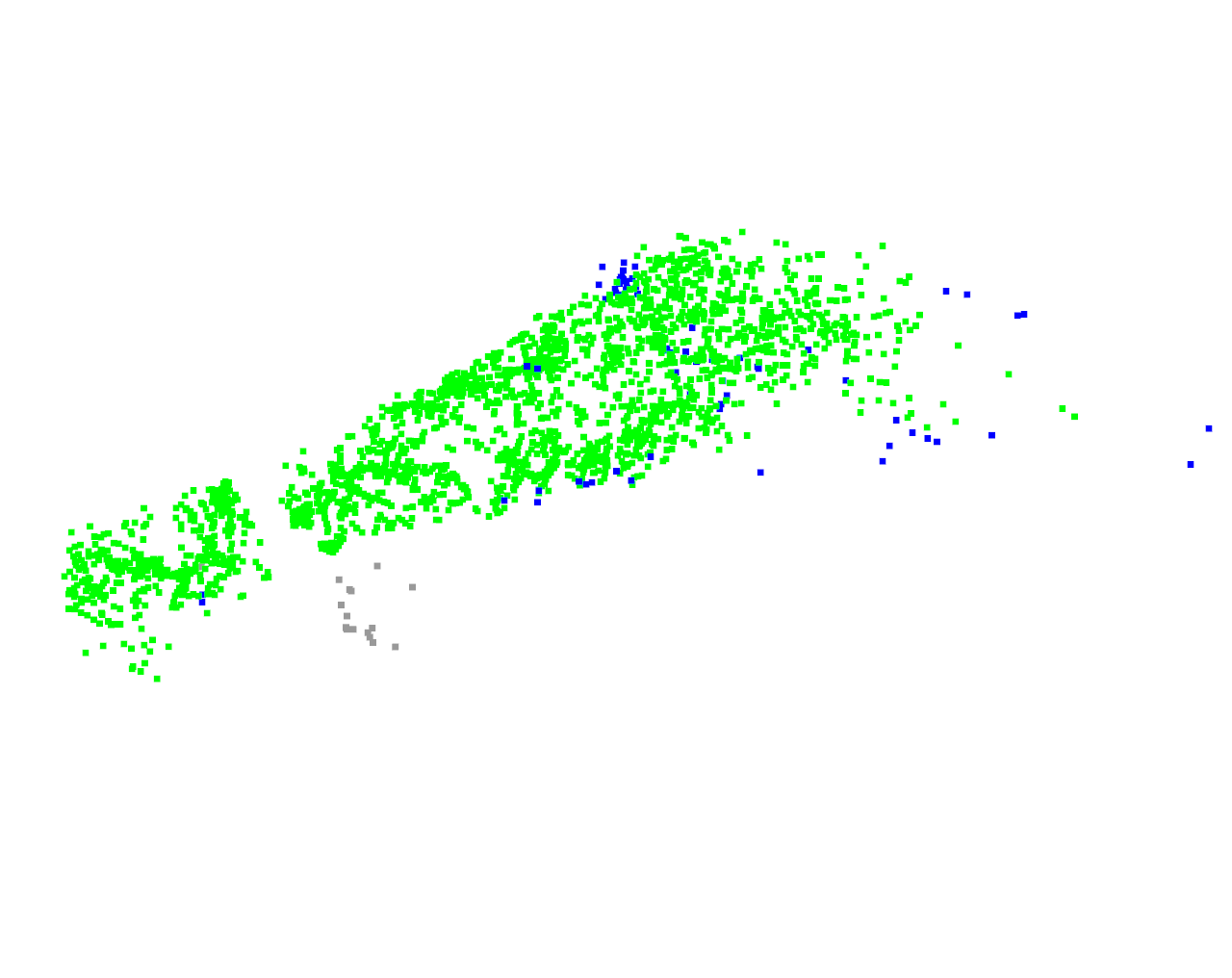}
        & \segimg{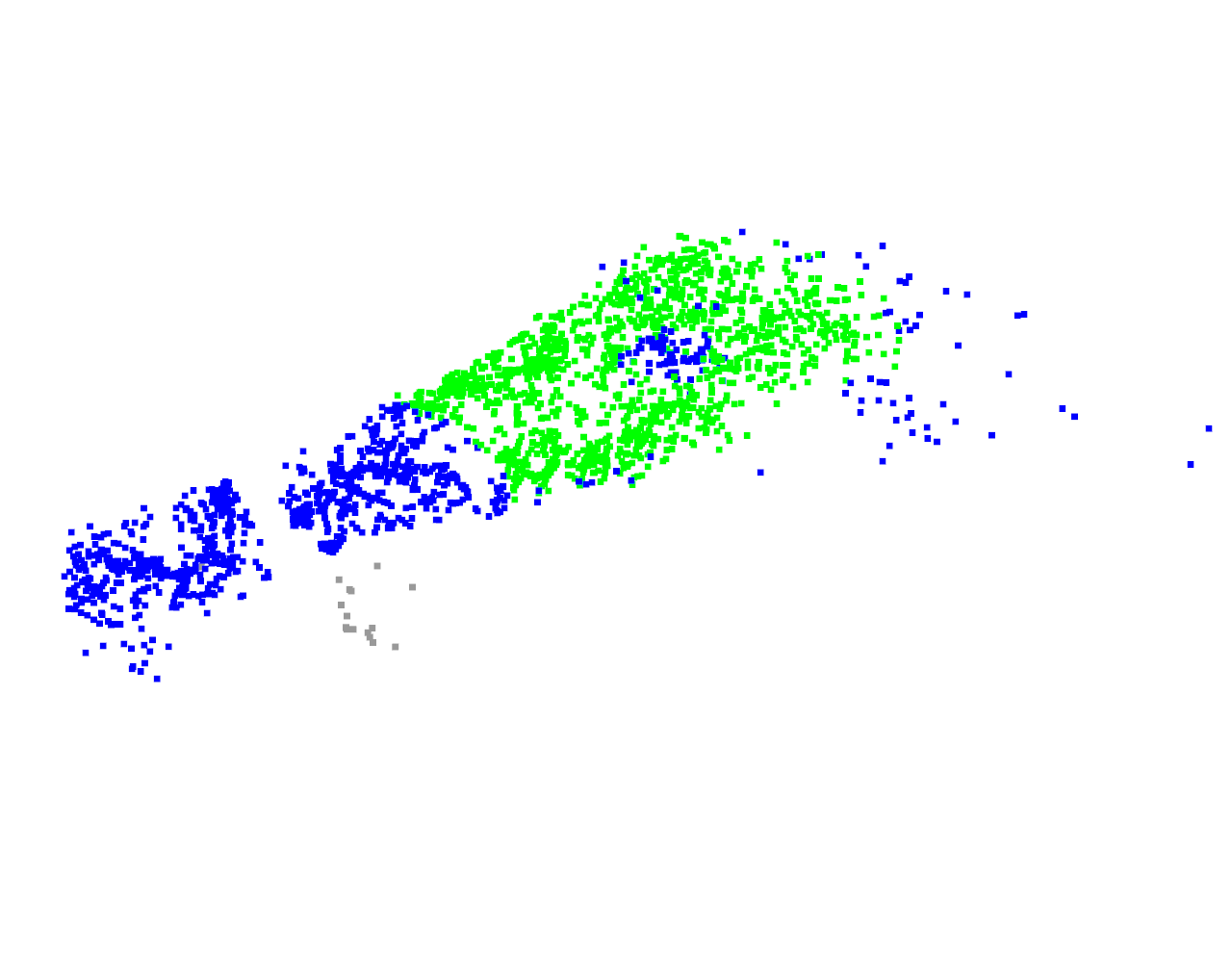}
        & \segimg{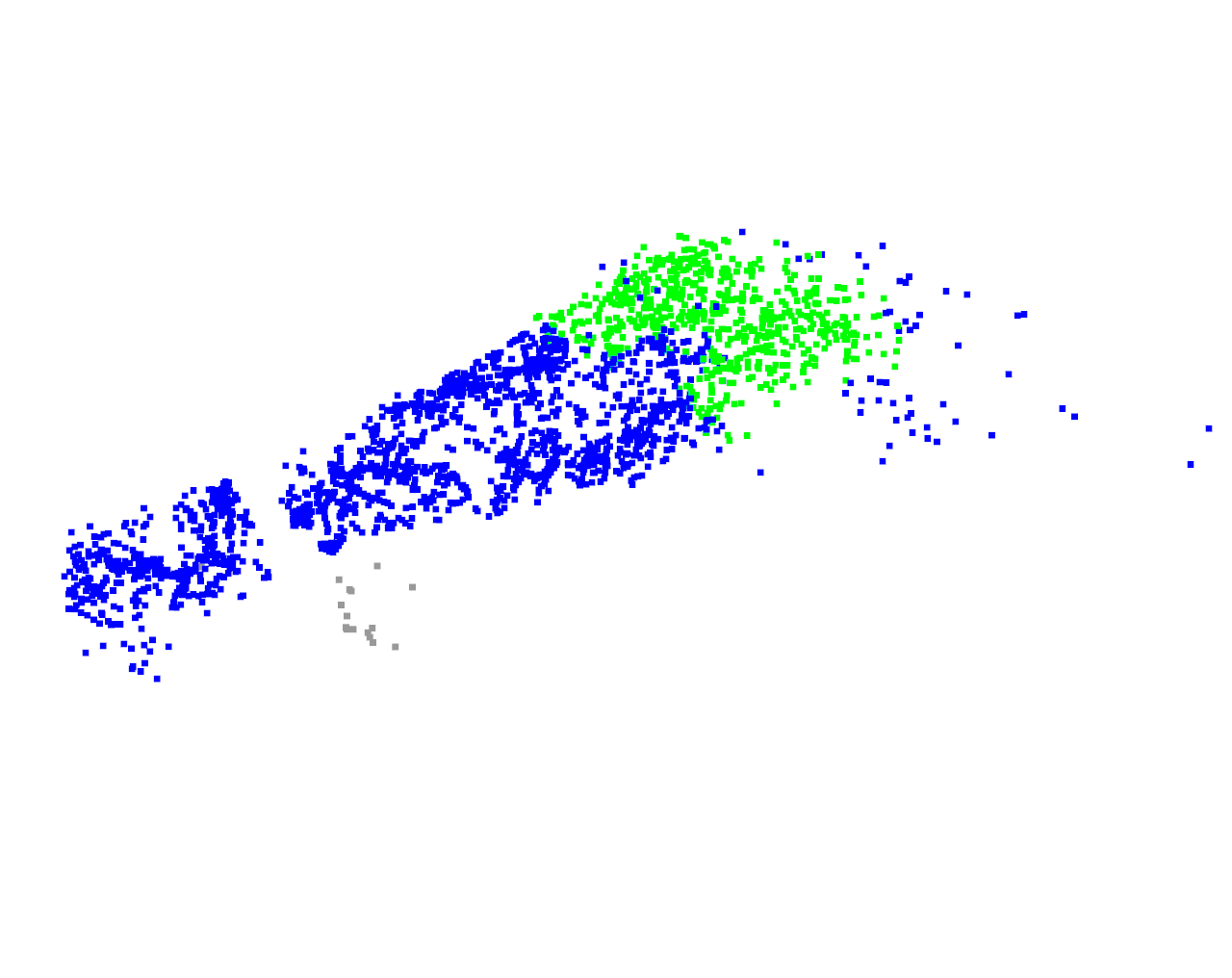}
        & \segimg{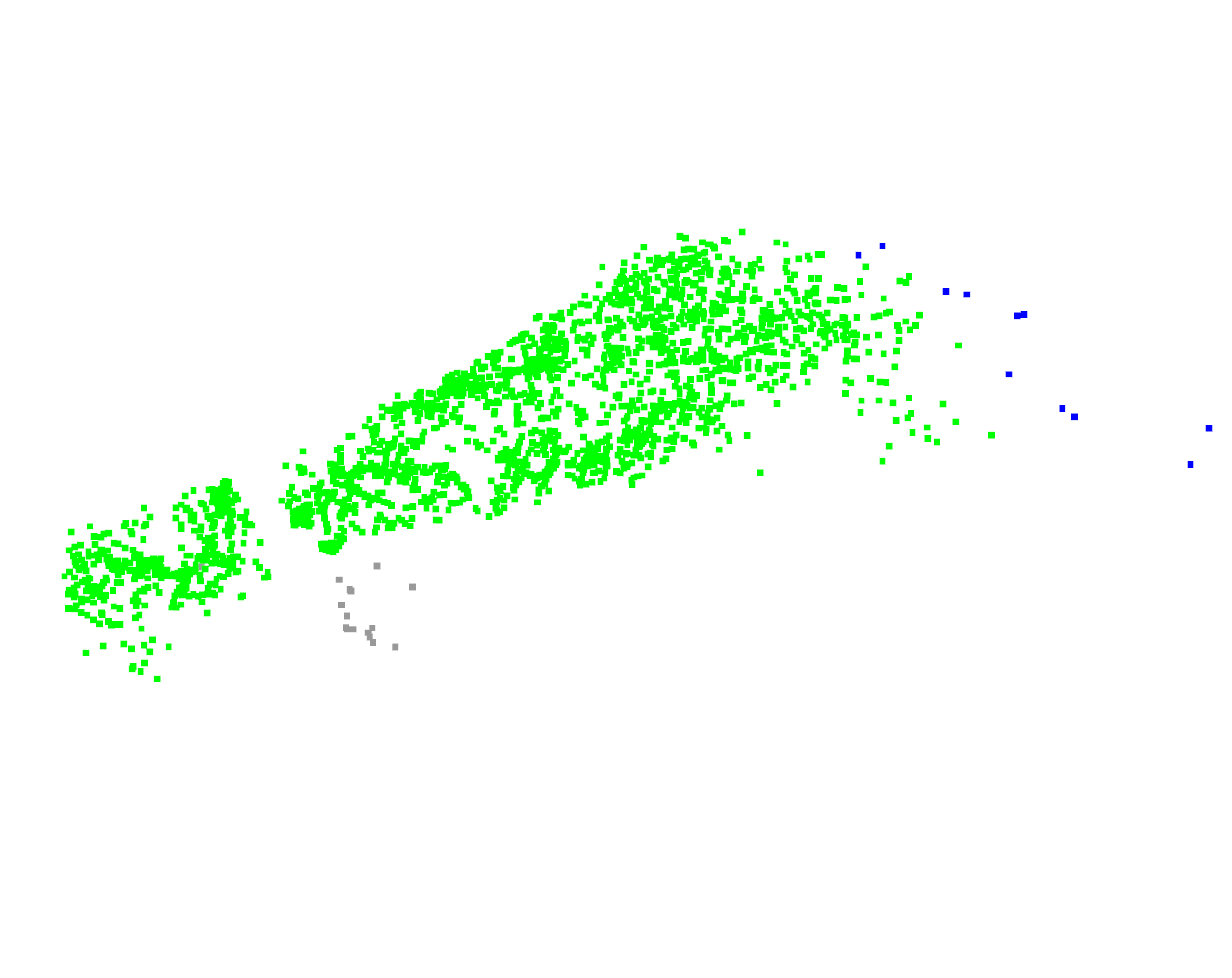} \\

        \seglabel{Slope}
        & \segimg{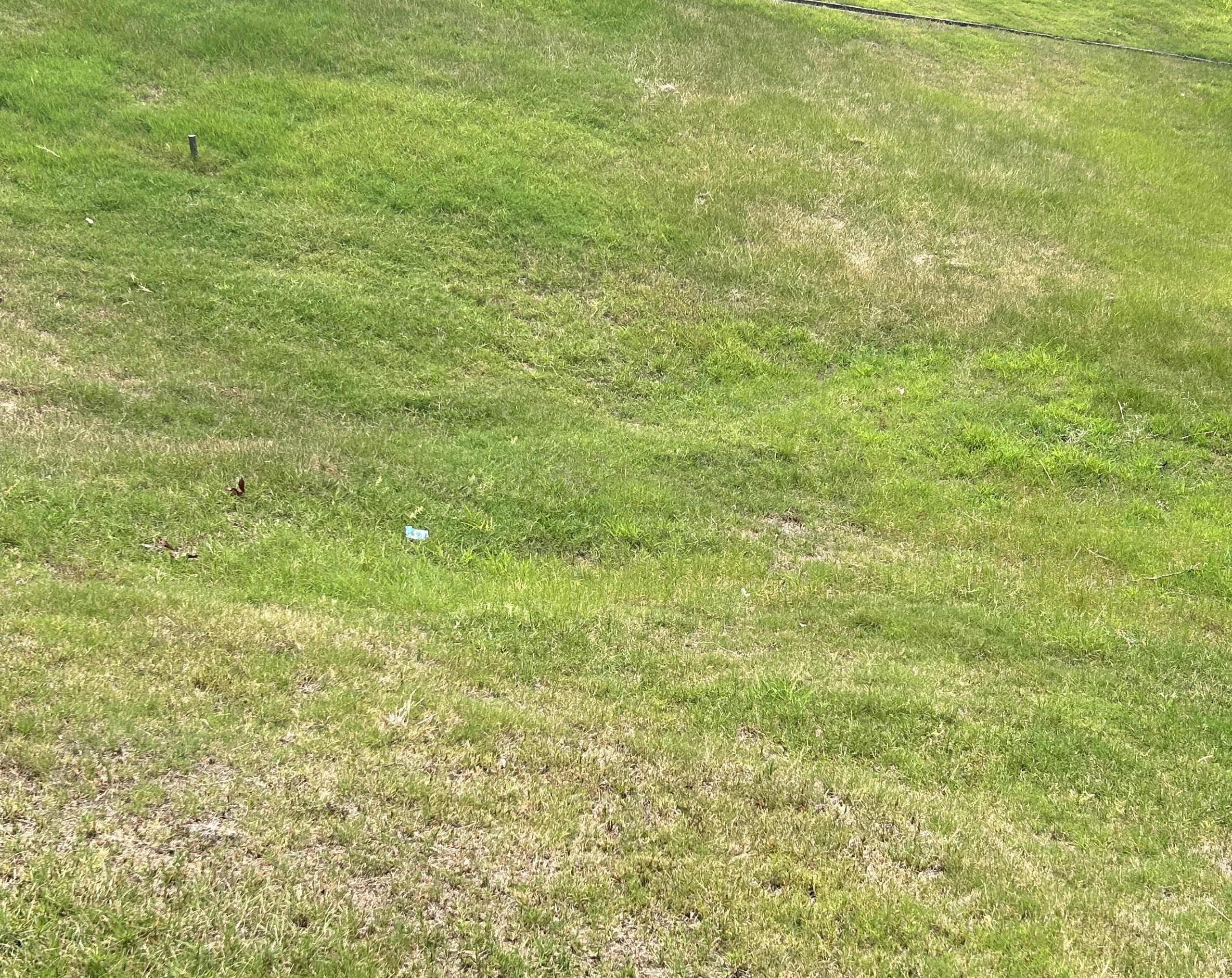}
        & \segimg{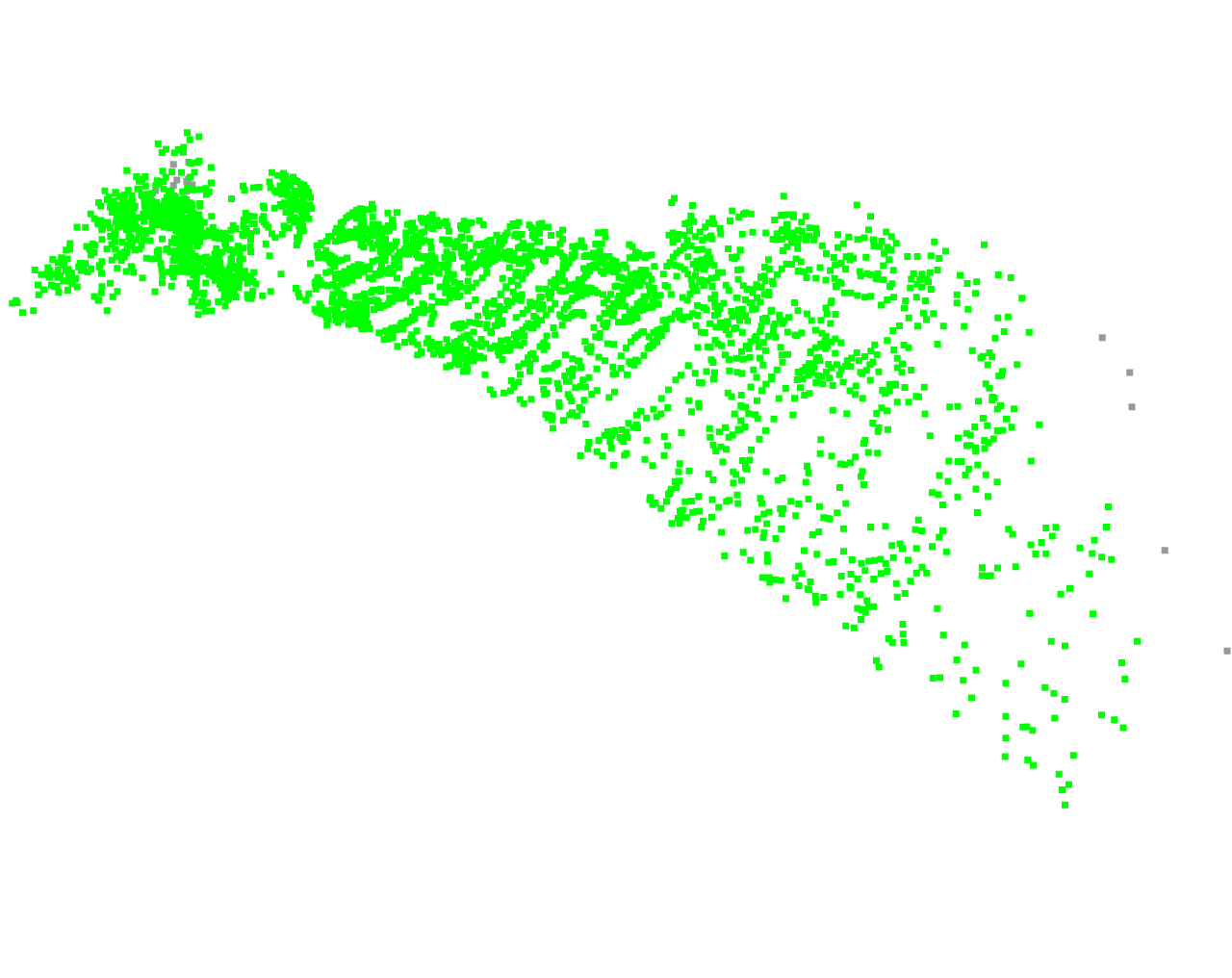}
        & \segimg{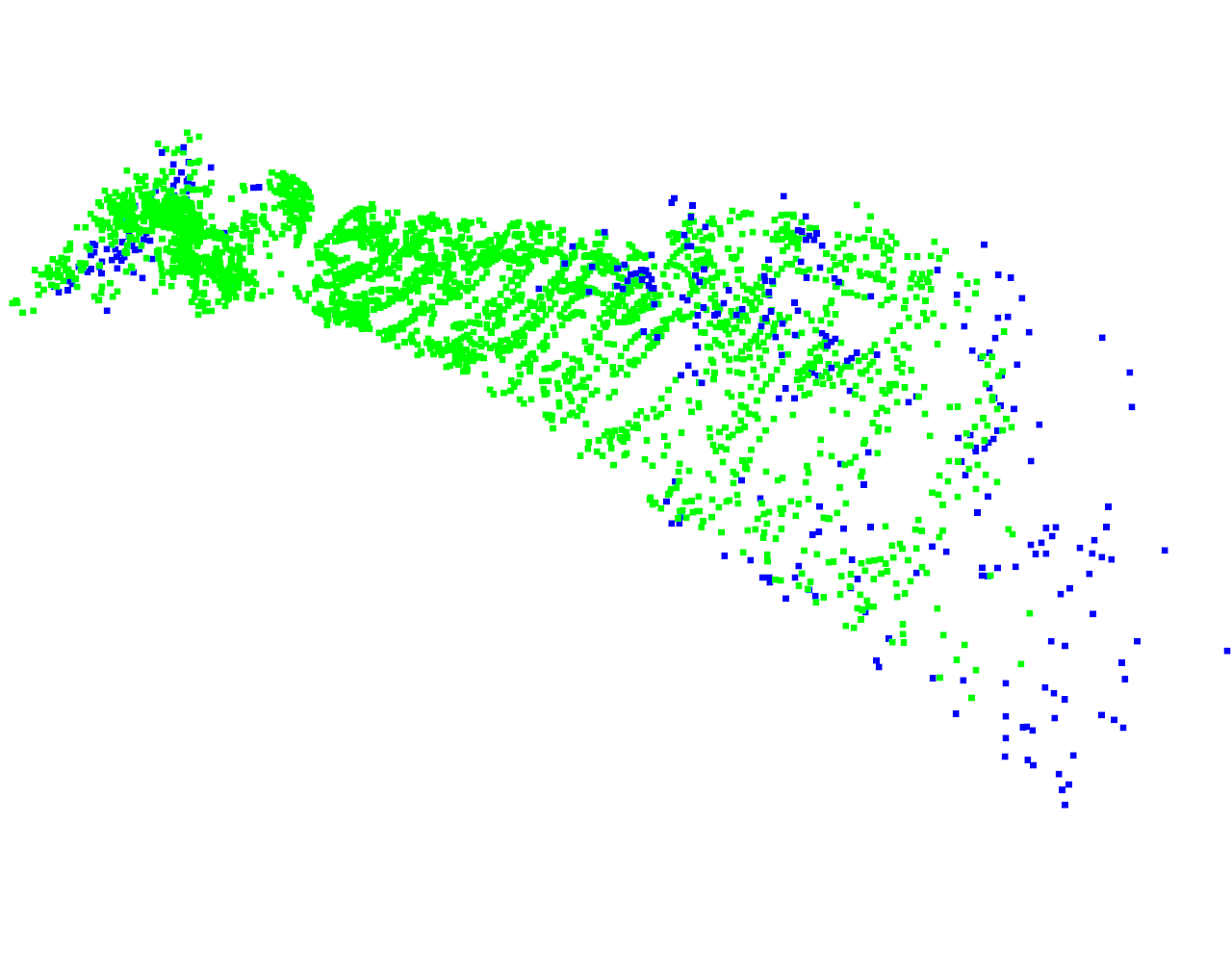}
        & \segimg{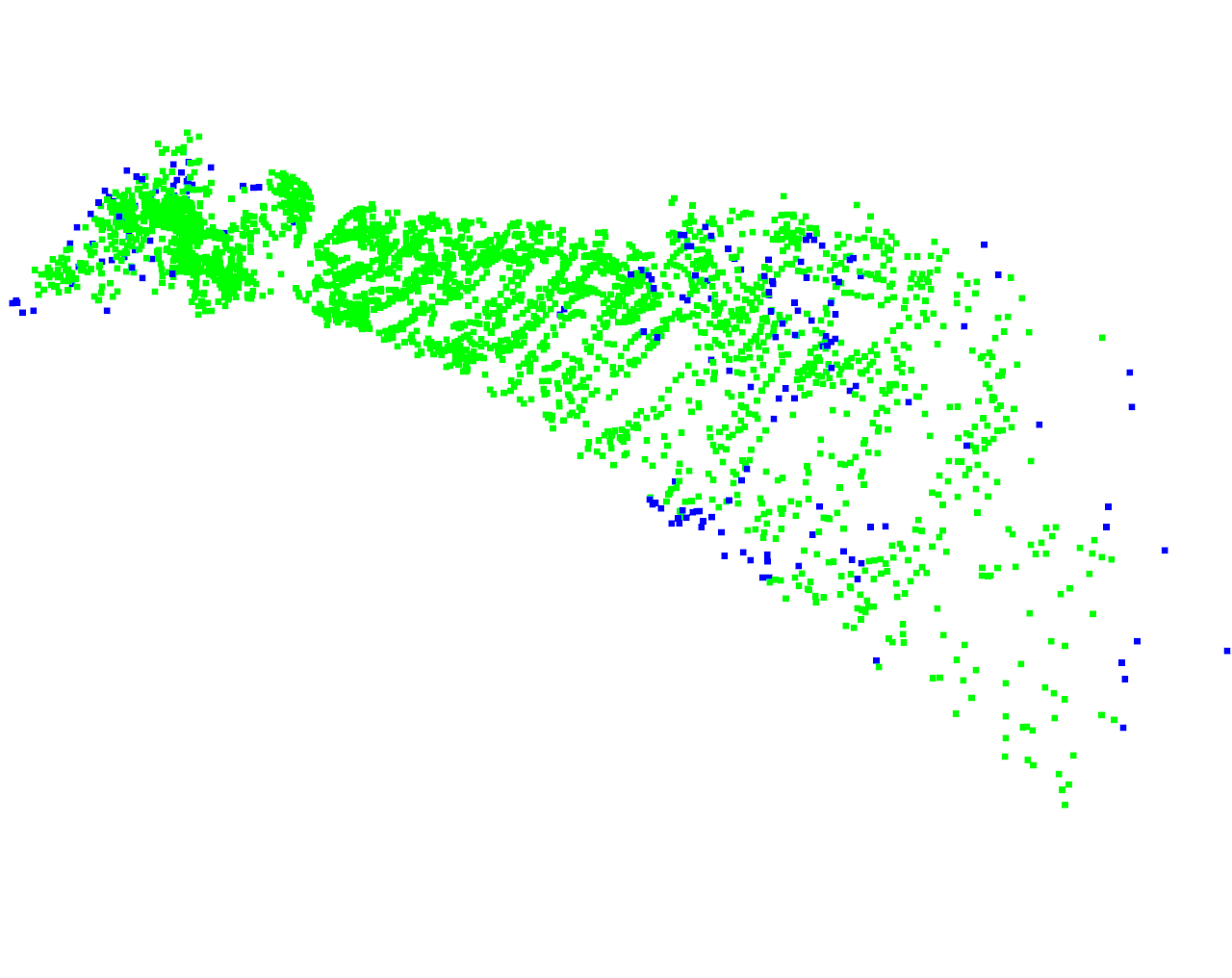}
        & \segimg{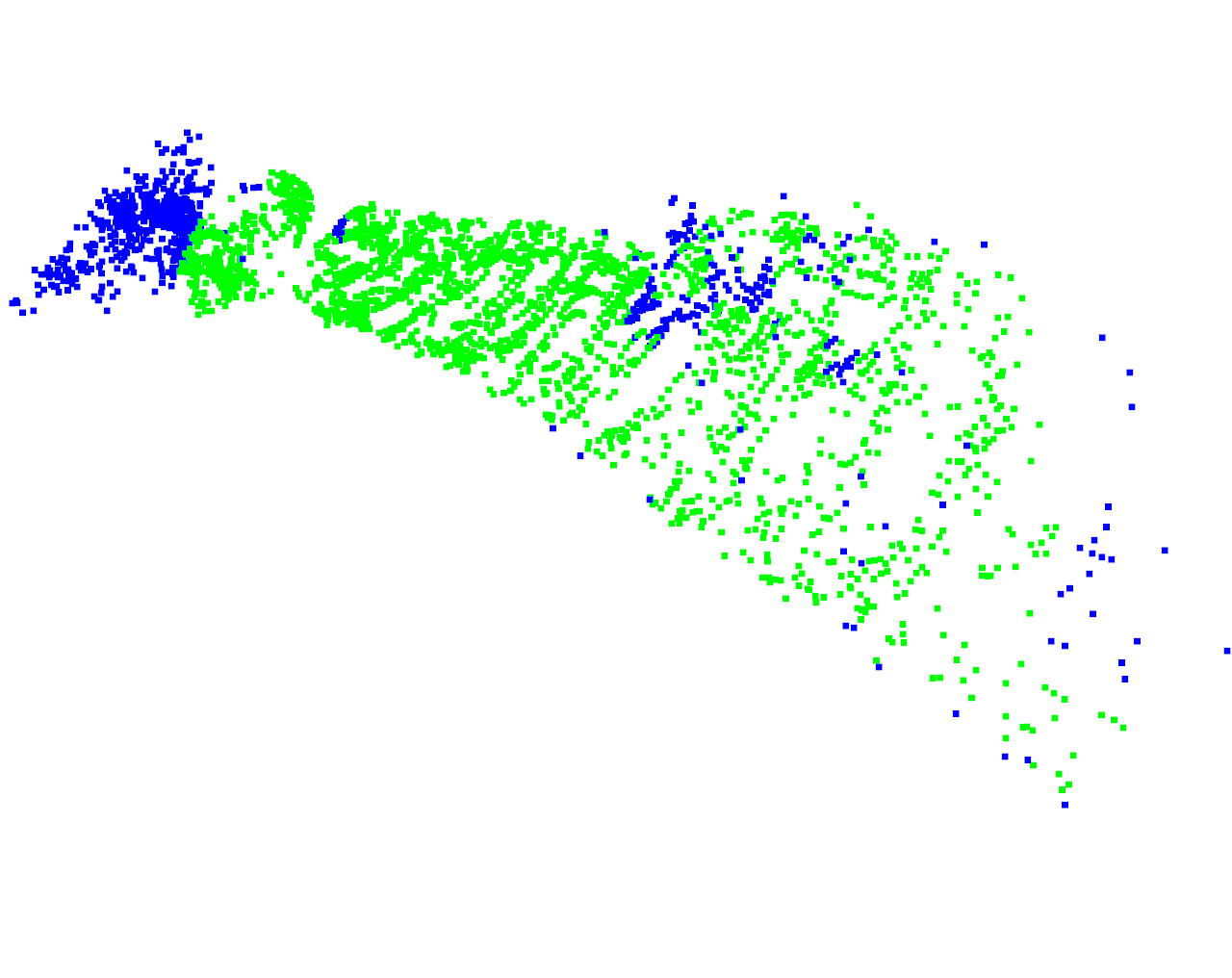}
        & \segimg{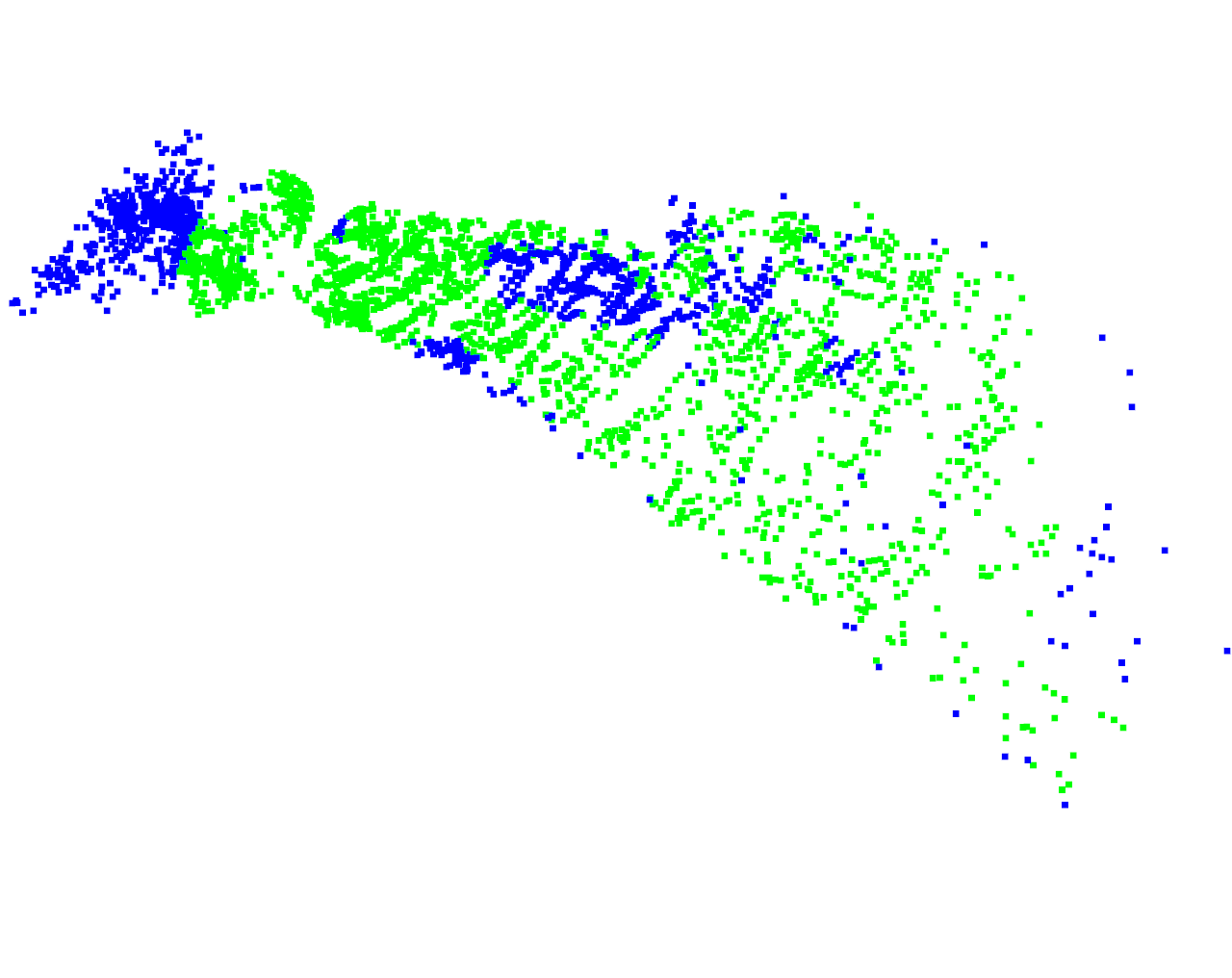}
        & \segimg{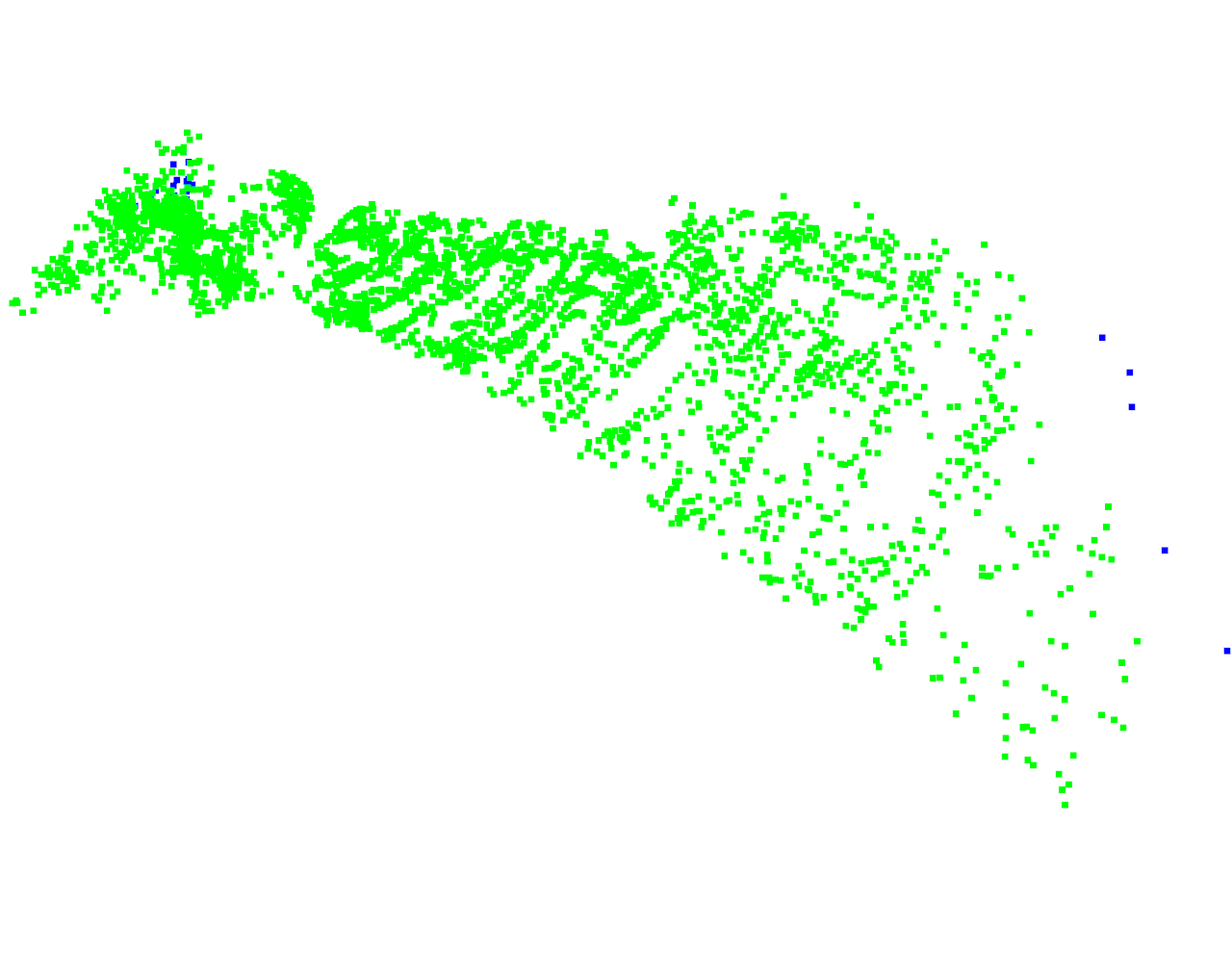} \\

        \seglabel{Water}
        & \segimg{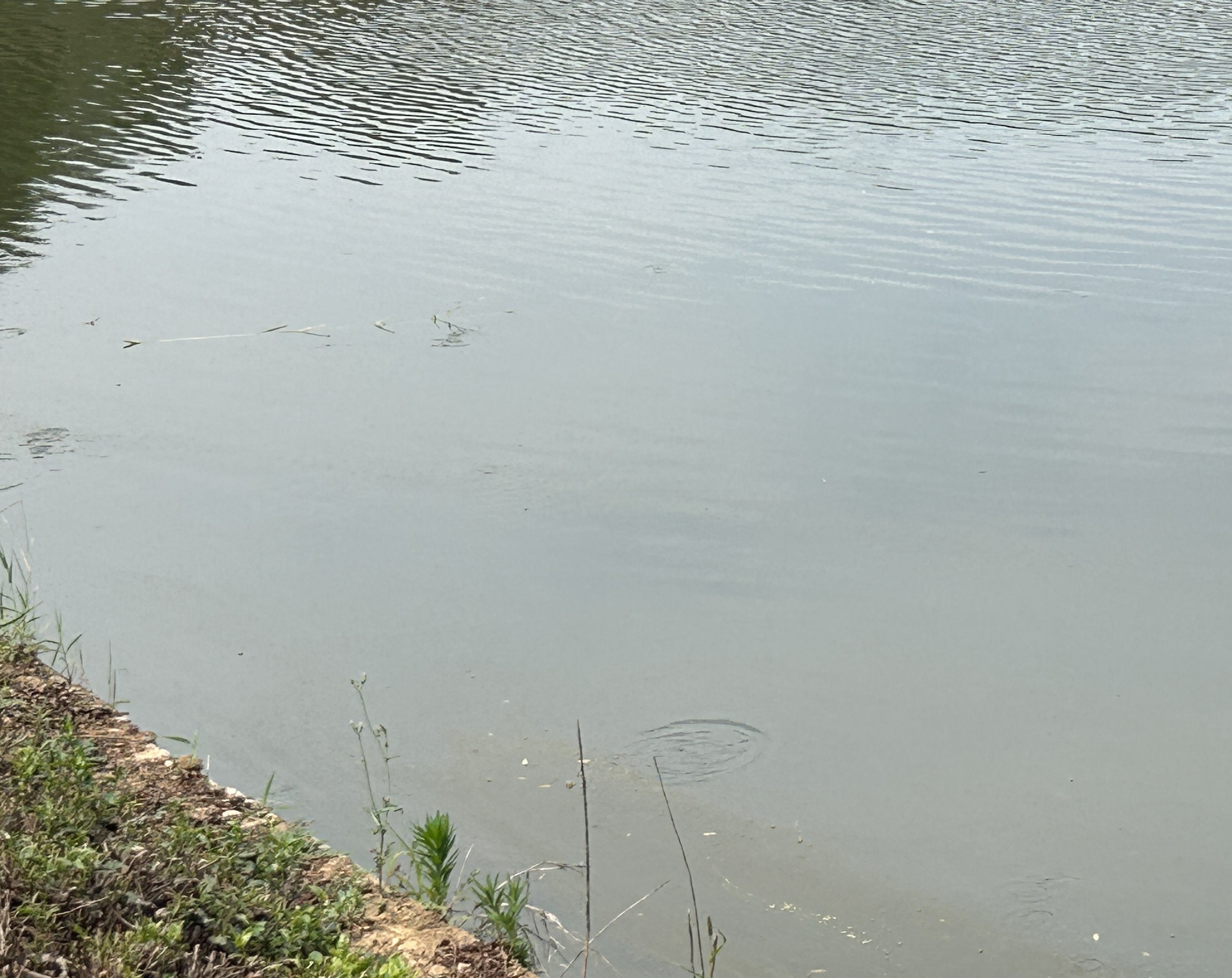}
        & \segimg{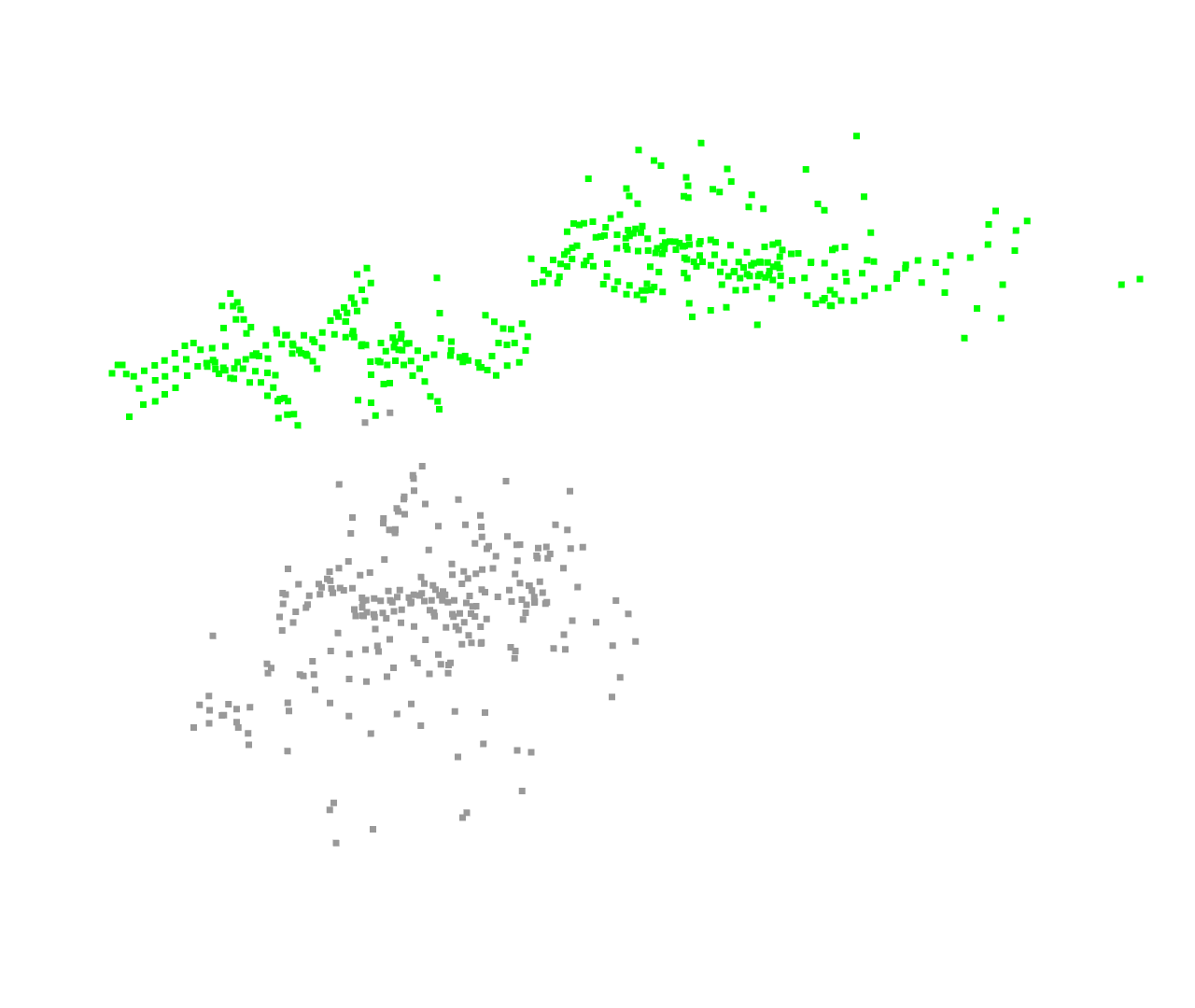}
        & \segimg{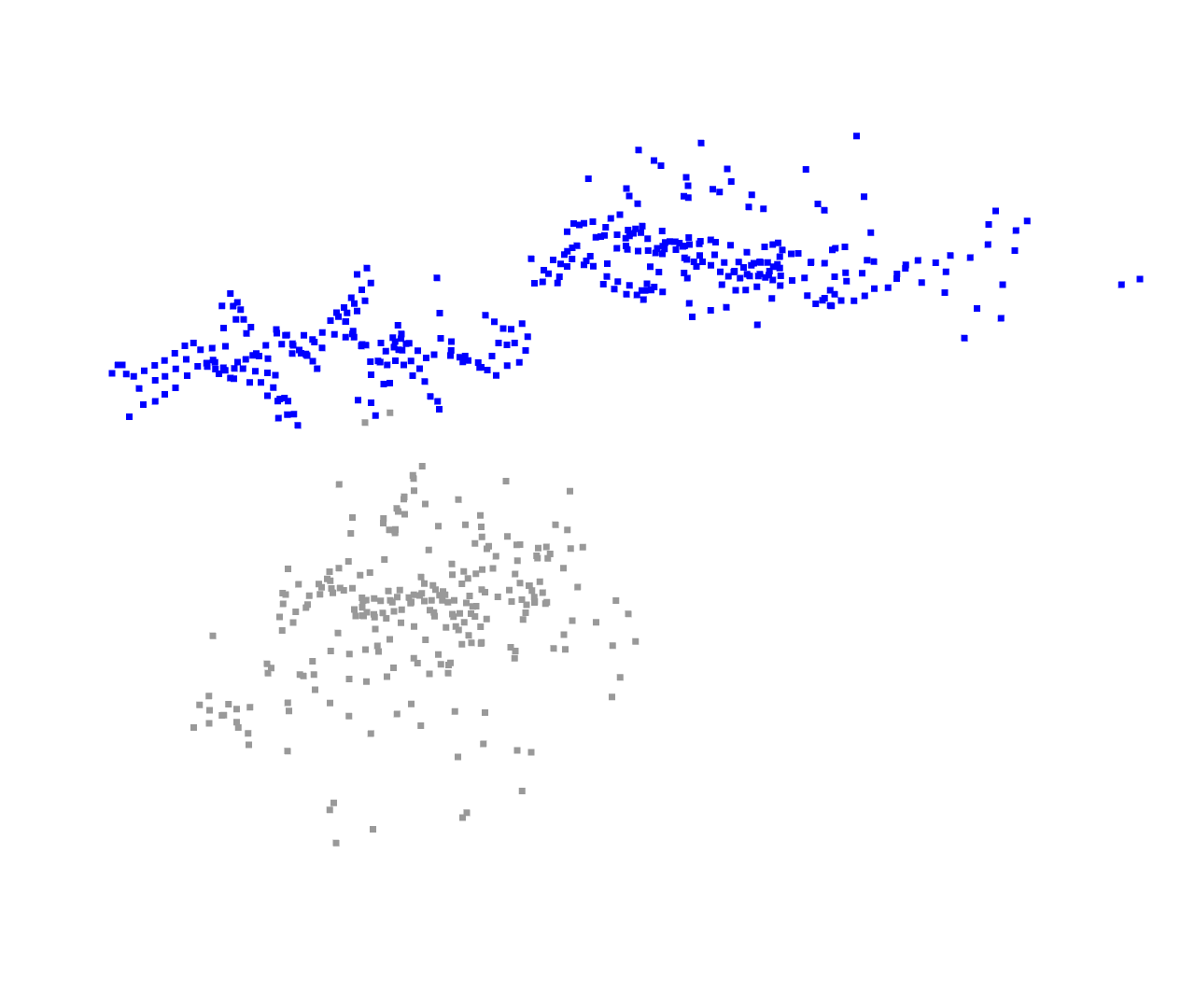}
        & \segimg{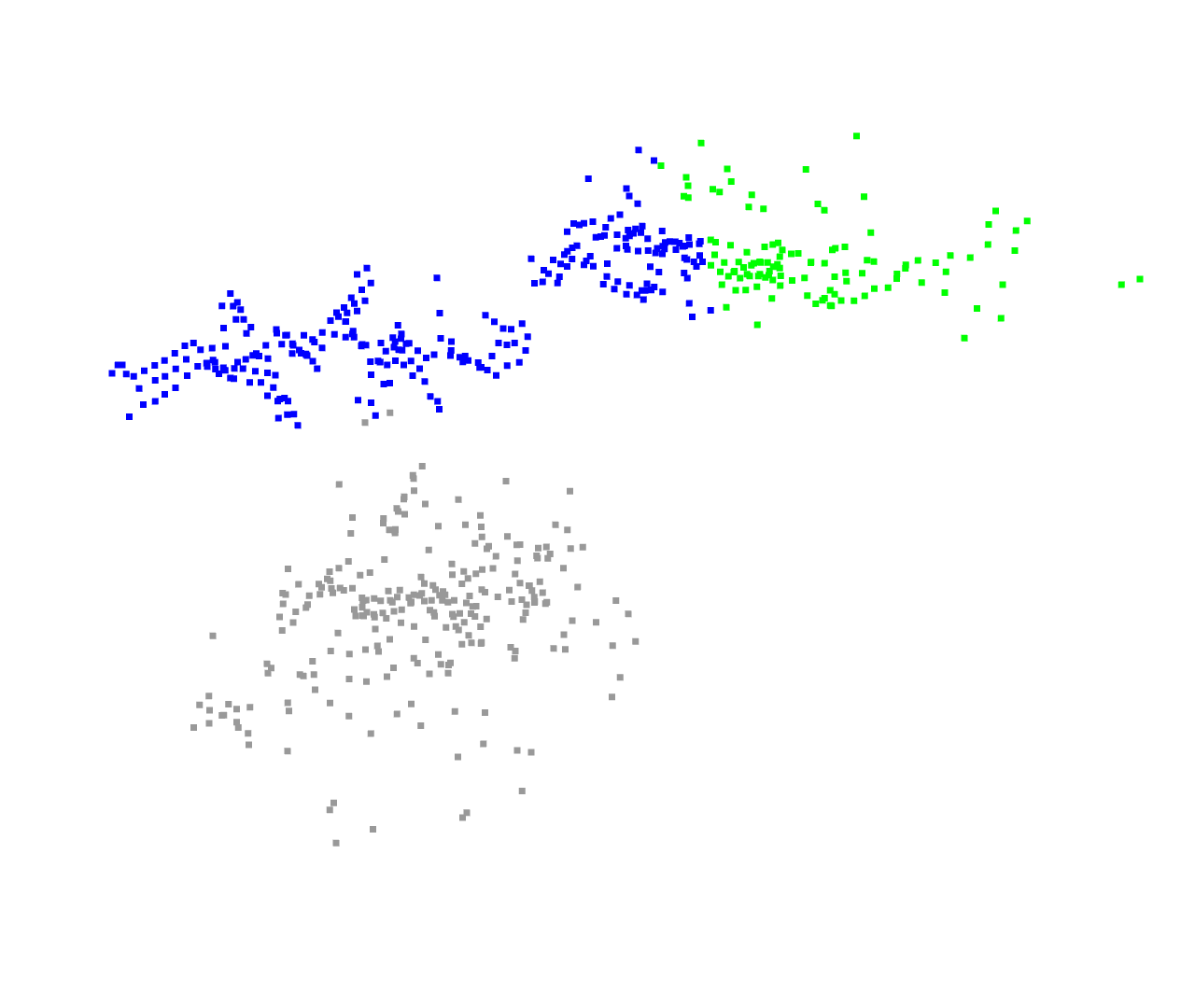}
        & \segimg{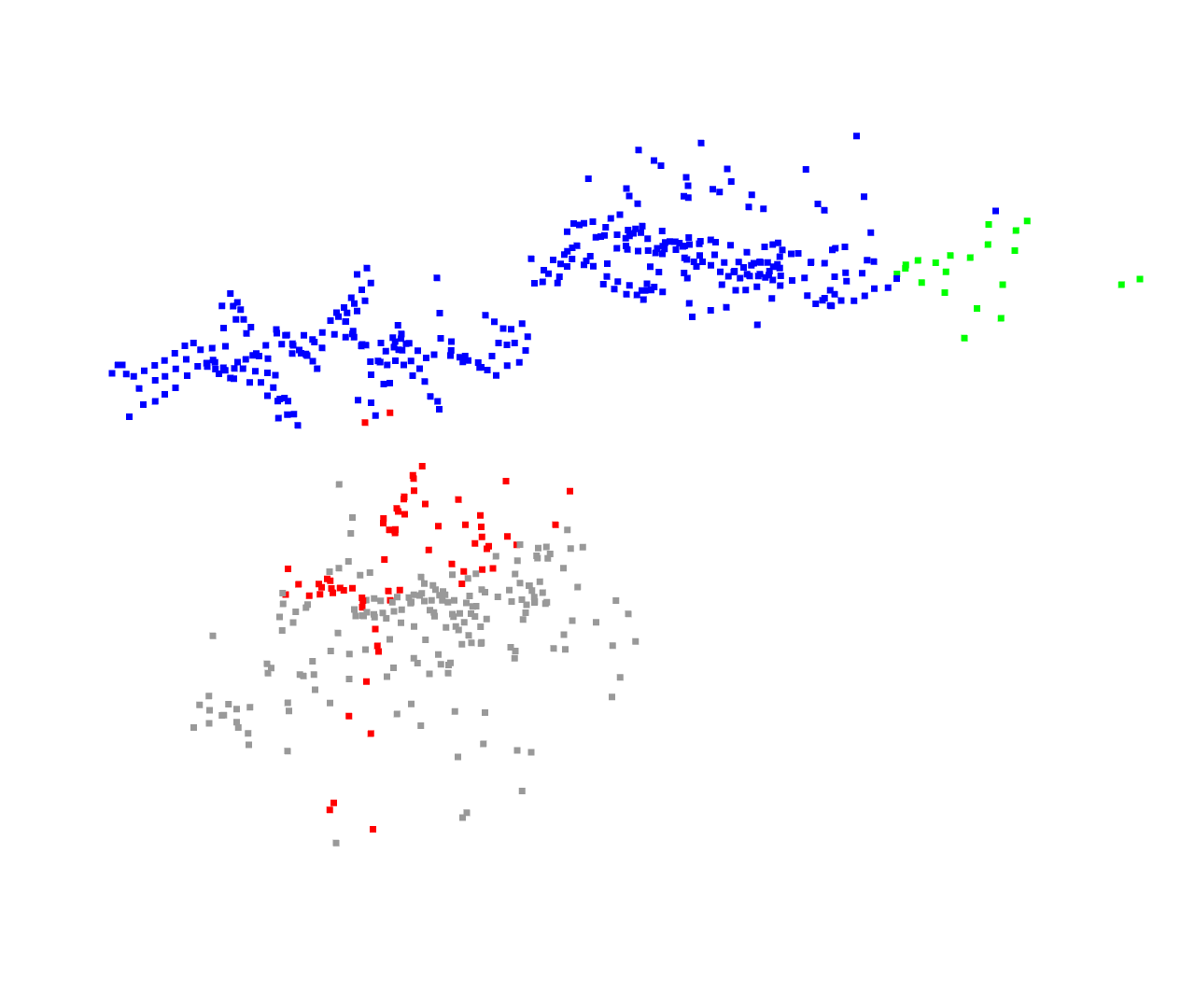}
        & \segimg{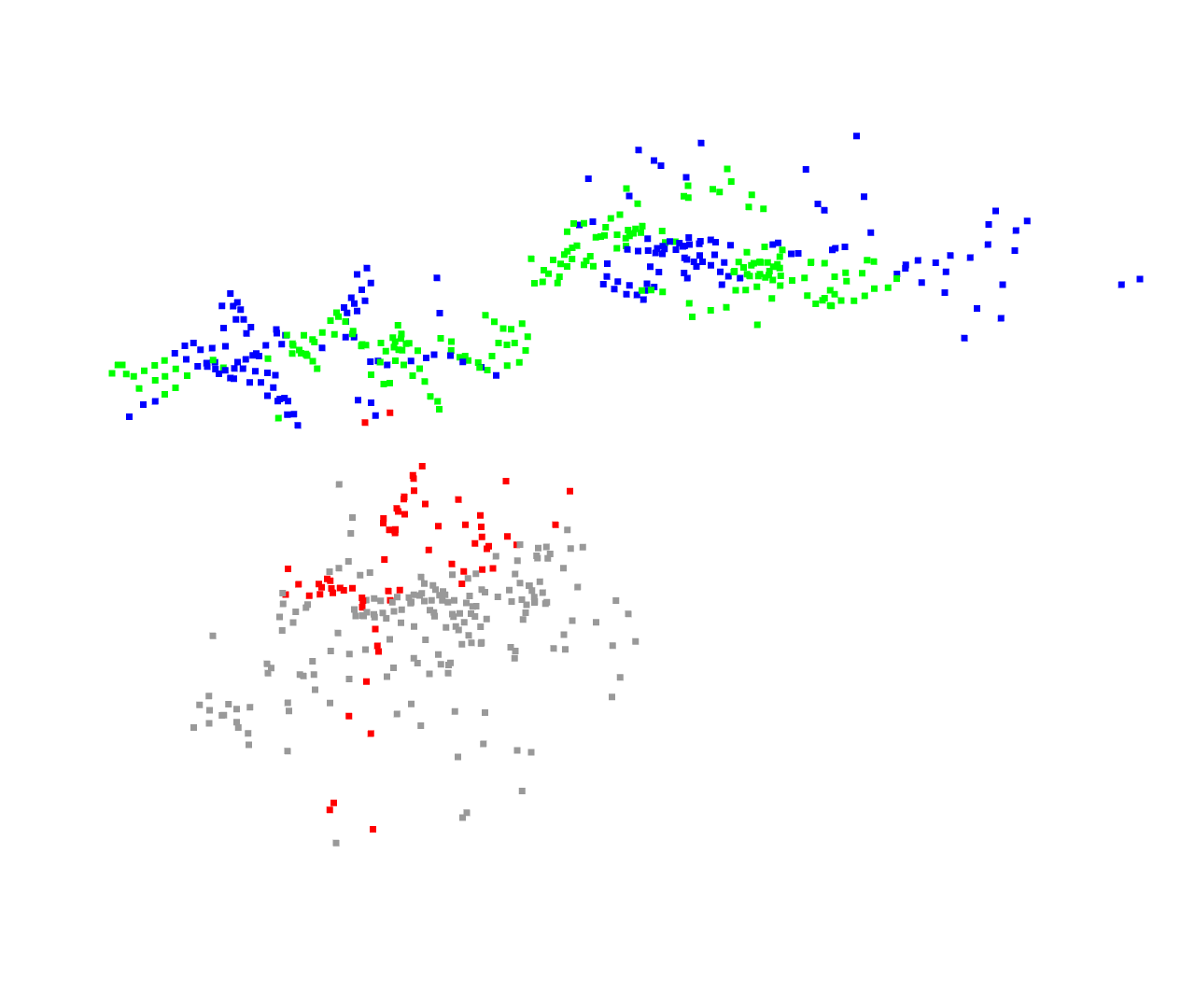}
        & \segimg{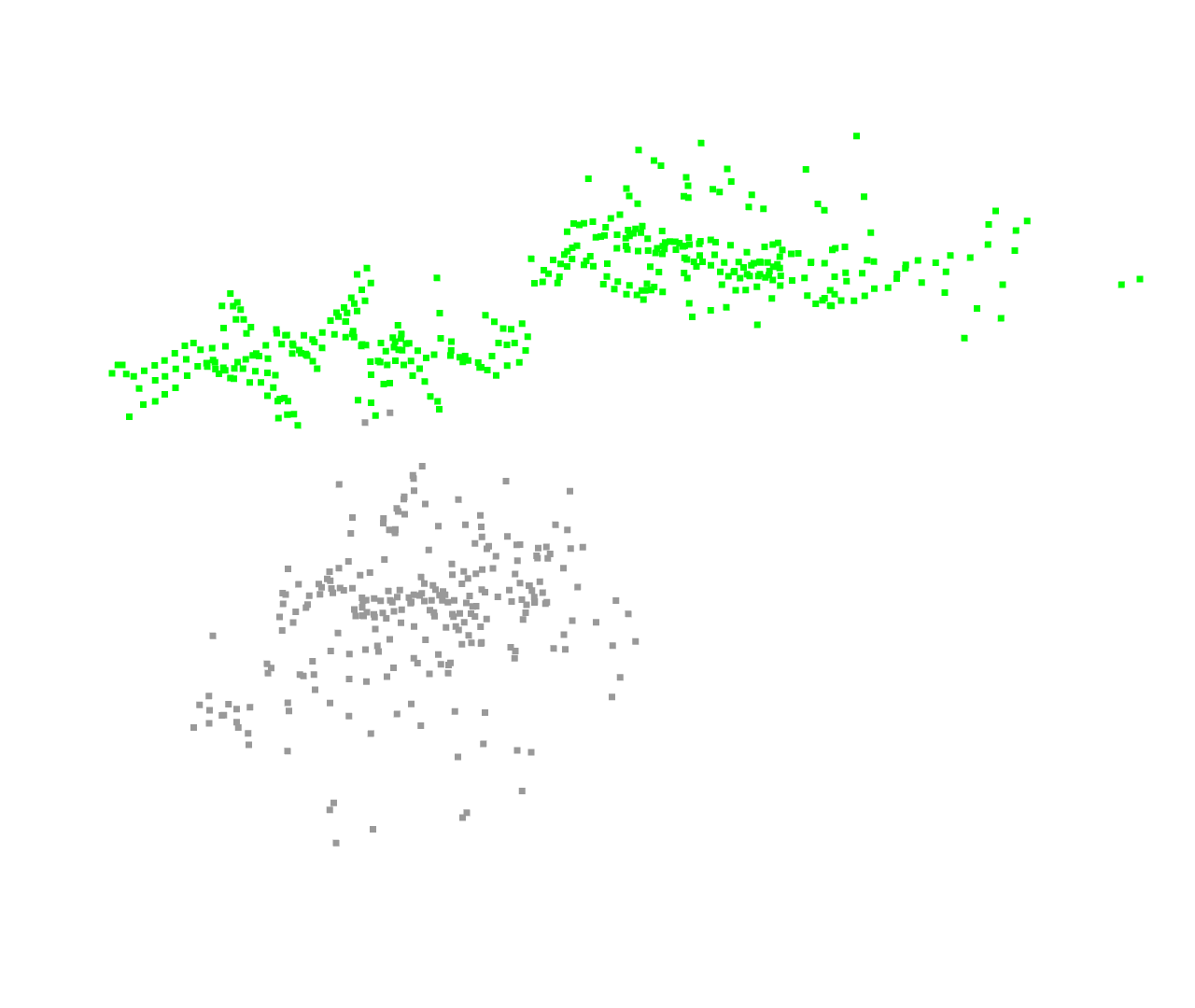} \\

        \seglabel{Tea-1}
        & \segimg{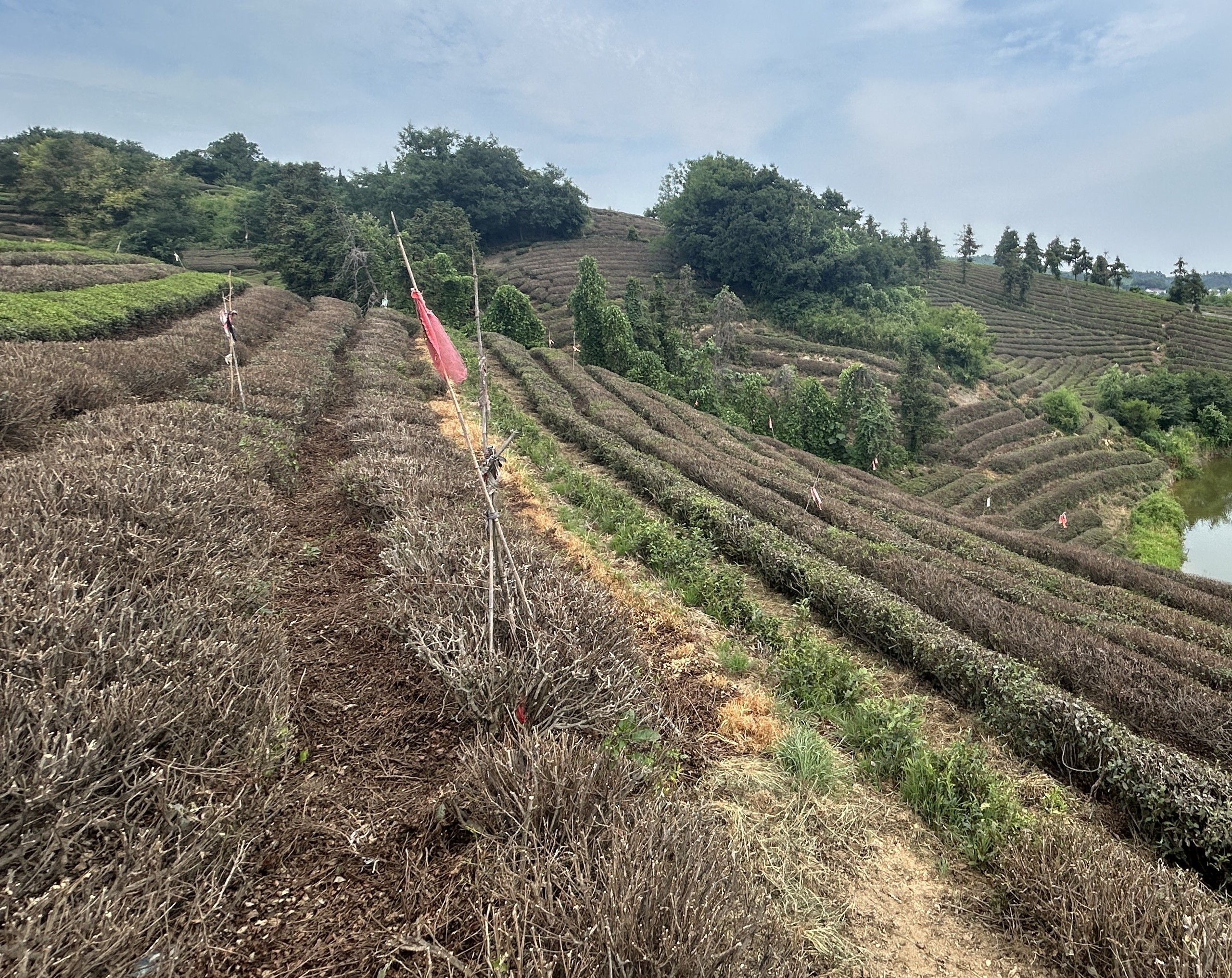}
        & \segimg{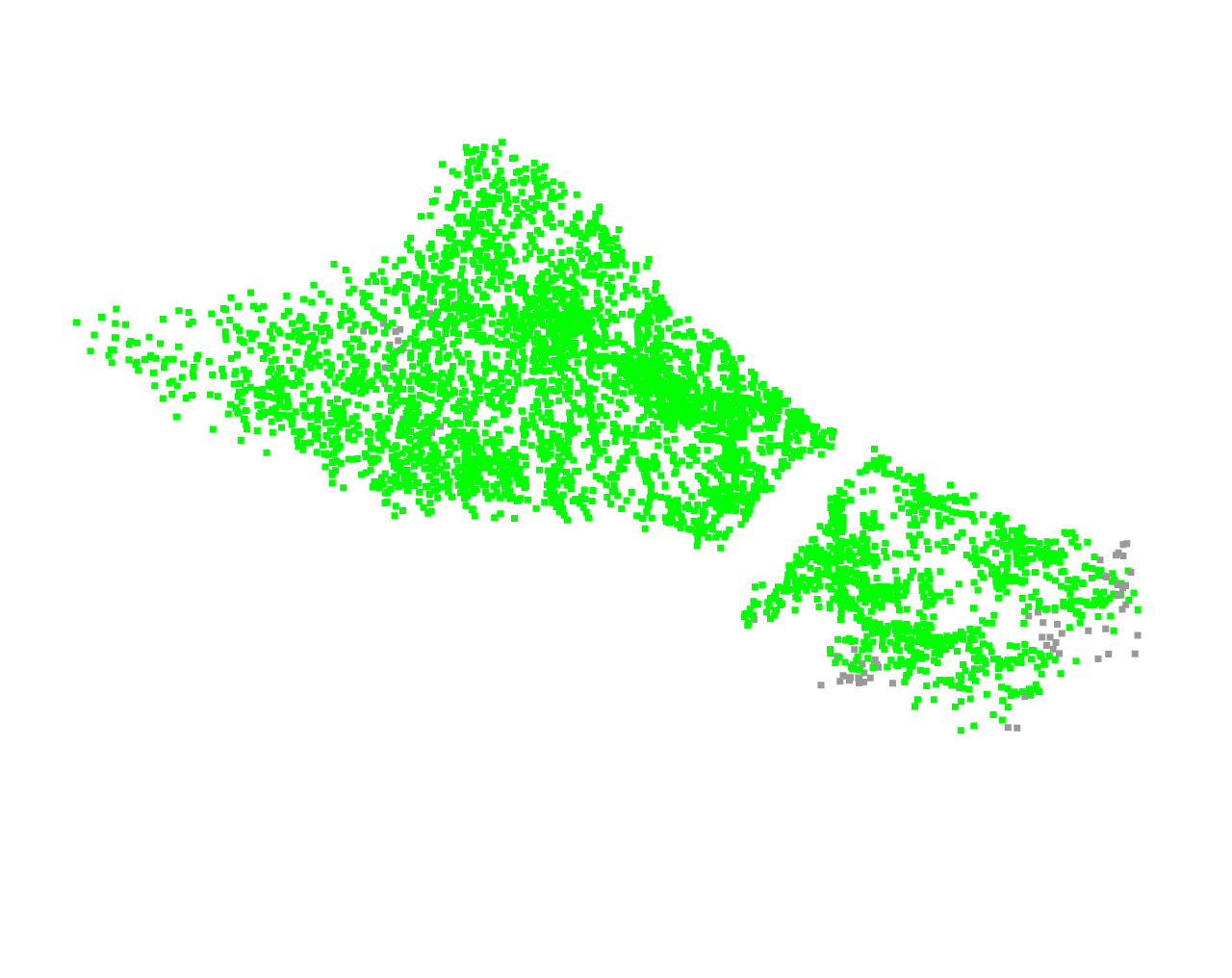}
        & \segimg{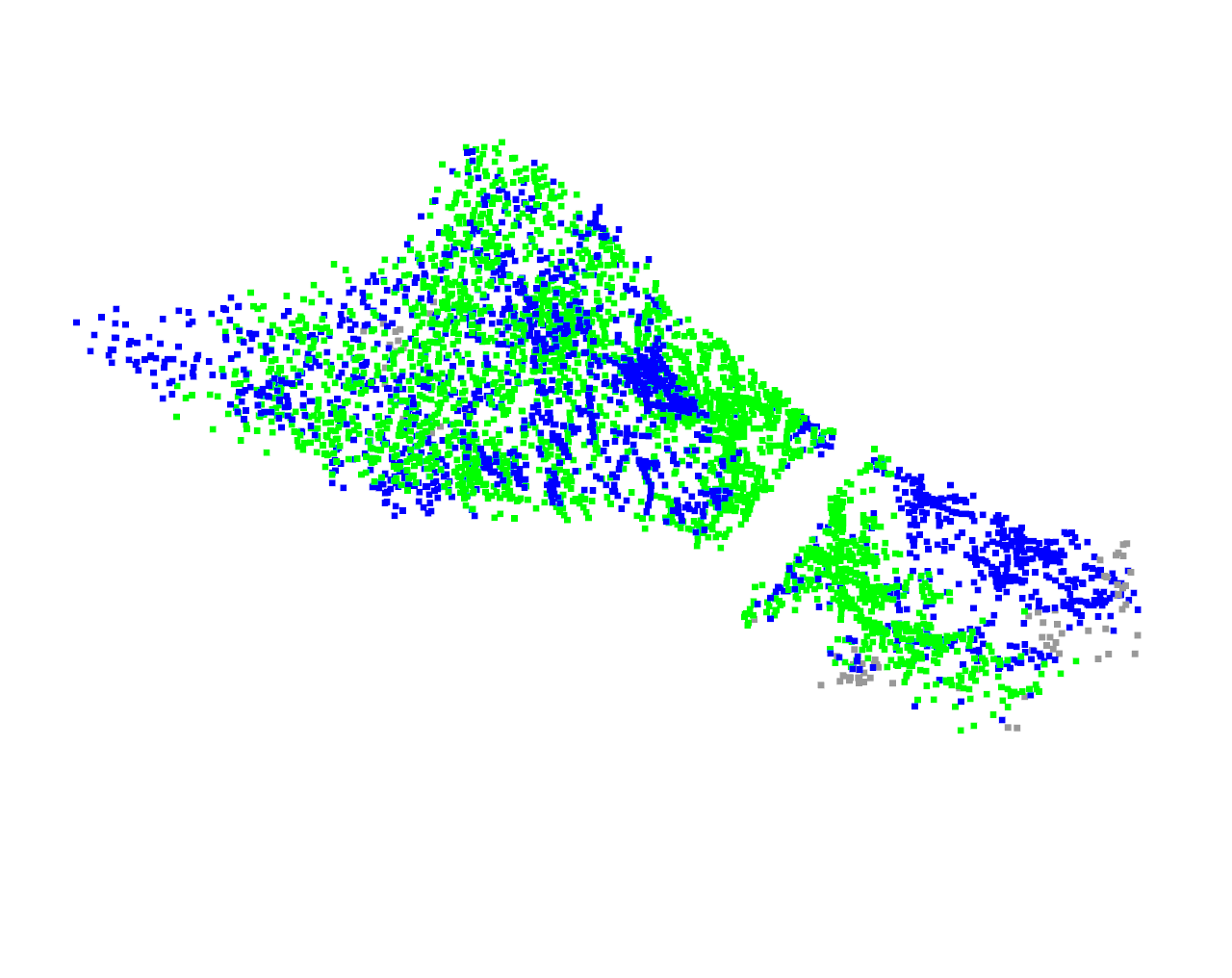}
        & \segimg{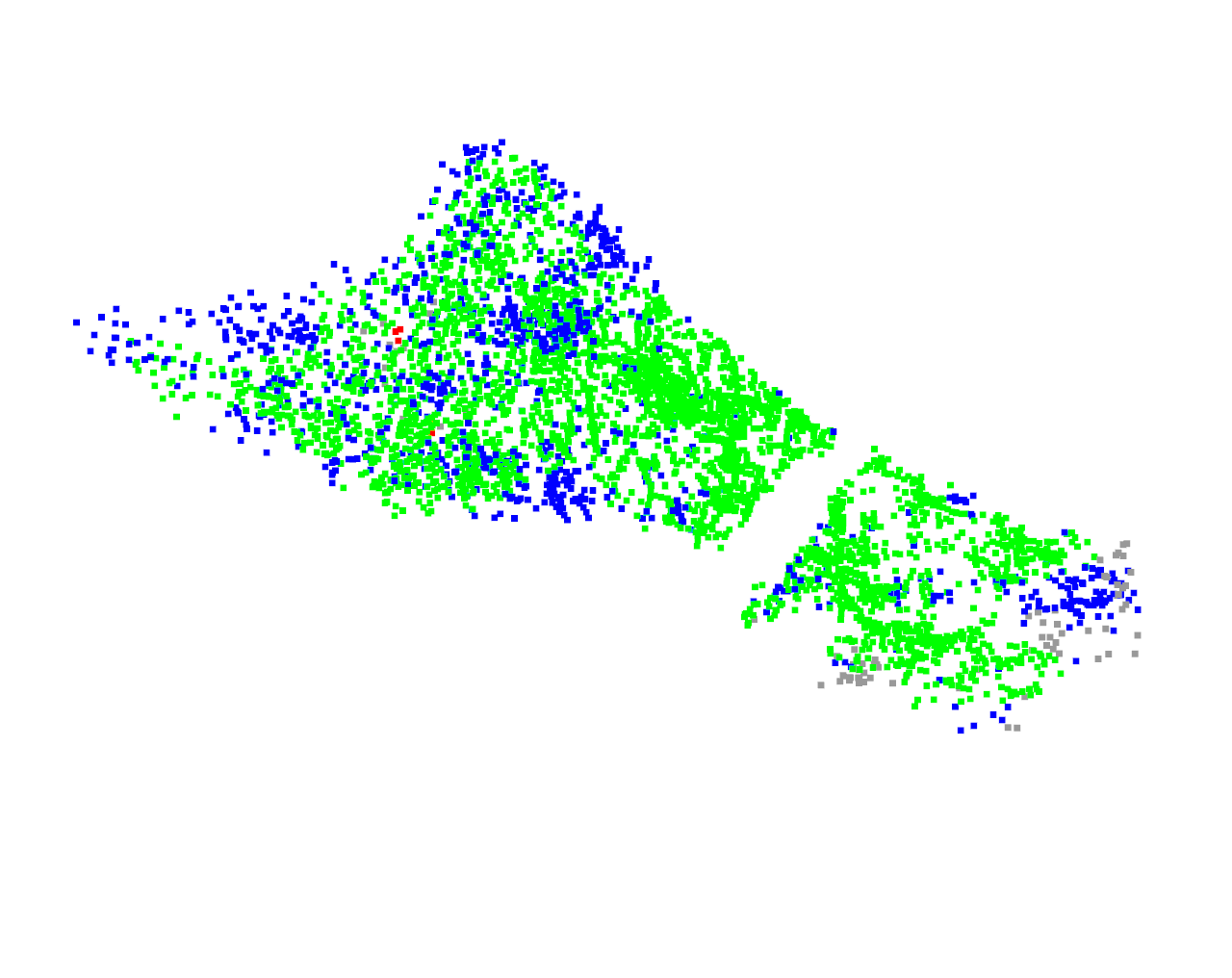}
        & \segimg{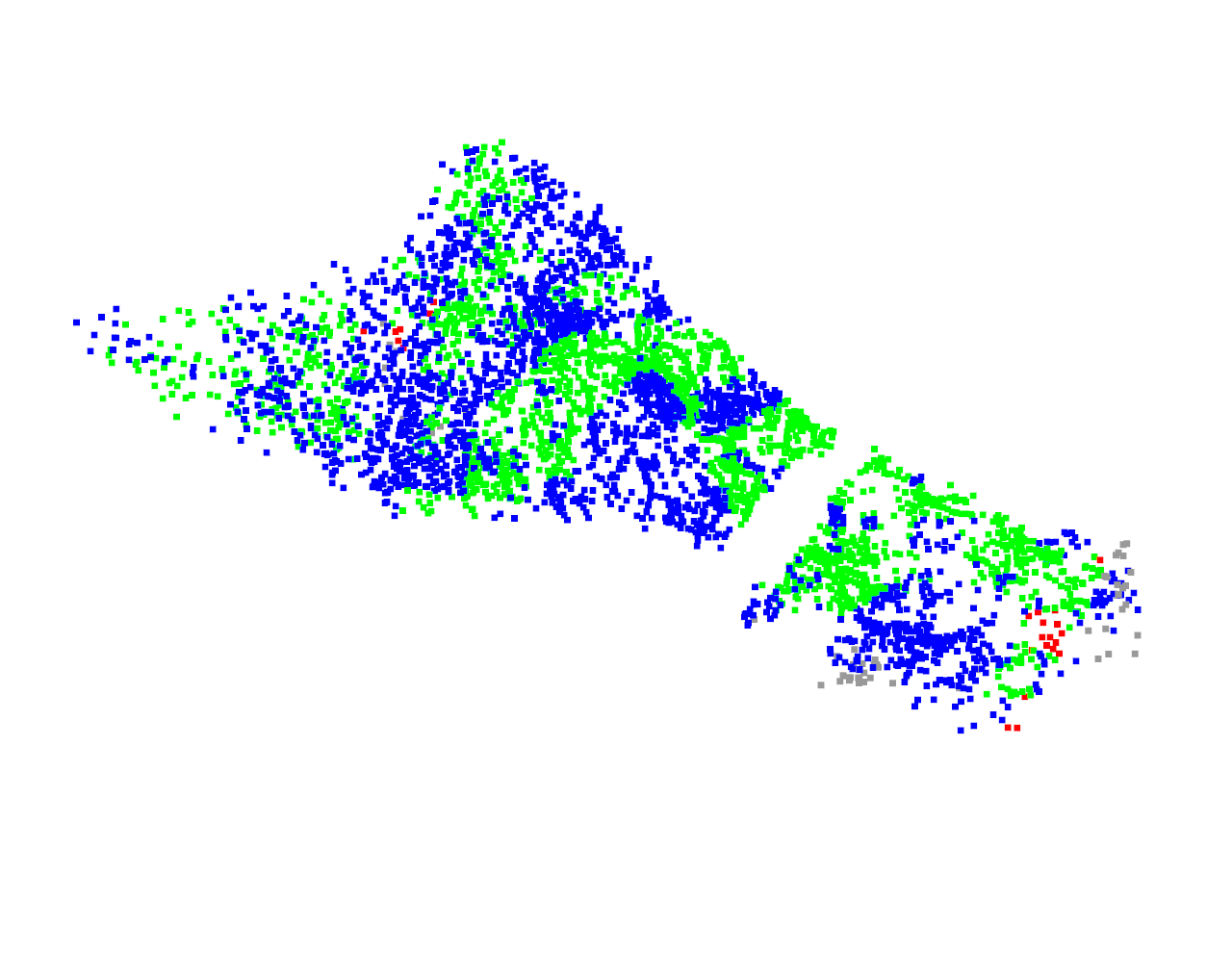}
        & \segimg{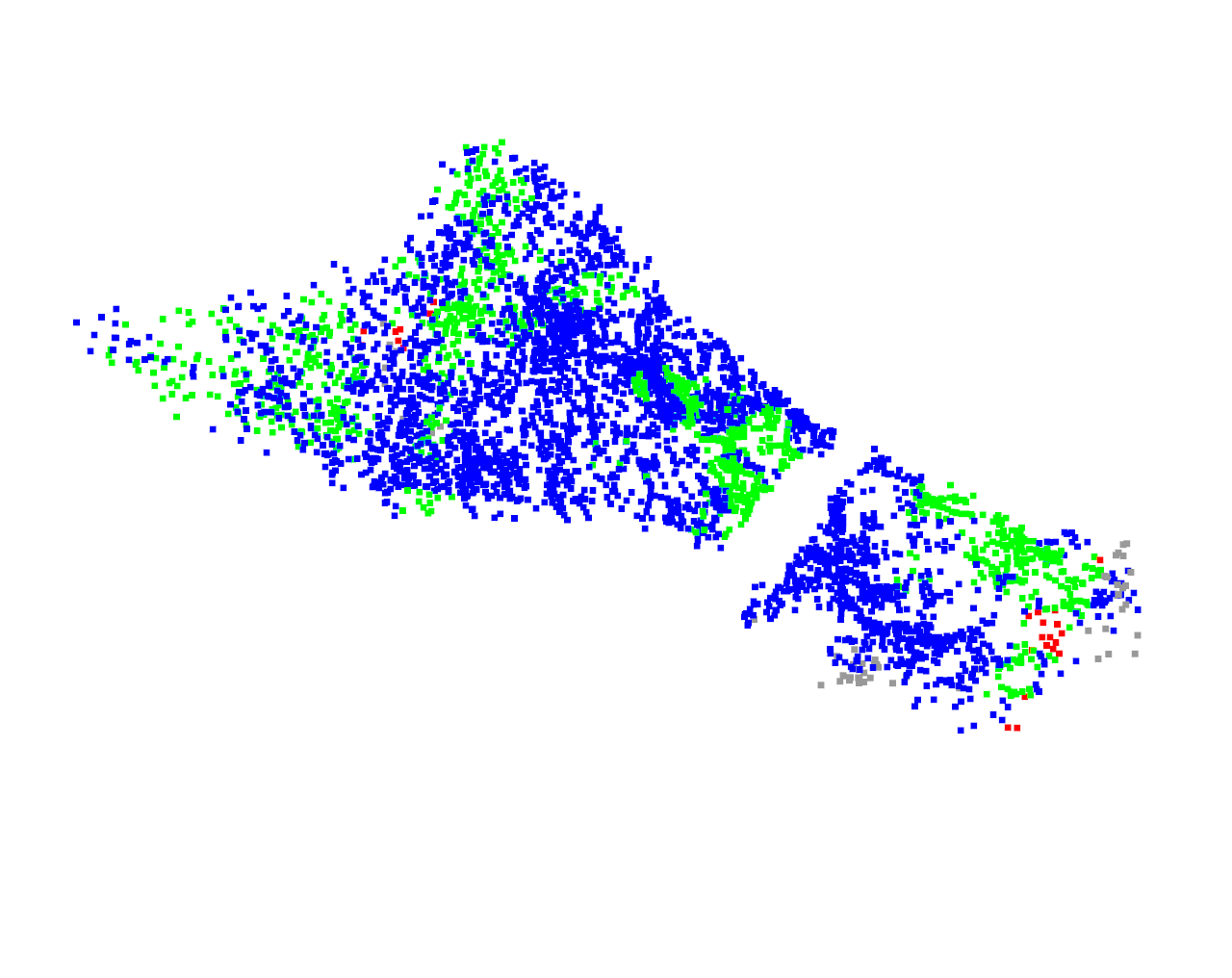}
        & \segimg{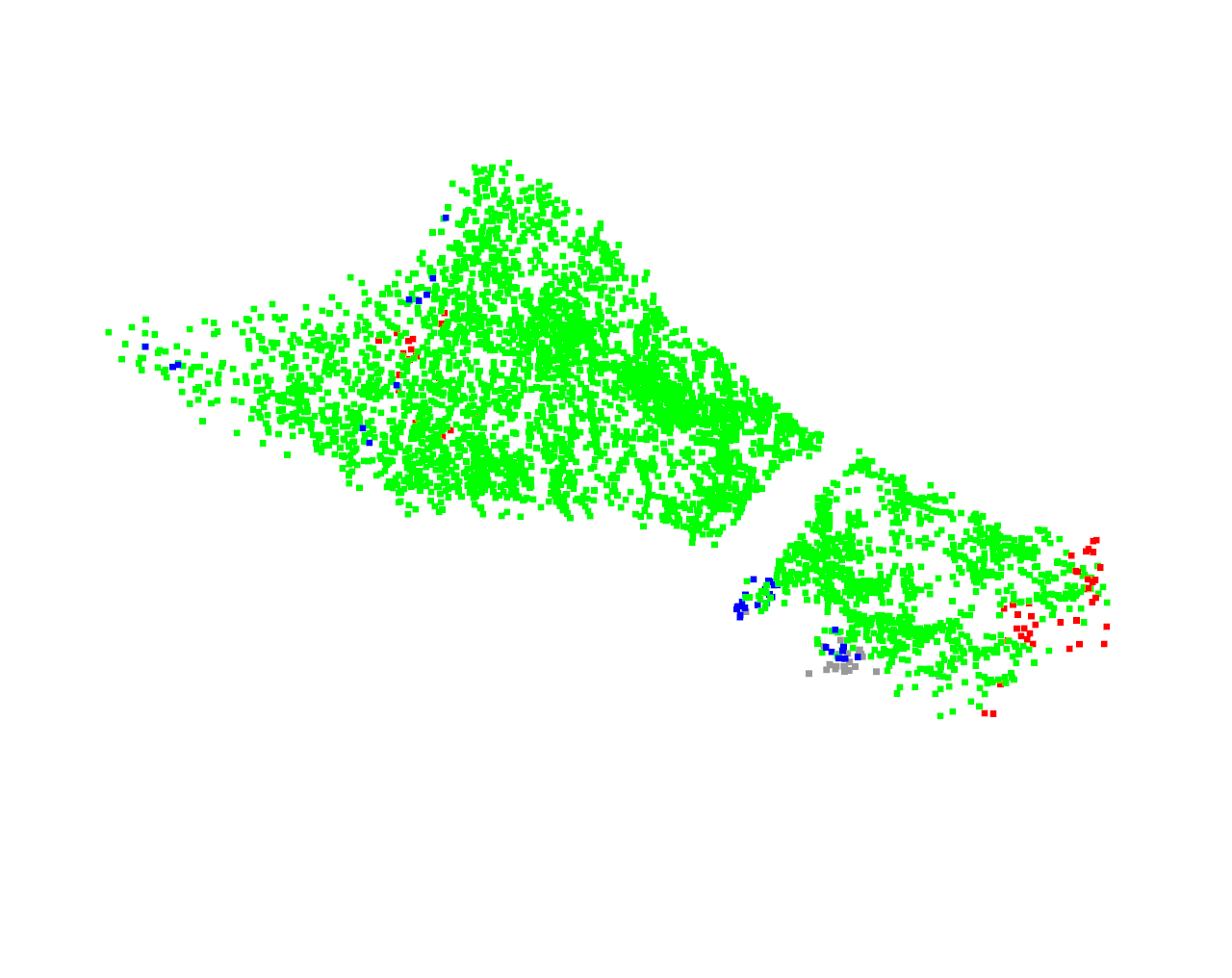} \\

        \seglabel{Tea-2}
        & \segimg{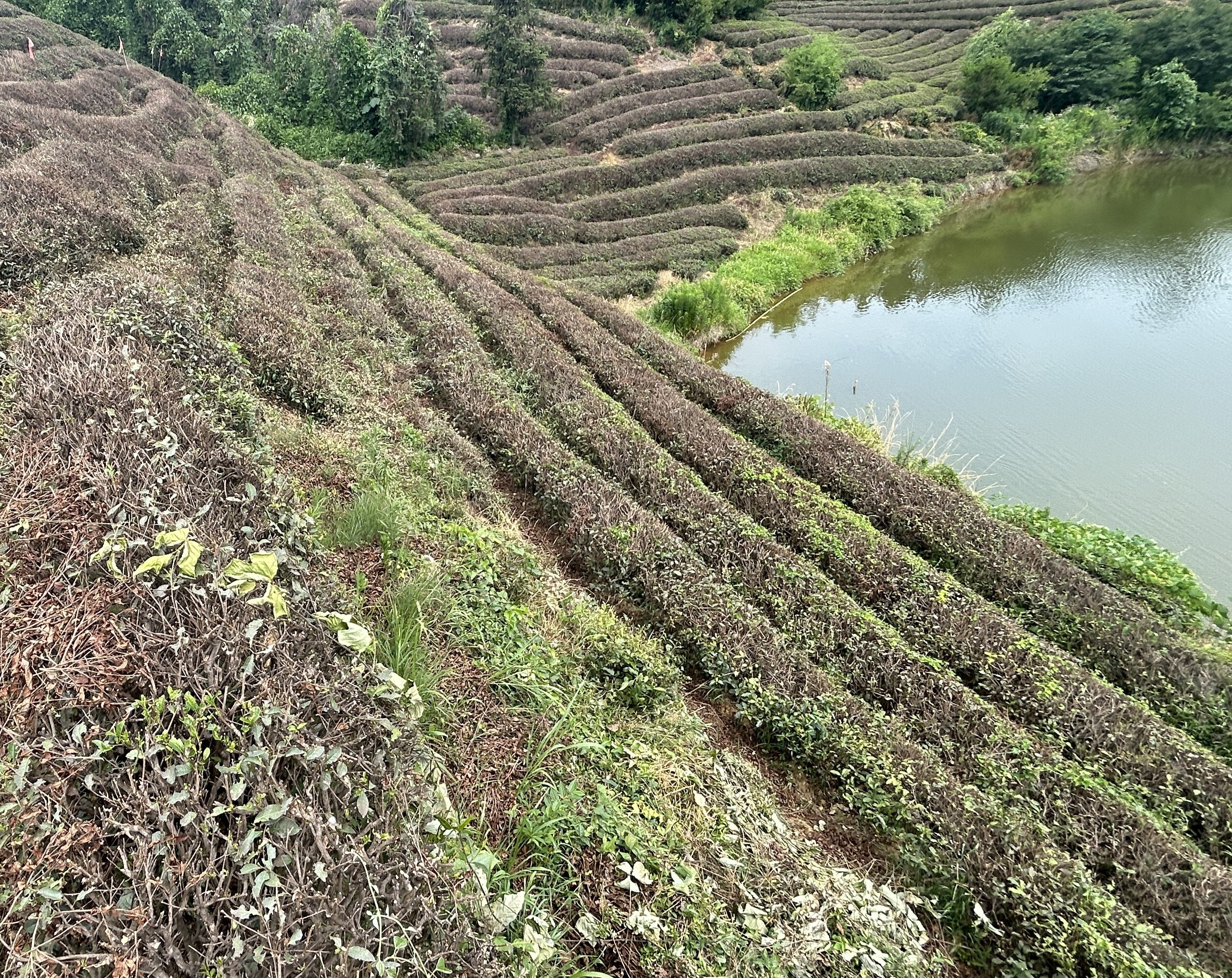}
        & \segimg{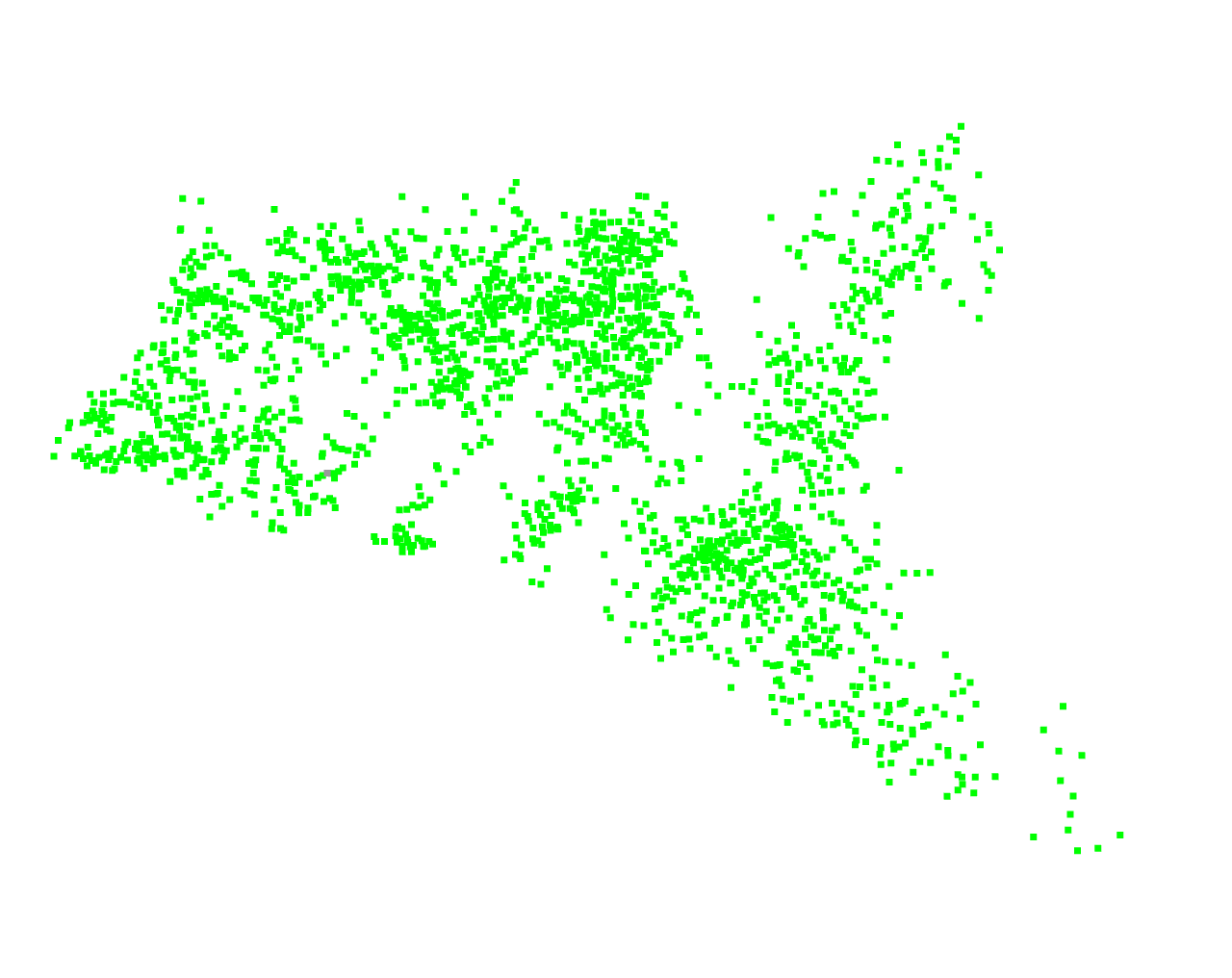}
        & \segimg{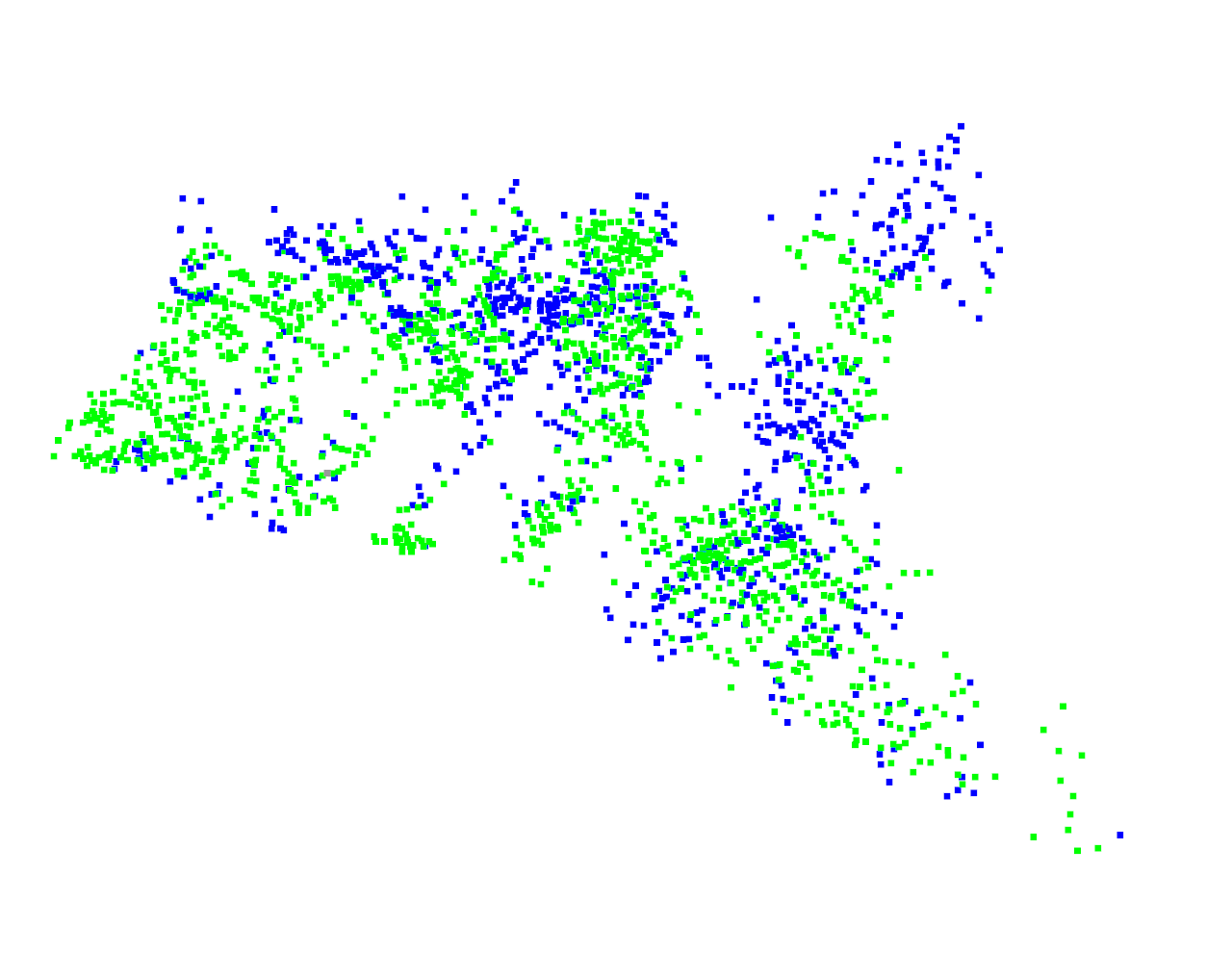}
        & \segimg{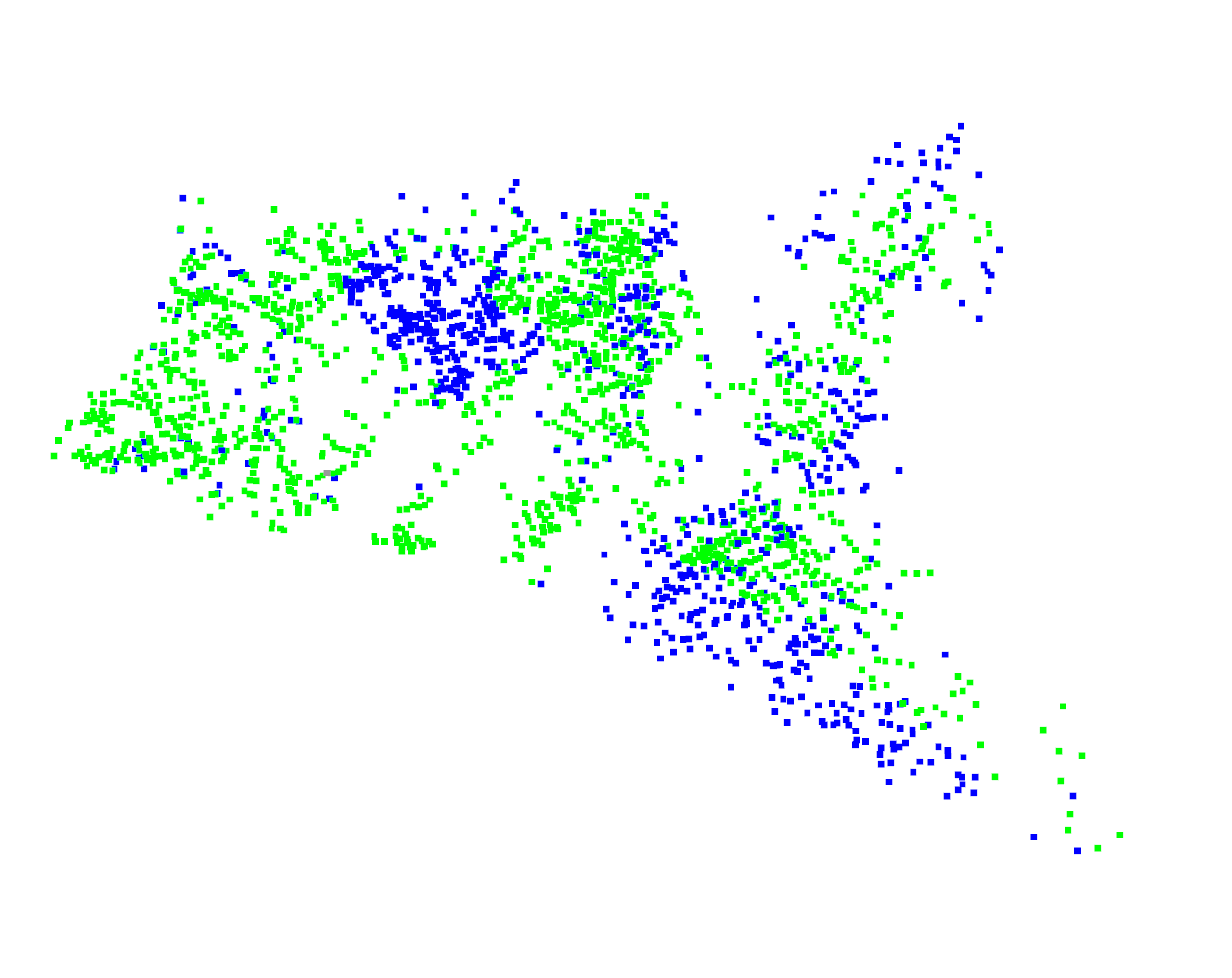}
        & \segimg{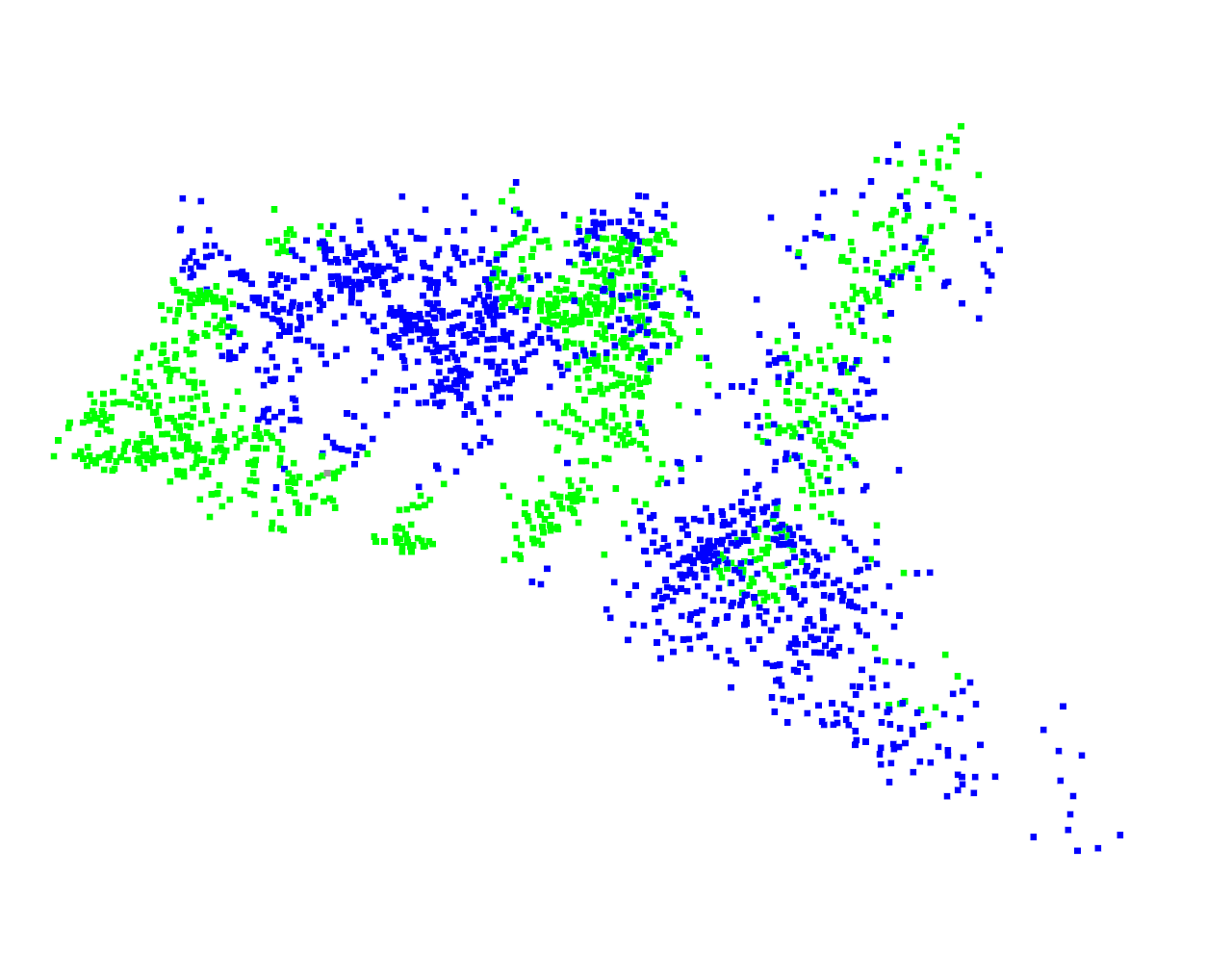}
        & \segimg{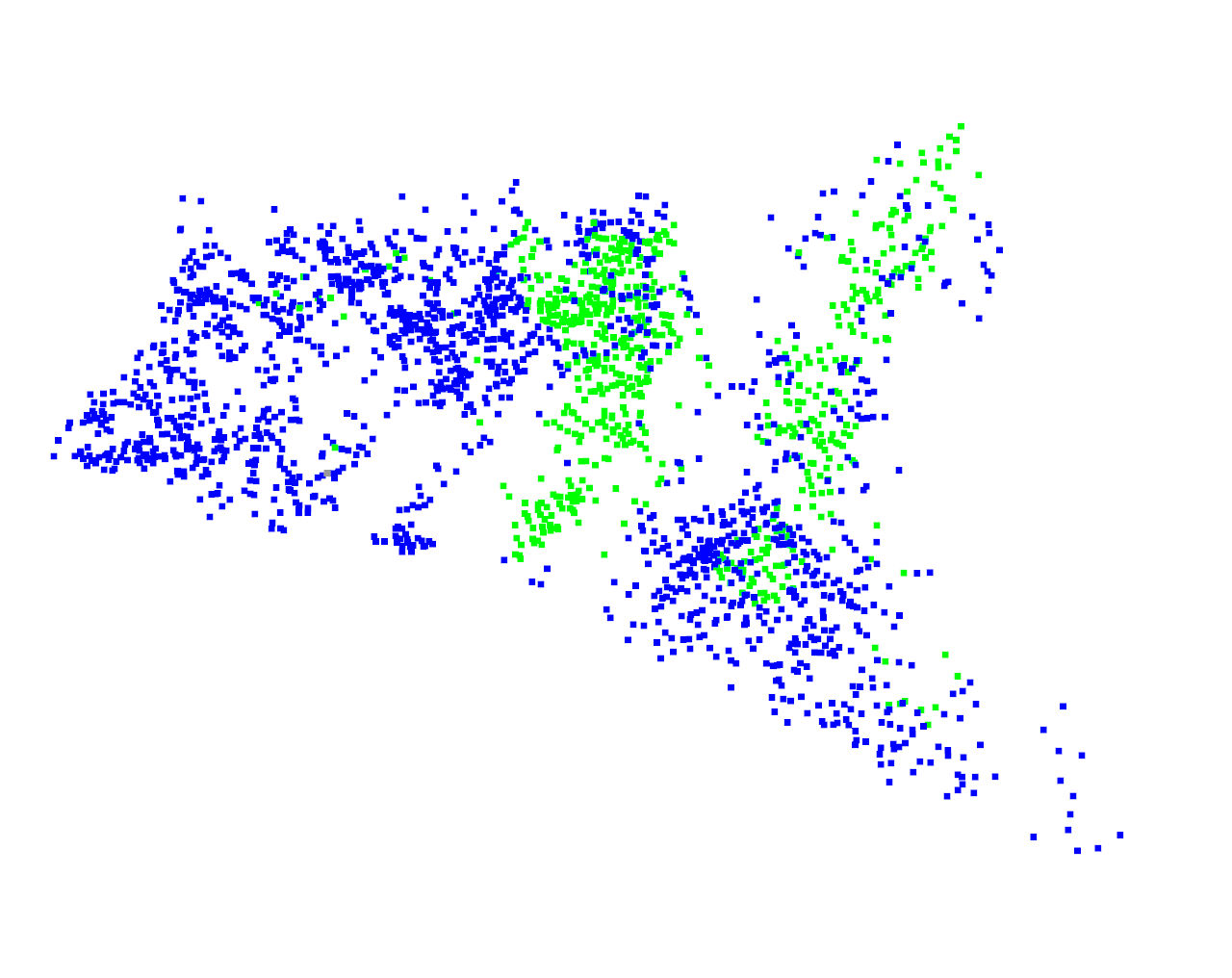}
        & \segimg{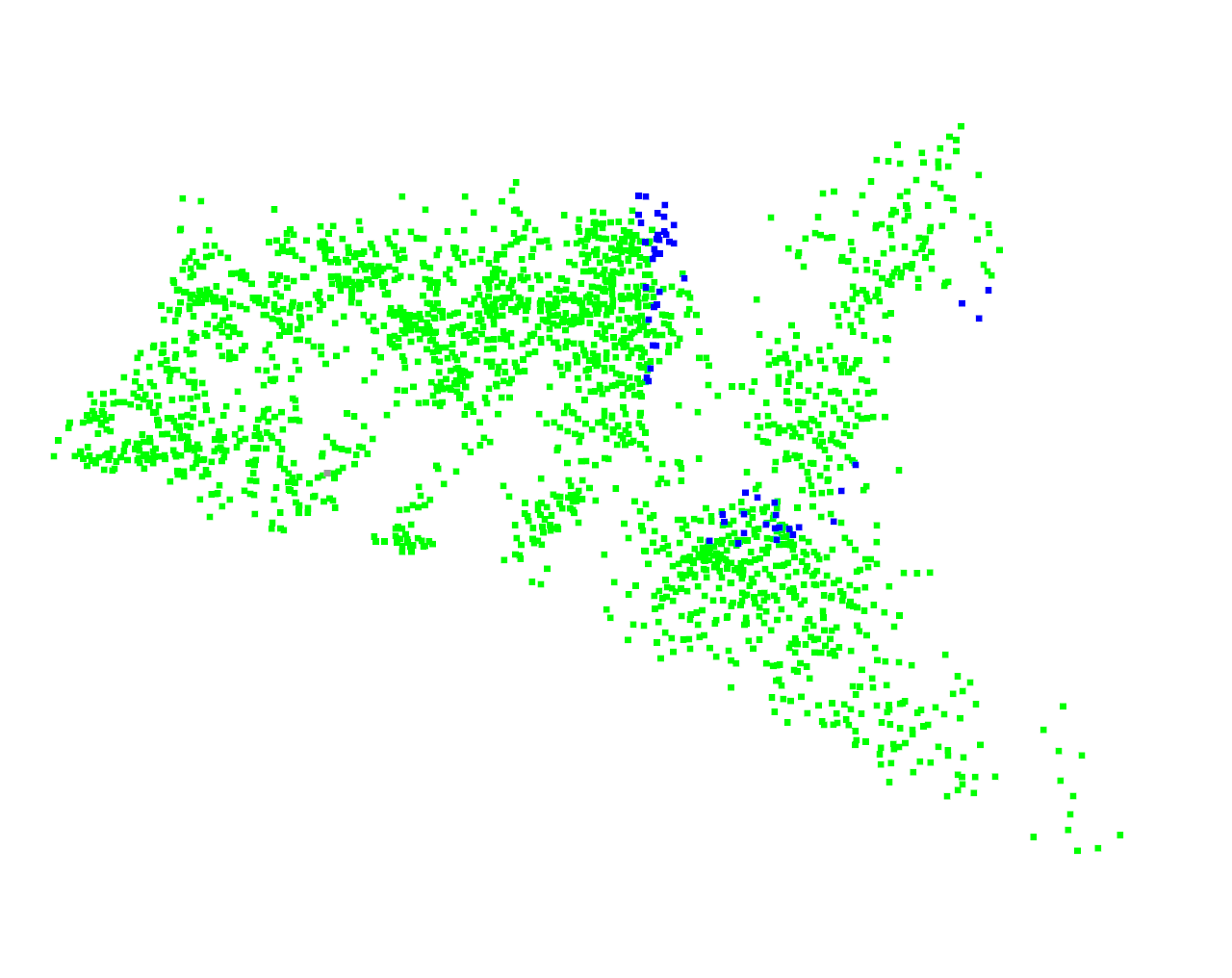} \\

        \seglabel{Tea-3}
        & \segimg{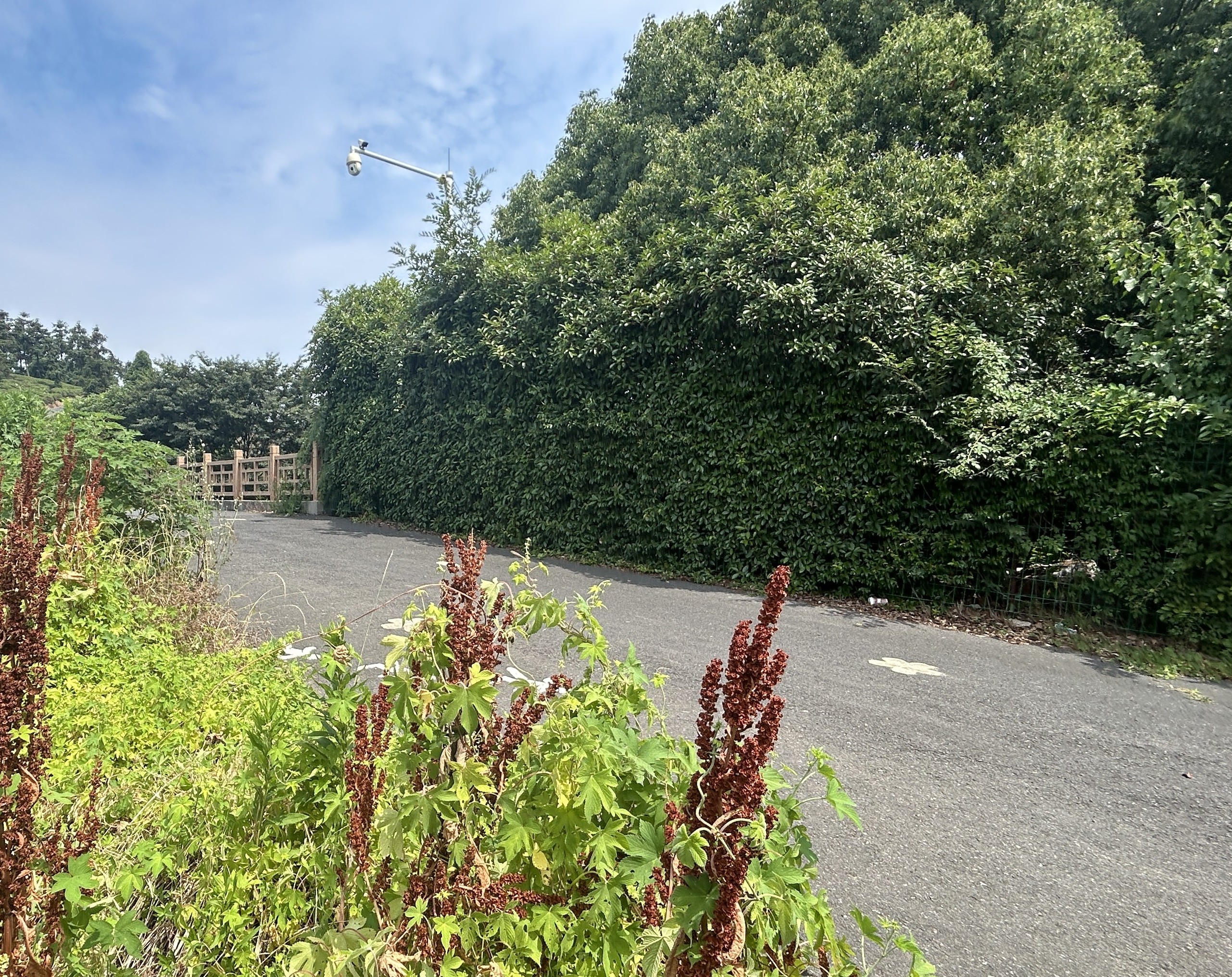}
        & \segimg{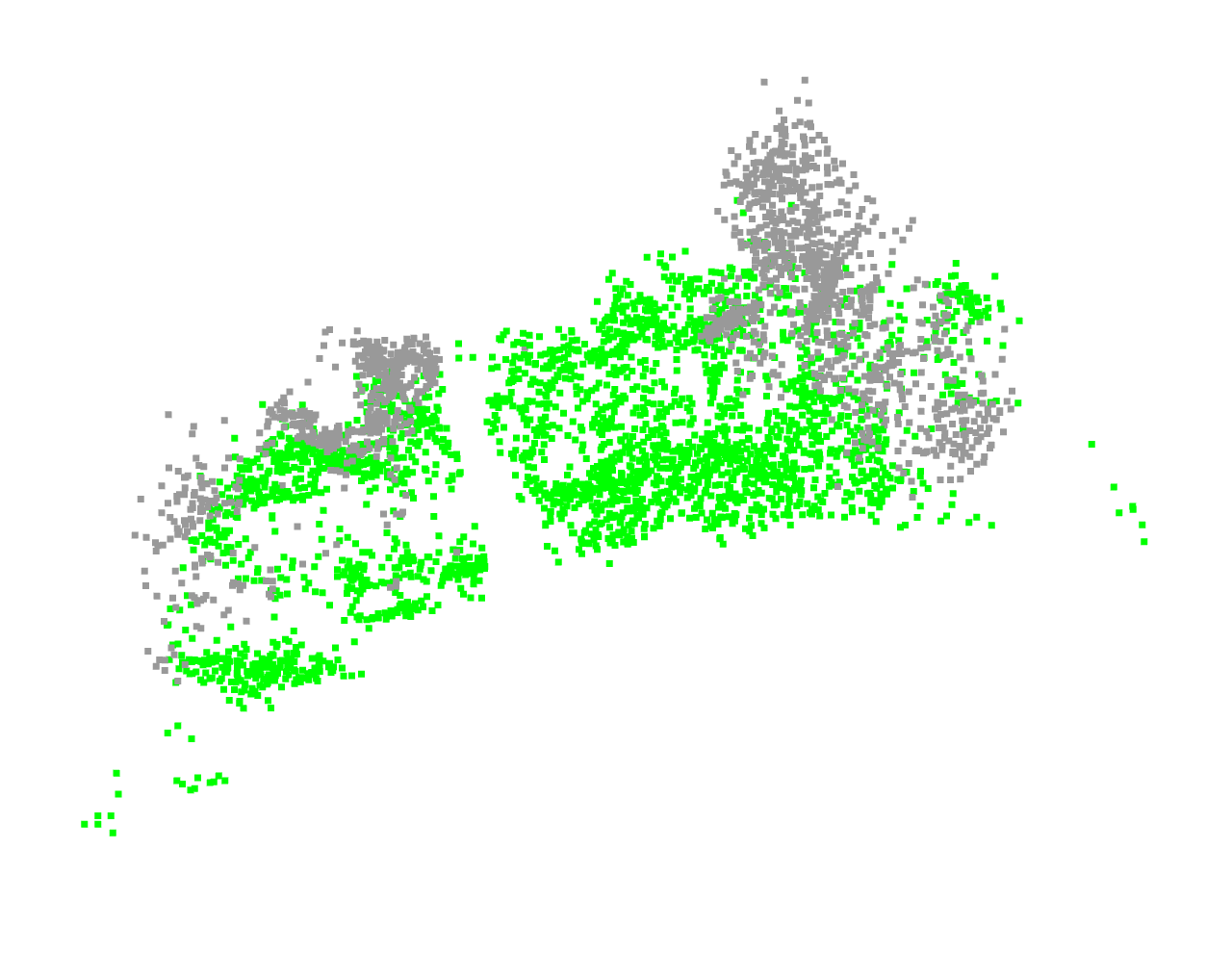}
        & \segimg{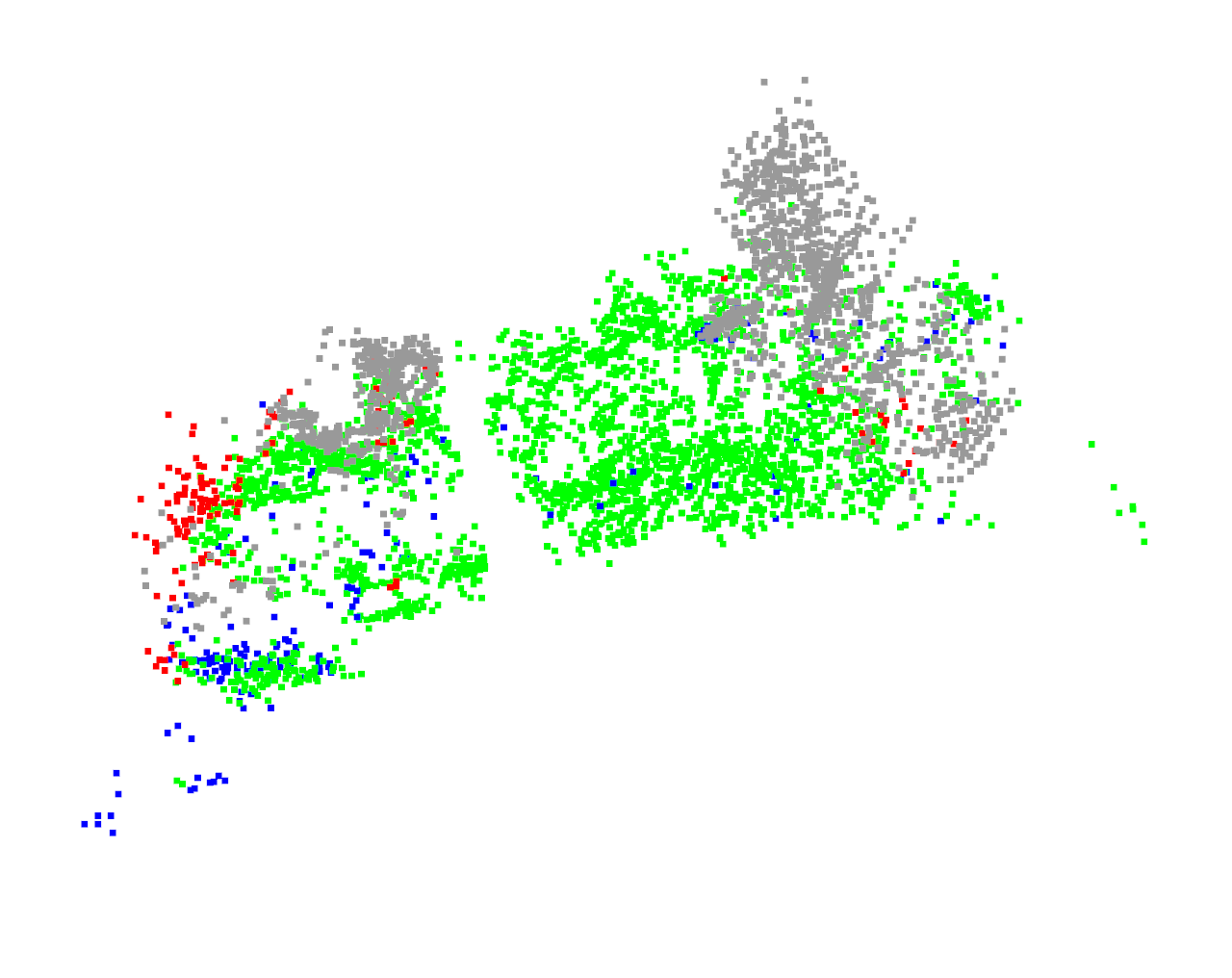}
        & \segimg{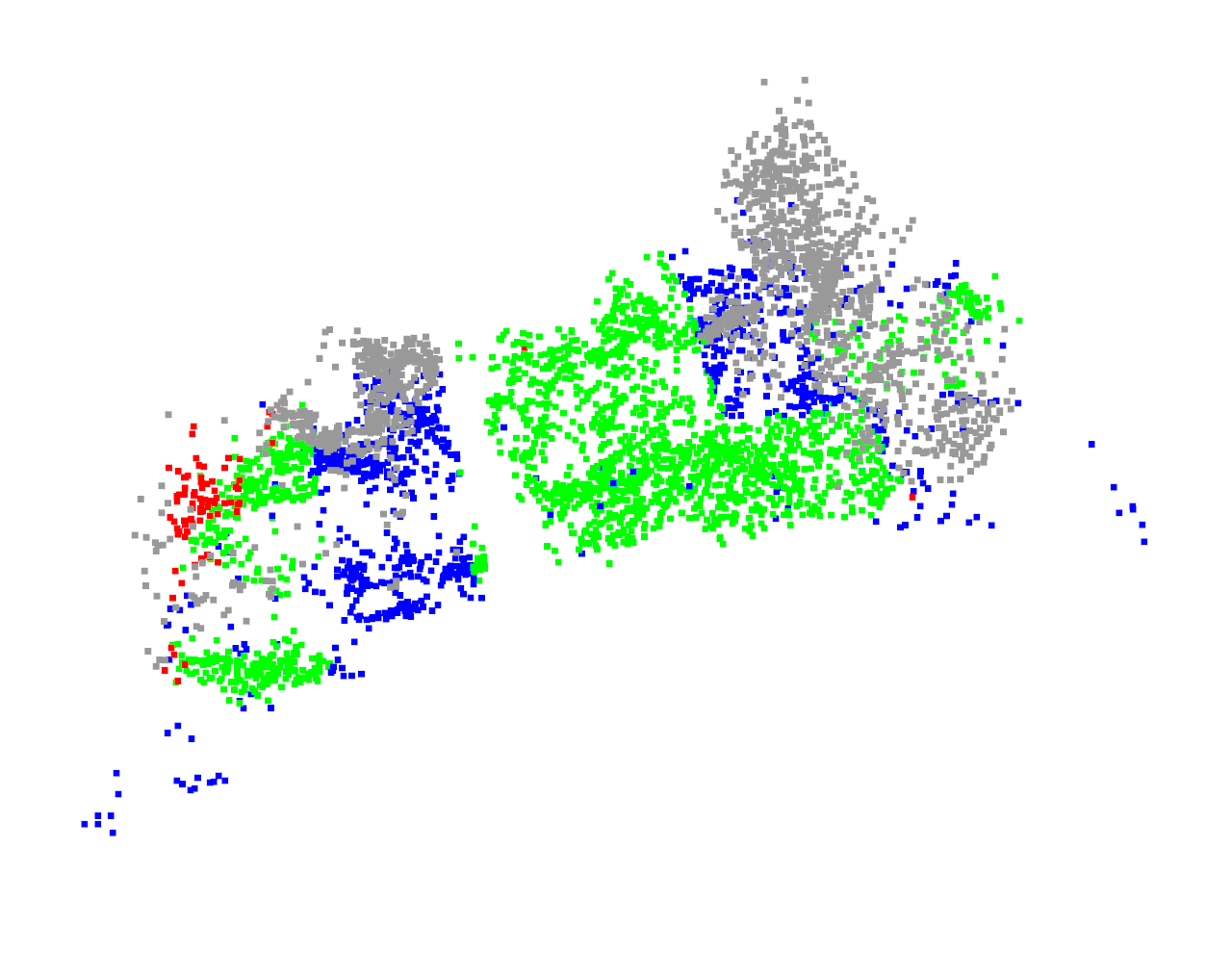}
        & \segimg{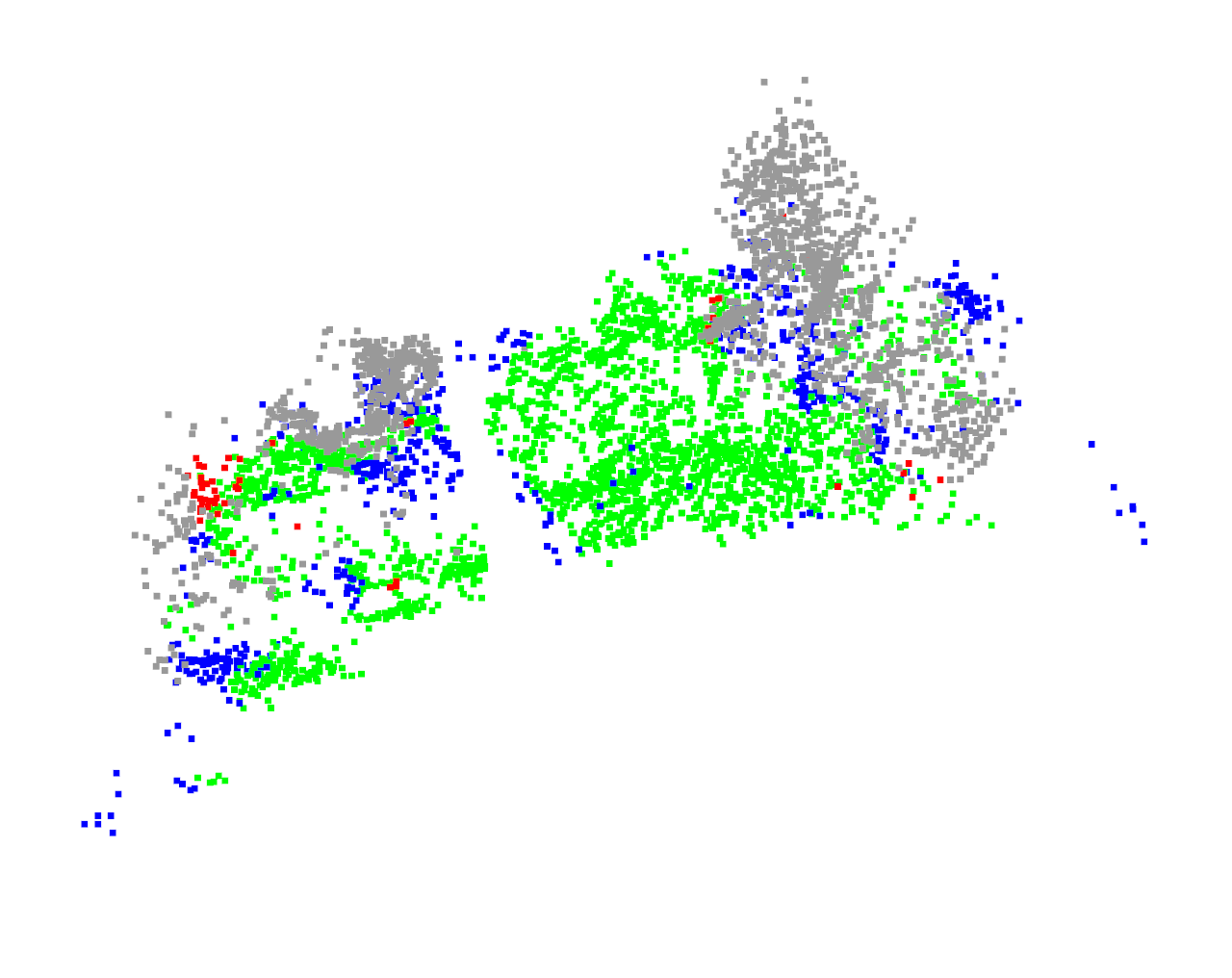}
        & \segimg{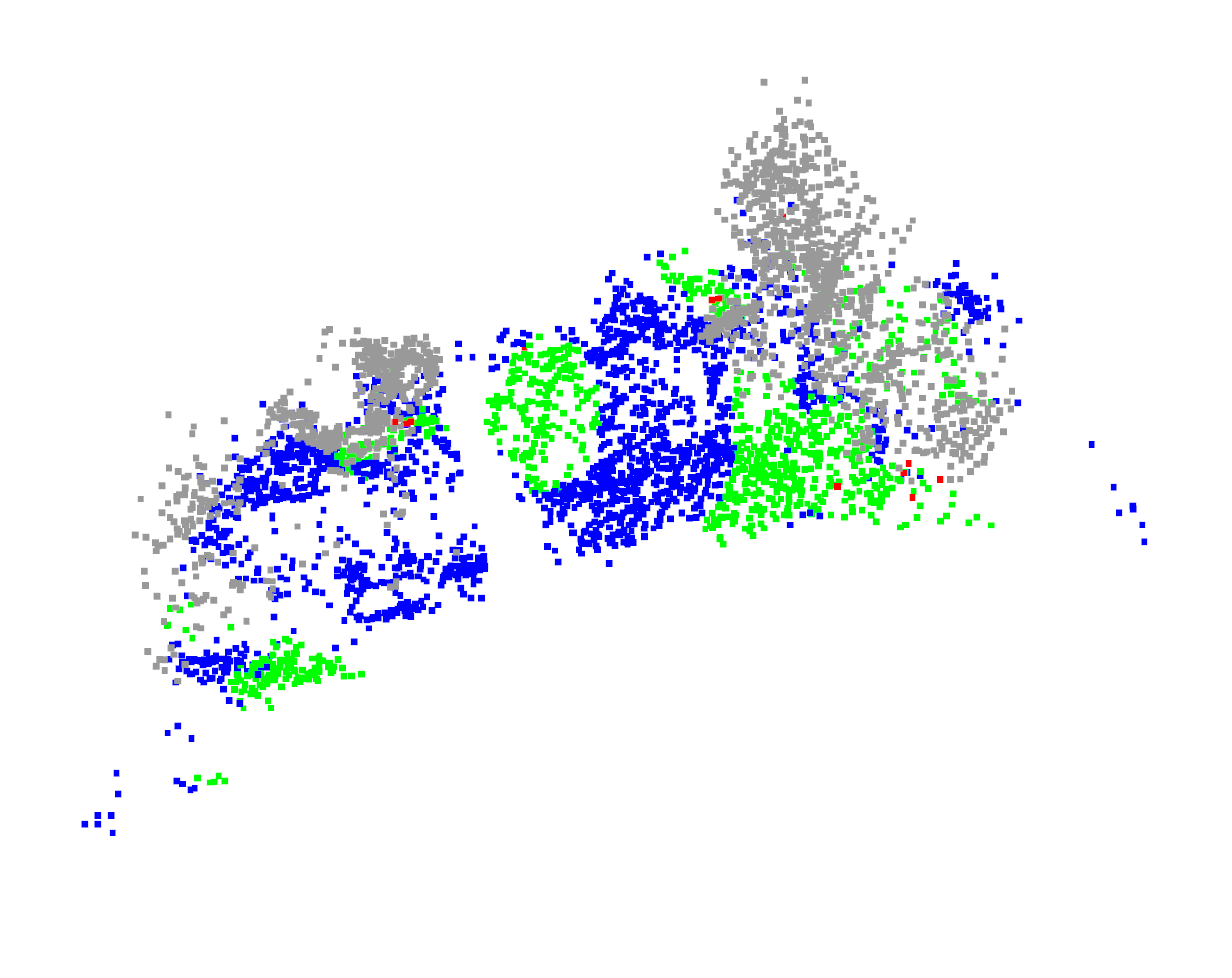}
        & \segimg{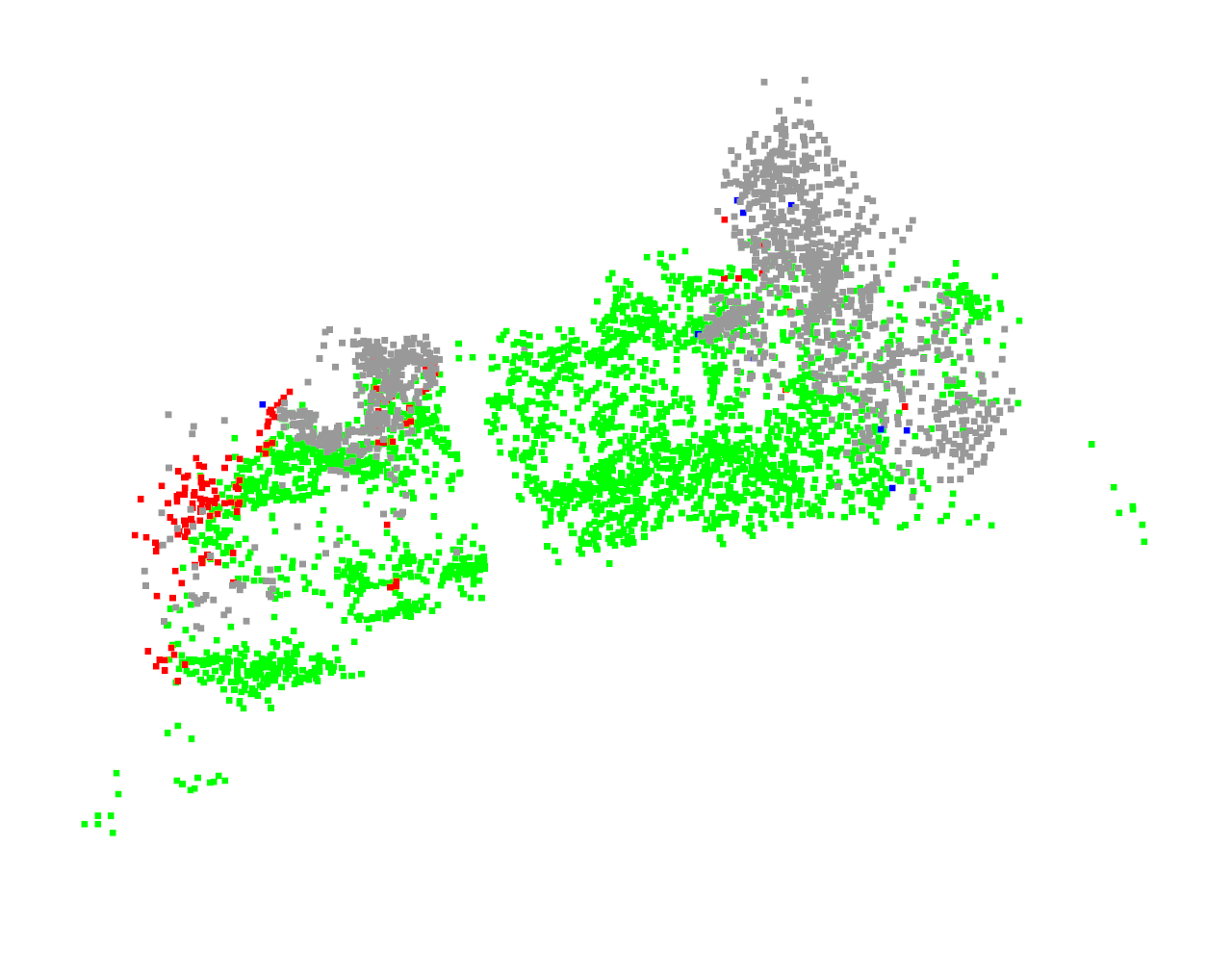} \\

        \seglabel{Tea-4}
        & \segimg{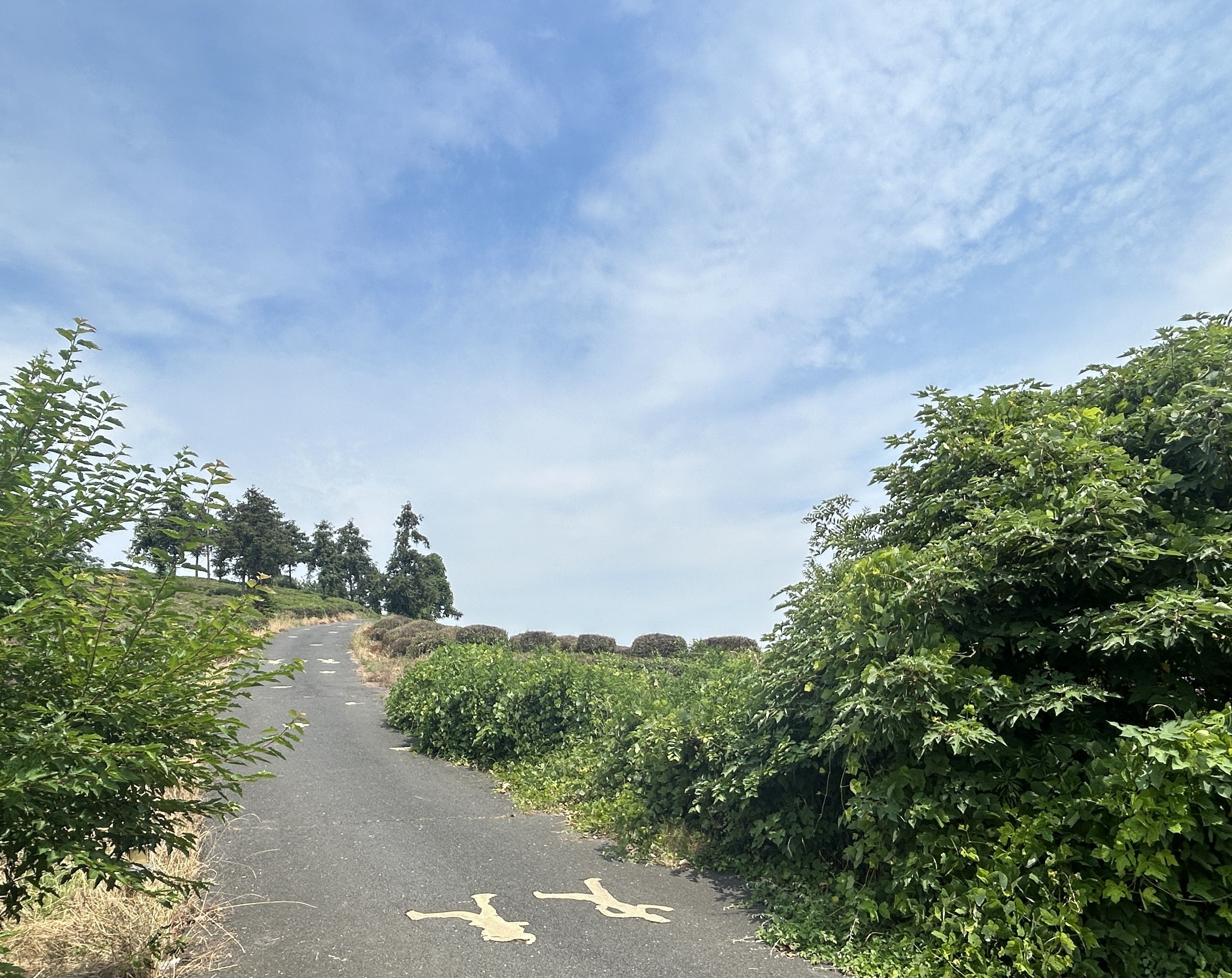}
        & \segimg{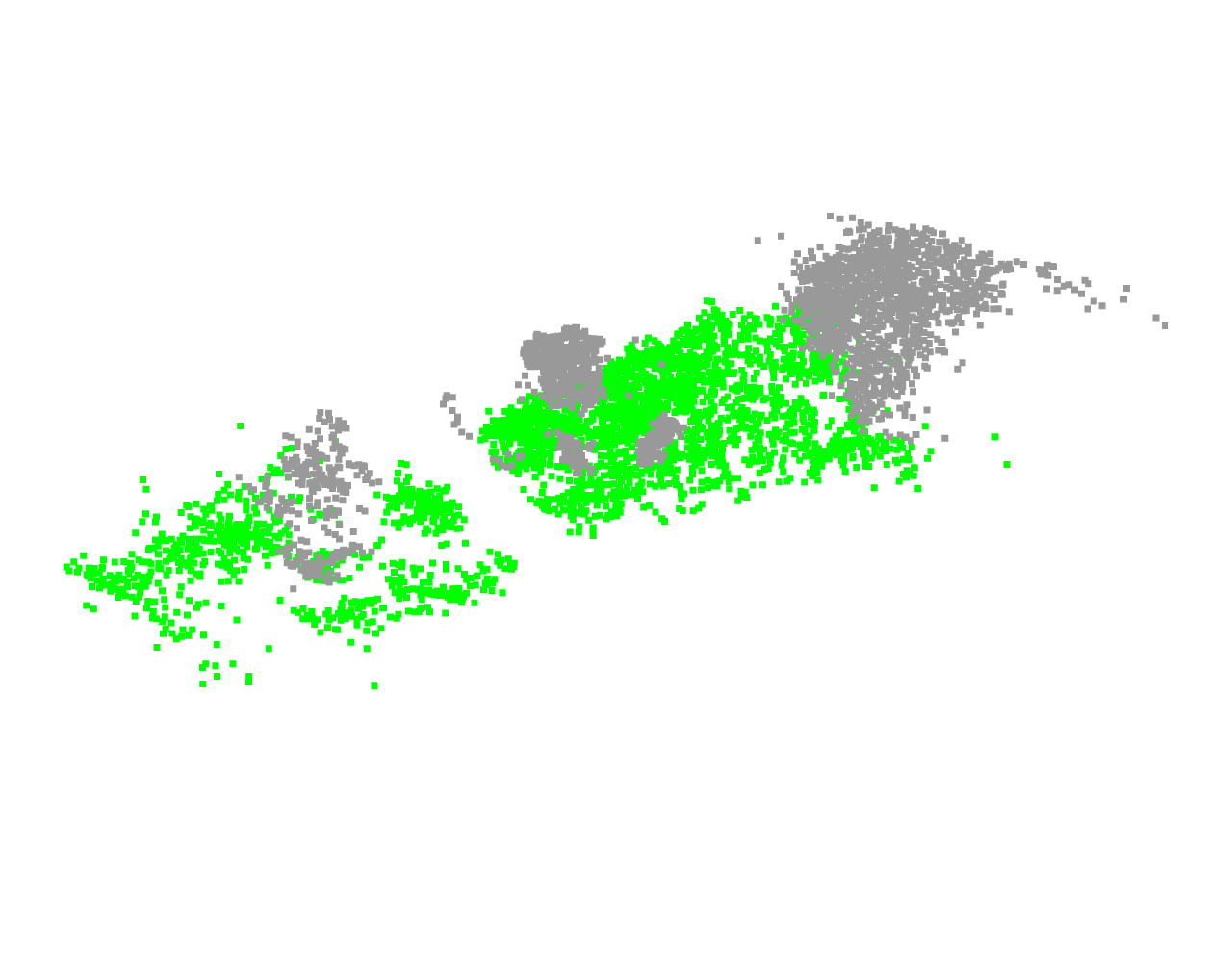}
        & \segimg{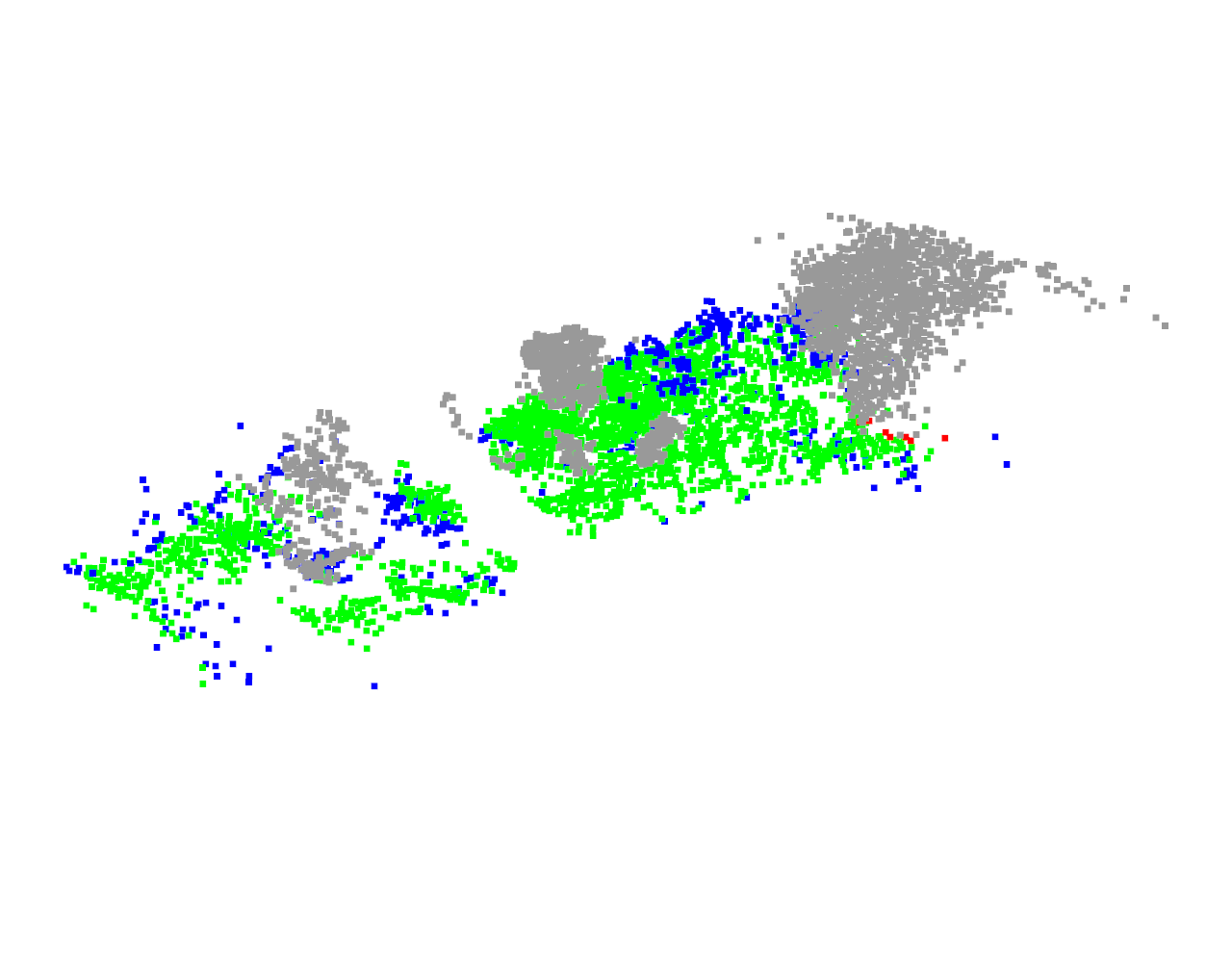}
        & \segimg{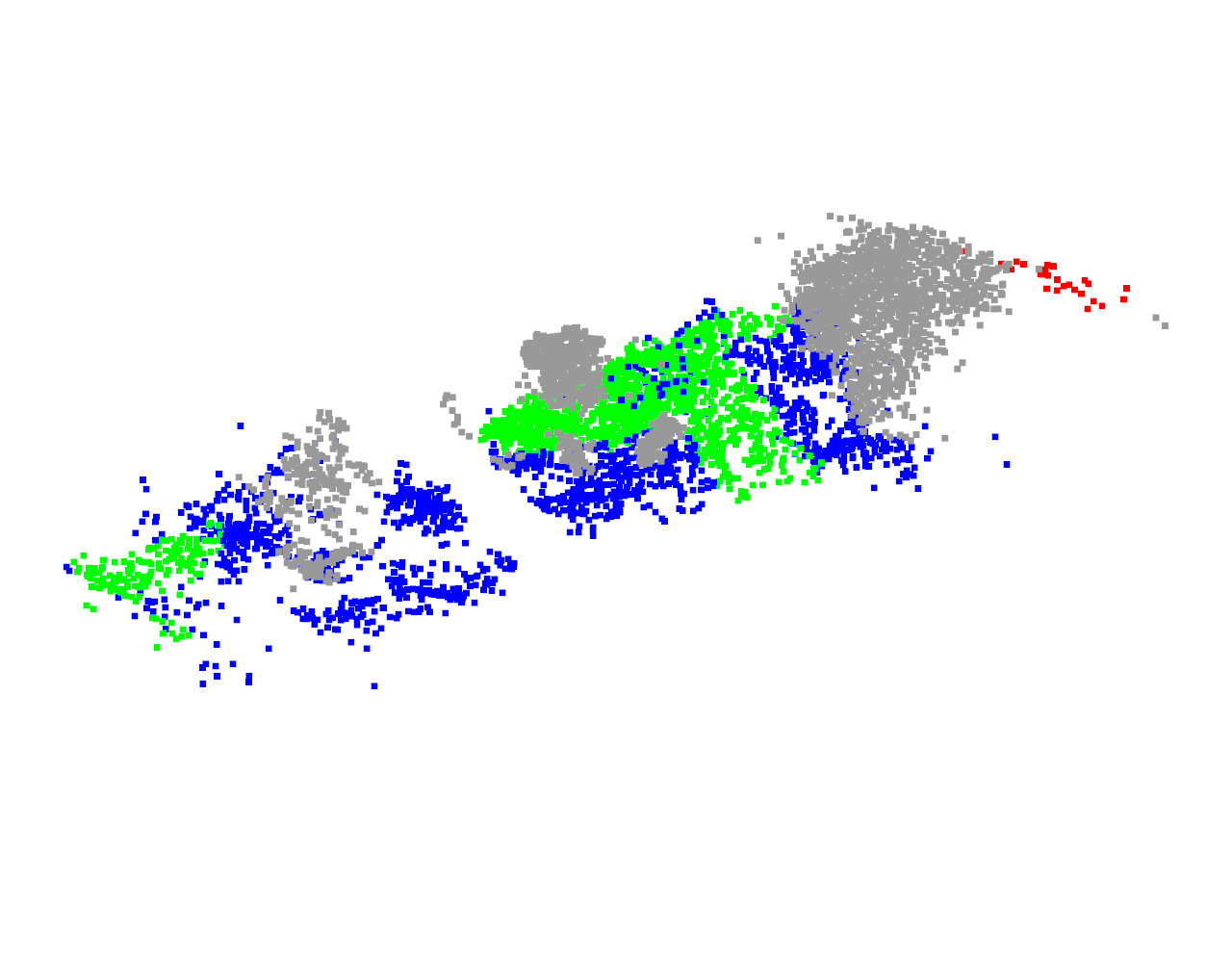}
        & \segimg{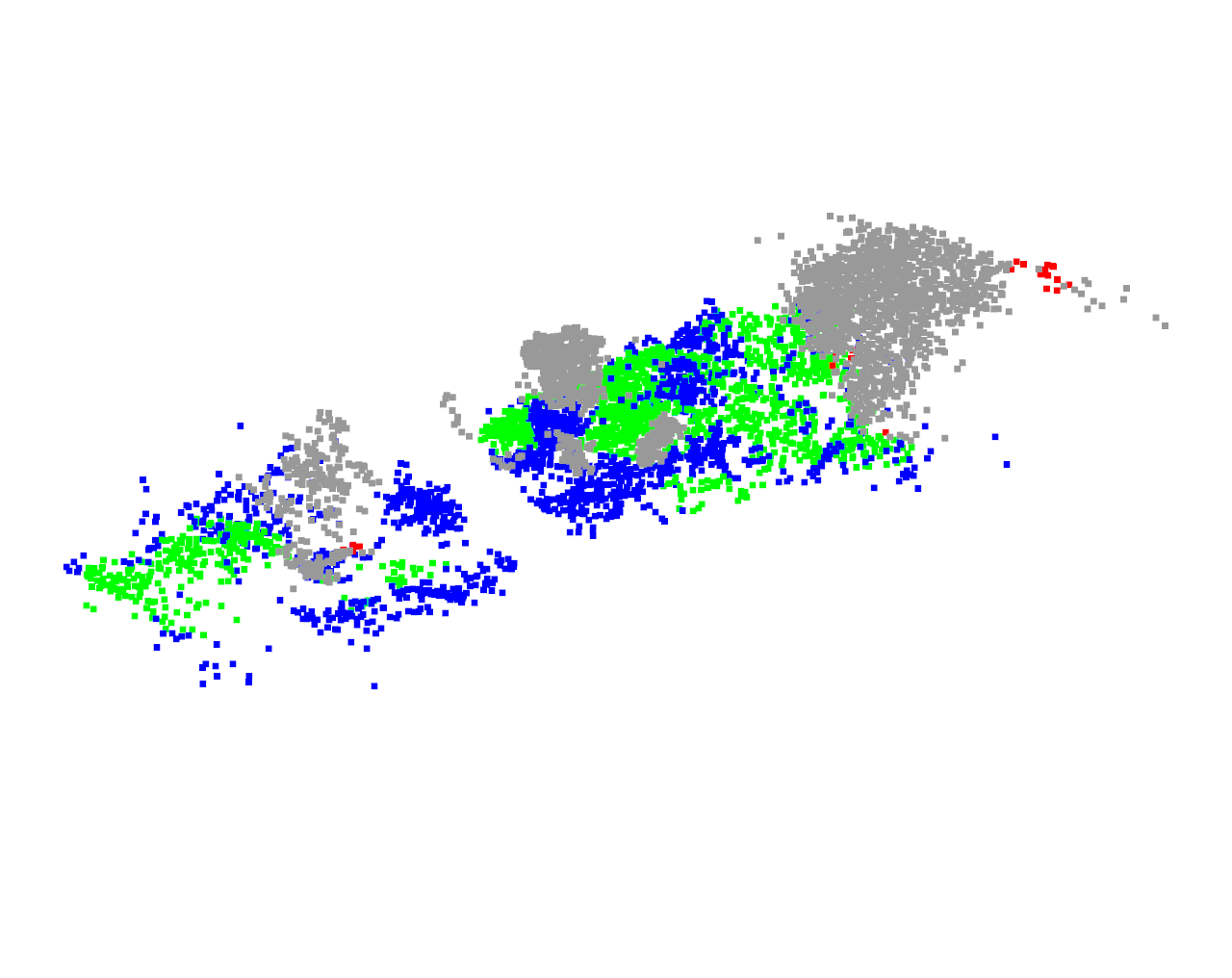}
        & \segimg{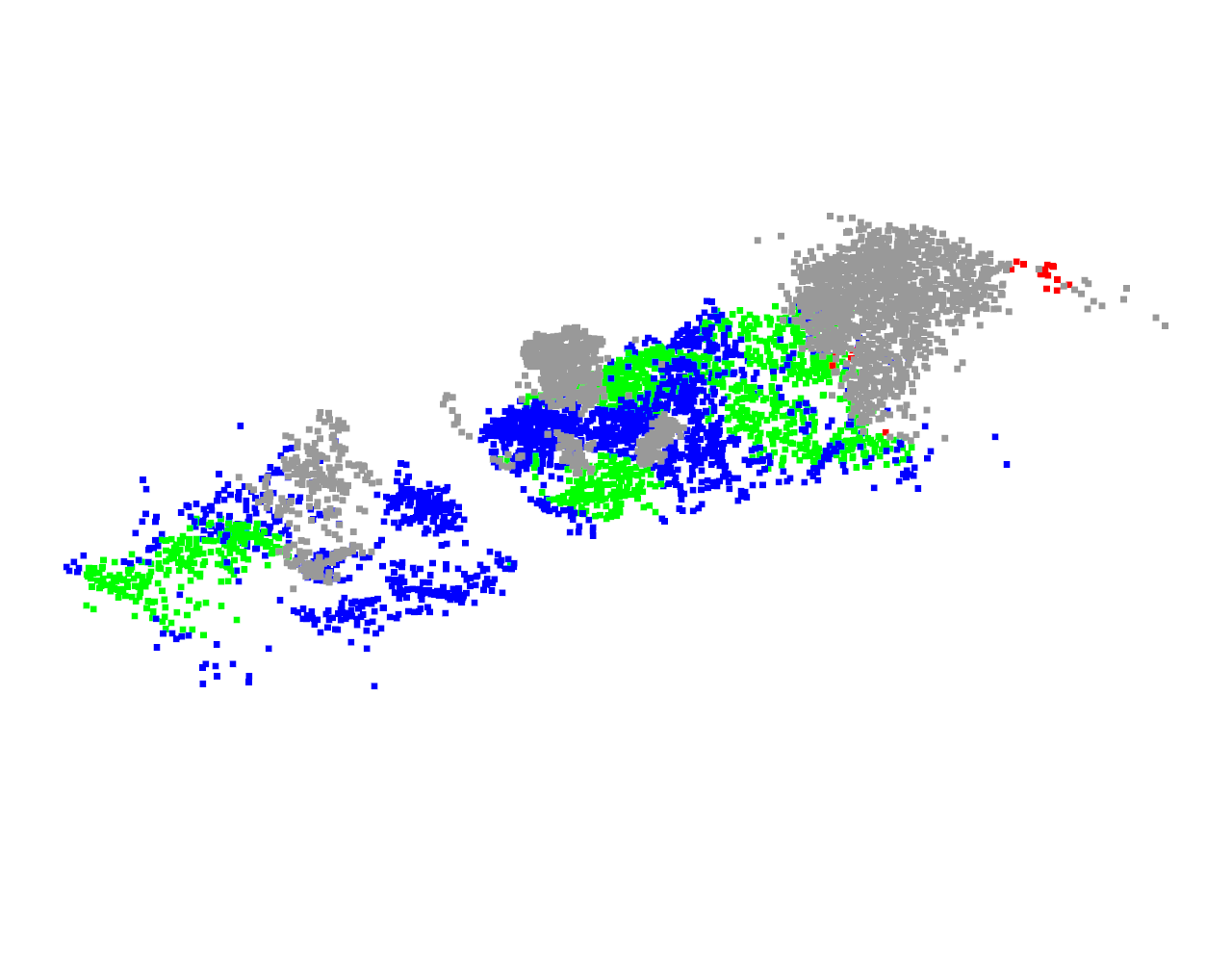}
        & \segimg{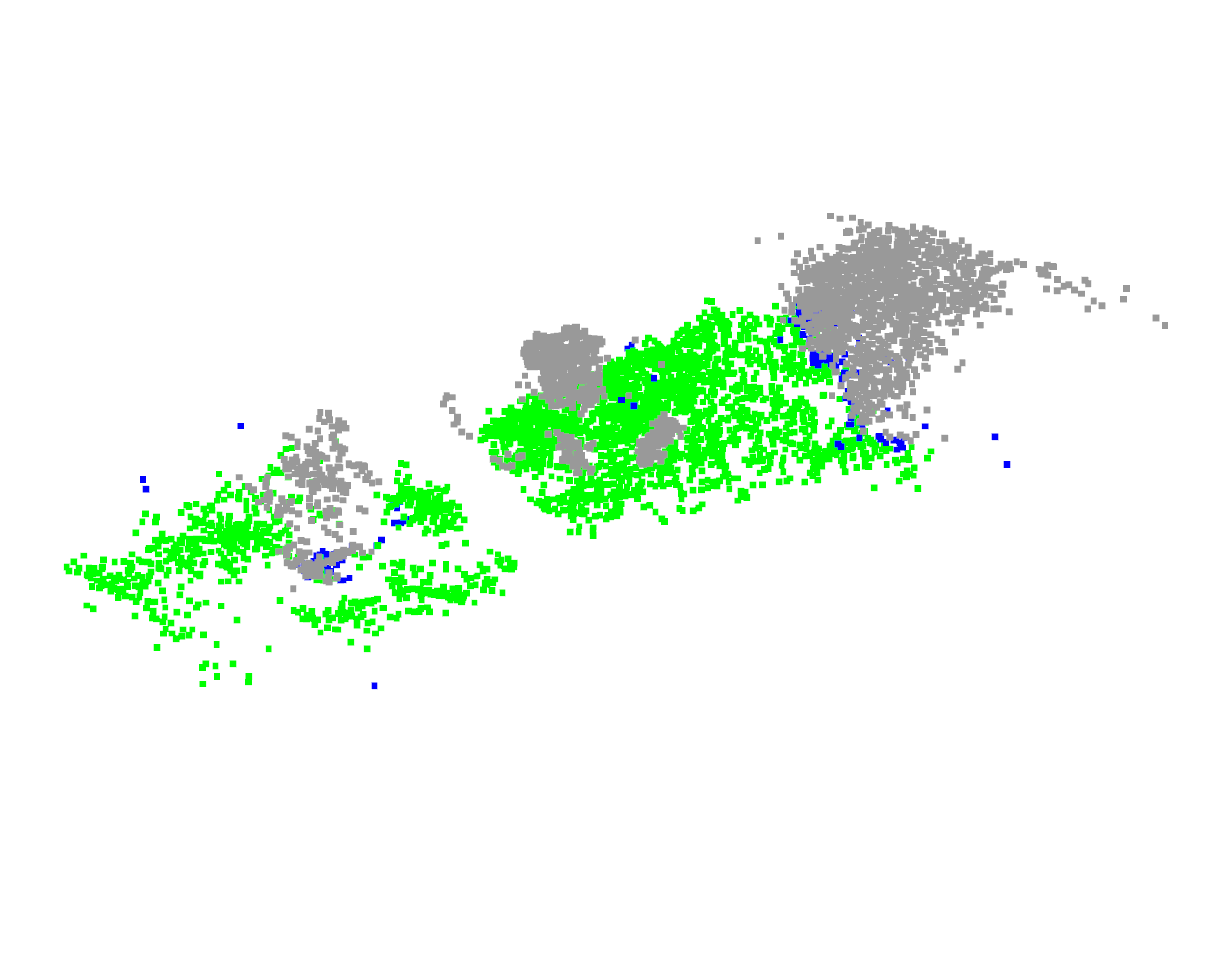} \\
        
        \bottomrule
    \end{tabular}
    \caption{\textbf{Qualitative comparison of different ground segmentation methods across diverse terrain scenarios.}
    The first three rows correspond to the flat field, slope, and water scenes, respectively, whereas the last four rows correspond to tea plantation hill scenes (Tea-1 to Tea-4). True positives (TP), false positives (FP), false negatives (FN), and true negatives (TN) are visualized in \textcolor{green}{green}, \textcolor{red}{red}, \textcolor{blue}{blue}, and \textcolor{gray}{gray}, respectively.
    }
    \label{fig:seg_comparison}
\end{figure*}

To further illustrate the qualitative differences among the compared methods, representative segmentation results are shown in Fig.~\ref{fig:seg_comparison}, including flat field, slope, water, and four tea plantation hill cases (Tea-1 to Tea-4). 
Tea-3 and Tea-4 are collected near the plantation boundary and contain flat road regions.

In both the flat field and slope scenarios, most methods achieve reasonable segmentation results. 
However, the proposed method exhibits better boundary integrity and spatial consistency. 
In more challenging scenarios, especially tea plantation hill, the compared methods frequently suffer from incomplete ground extraction or severe misclassification. 
In contrast, the proposed method more effectively suppresses false positives and false negatives, resulting in more spatially coherent ground extraction in challenging regions.
The results on Tea-3 and Tea-4 further show that the proposed method remains robust in mixed scenes containing both terraced terrain and flat roads.

The water scenario is particularly challenging for mmWave radar due to strong reflections and multi-path effects, which may introduce a large number of spurious observations below the actual water surface. 
Under such conditions, the compared methods are more easily affected by false detections or local failure, whereas the proposed method remains more robust by suppressing a substantial portion of interference-induced outliers and preserving more consistent ground extraction.

A small number of false negatives still remain in some local regions. 
This is mainly caused by the inherent sparsity of mmWave point clouds and the limited adaptability of globally fixed thresholds to locally varying terrain geometry. 
Nevertheless, the proposed method maintains more robust and consistent performance across diverse agricultural environments.

\subsubsection{Runtime Analysis}
\begin{table*}[t]
\centering
\caption{\textbf{Runtime comparison of ground segmentation methods on the RK3588 platform.}
Latency is reported in milliseconds per frame, and FPS denotes frames per second. \textbf{Bold} values indicate the best performance.}
\label{tab:runtime}

\begin{tabular}{l cc cc cc cc cc}
\toprule
\multirow{2}{*}{\textbf{Method}} 
& \multicolumn{2}{c}{\textbf{Flat Field}} 
& \multicolumn{2}{c}{\textbf{Slope}} 
& \multicolumn{2}{c}{\textbf{Water}} 
& \multicolumn{2}{c}{\textbf{Tea Plantation Hill}} 
& \multicolumn{2}{c}{\textbf{Overall}} \\
\cmidrule(lr){2-3} \cmidrule(lr){4-5} \cmidrule(lr){6-7} \cmidrule(lr){8-9} \cmidrule(lr){10-11}
& Latency & FPS & Latency & FPS & Latency & FPS & Latency & FPS & Latency & FPS \\
\midrule

RANSAC-Single 
& 0.433 & 2307 & 0.680 & 1470 & 0.101 & 9881 & 0.777 & 1287 & 0.540 & 1851 \\

RANSAC-Patch  
& 0.719 & 1391 & 1.116 & 896 & 0.139 & 7207 & 0.921 & 1086 & 0.790 & 1265 \\

Patchwork     
& 1.270 & 787  & 1.736 & 576 & 0.306 & 3268 & 1.501 & 666  & 1.310 & 763 \\

Patchwork++   
& 1.266 & 790  & 1.804 & 554 & 0.271 & 3683 & 1.530 & 653  & 1.329 & 752 \\

\textbf{Proposed} 
& \textbf{0.422} & \textbf{2370} 
& \textbf{0.667} & \textbf{1499} 
& \textbf{0.093} & \textbf{10709} 
& \textbf{0.578} & \textbf{1731} 
& \textbf{0.491} & \textbf{2036} \\

\bottomrule
\end{tabular}

\end{table*}

To evaluate the onboard efficiency of different ground segmentation methods, runtime performance is measured on the RK3588 embedded computing platform used in the UAV system. 
For each method, the average latency per frame and the corresponding processing frequency are reported across different scenarios, as shown in Table~\ref{tab:runtime}.

The proposed method achieves the lowest overall latency of 0.491~ms and the highest overall processing frequency of 2036~FPS, indicating that it can support real-time onboard terrain perception.
Compared with Patchwork and Patchwork++, the proposed method is substantially faster because it avoids sensor-centric concentric-zone processing and relies on lightweight grid-based partitioning and PCA-based local plane estimation. 
Compared with RANSAC-Patch, the proposed method also achieves lower overall latency while providing higher segmentation accuracy, demonstrating a better balance between computational efficiency and segmentation robustness.

This runtime advantage is important for agricultural UAV deployment, where the terrain perception module must operate together with state estimation, mapping, and flight control modules under limited onboard computational resources.

\subsubsection{Ablation Study}
\begin{table}[t]
\centering
\caption{\textbf{Ablation study of the proposed ground segmentation method.}
TI, PSI, and Ref denote temporal integration, prior-based seed initialization, and ground segmentation refinement, respectively. 
All metrics are reported in percentage. \textbf{Bold} values indicate the best performance, and \underline{underlined} values indicate the second-best performance.
}
\label{tab:ablation}

\begin{tabular}{c c c c cccc}
\toprule
\textbf{Baseline} & \textbf{TI} & \textbf{PSI} & \textbf{Ref} 
& \textbf{Prec.} & \textbf{Recall} & \textbf{IoU} & \textbf{F1} \\
\midrule
\checkmark &  &  &  
& 98.14 & 79.79 & 78.60 & 88.02 \\

\checkmark & \checkmark &  &  
& 97.82 & 81.65 & 80.19 & 89.01 \\

\checkmark & \checkmark & \checkmark &  
& \textbf{99.42} & 84.59 & 84.17 & 91.41 \\

\checkmark & \checkmark &  & \checkmark 
& 98.91 & \underline{88.98} & \underline{88.12} & \underline{93.68} \\

\checkmark & \checkmark & \checkmark & \checkmark 
& \underline{99.27} & \textbf{90.03} & \textbf{89.44} & \textbf{94.42} \\
\bottomrule
\end{tabular}

\end{table}

\begin{figure*}[htbp]
    \centering
    \setlength{\tabcolsep}{1.2pt}
    \renewcommand{\arraystretch}{1.0}

    \newcommand{\ablimg}[2][0.155\textwidth]{%
        \includegraphics[width=#1,valign=m]{#2}%
    }

    \newcommand{\caselabel}[1]{%
        \adjustbox{valign=m}{%
            \rotatebox[origin=c]{90}{%
                \makebox[3.2cm][c]{\textbf{#1}}%
            }%
        }%
    }
    
    \begin{tabular}{c c c c c c c c}
        \toprule
        & \textbf{Scene}
        & \textbf{Ground-Truth}
        & \textbf{Baseline} 
        & \textbf{Baseline + TI} 
        & \textbf{Baseline + TI + PSI} 
        & \textbf{Proposed} \\
        
        \midrule
        \caselabel{Case1}
        & \ablimg{fig/seg_flat_crop.jpg}
        & \ablimg{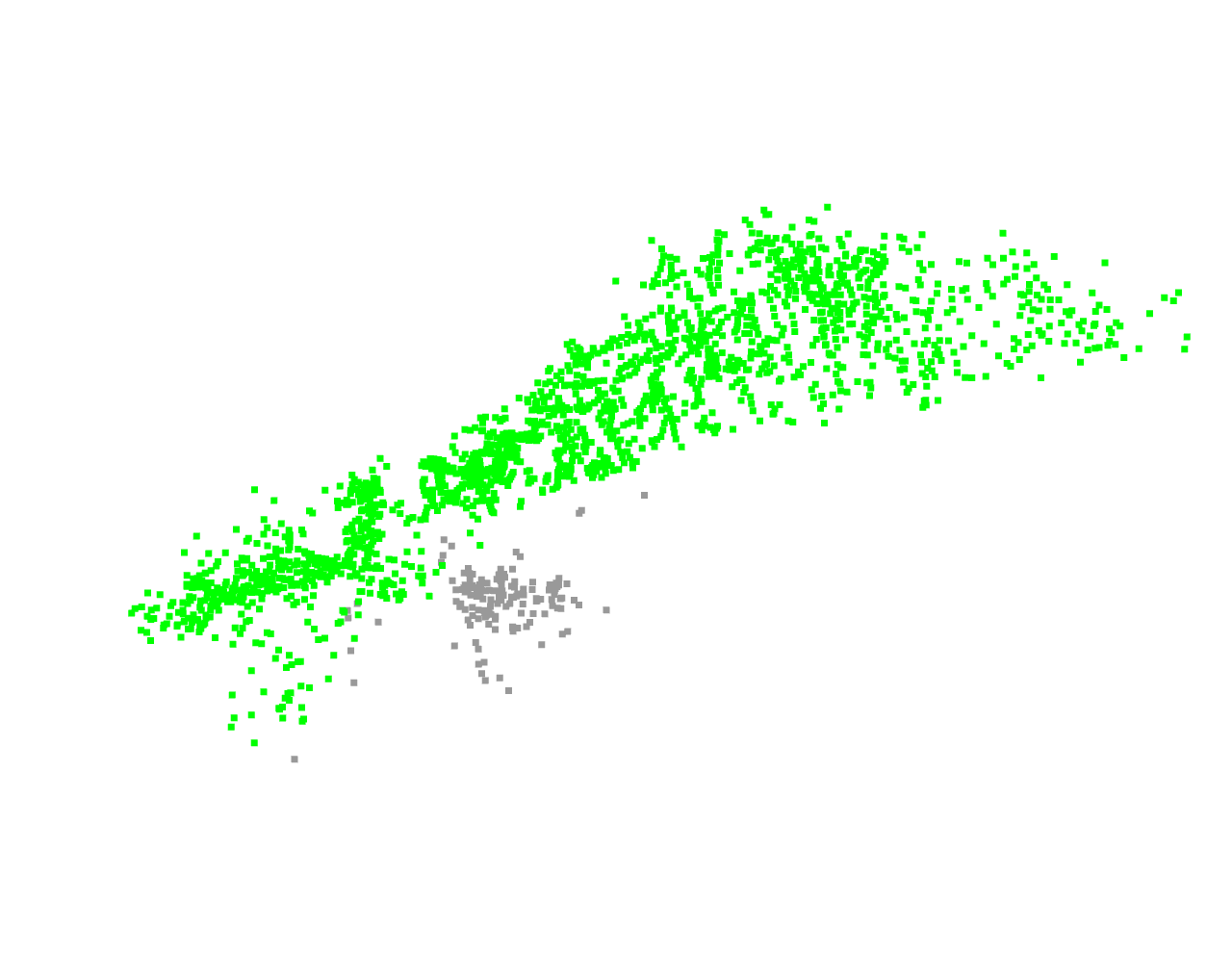}
        & \ablimg{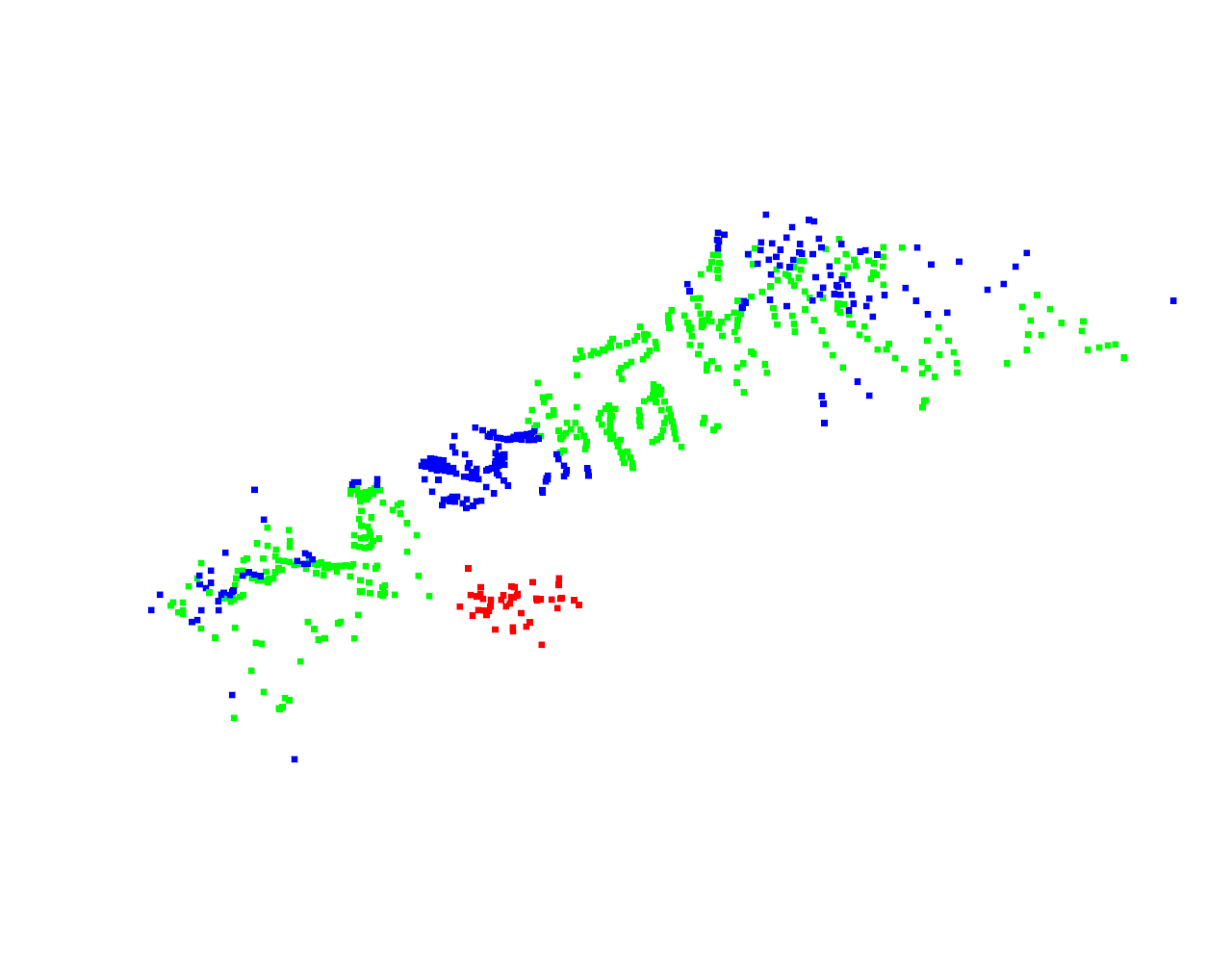}
        & \ablimg{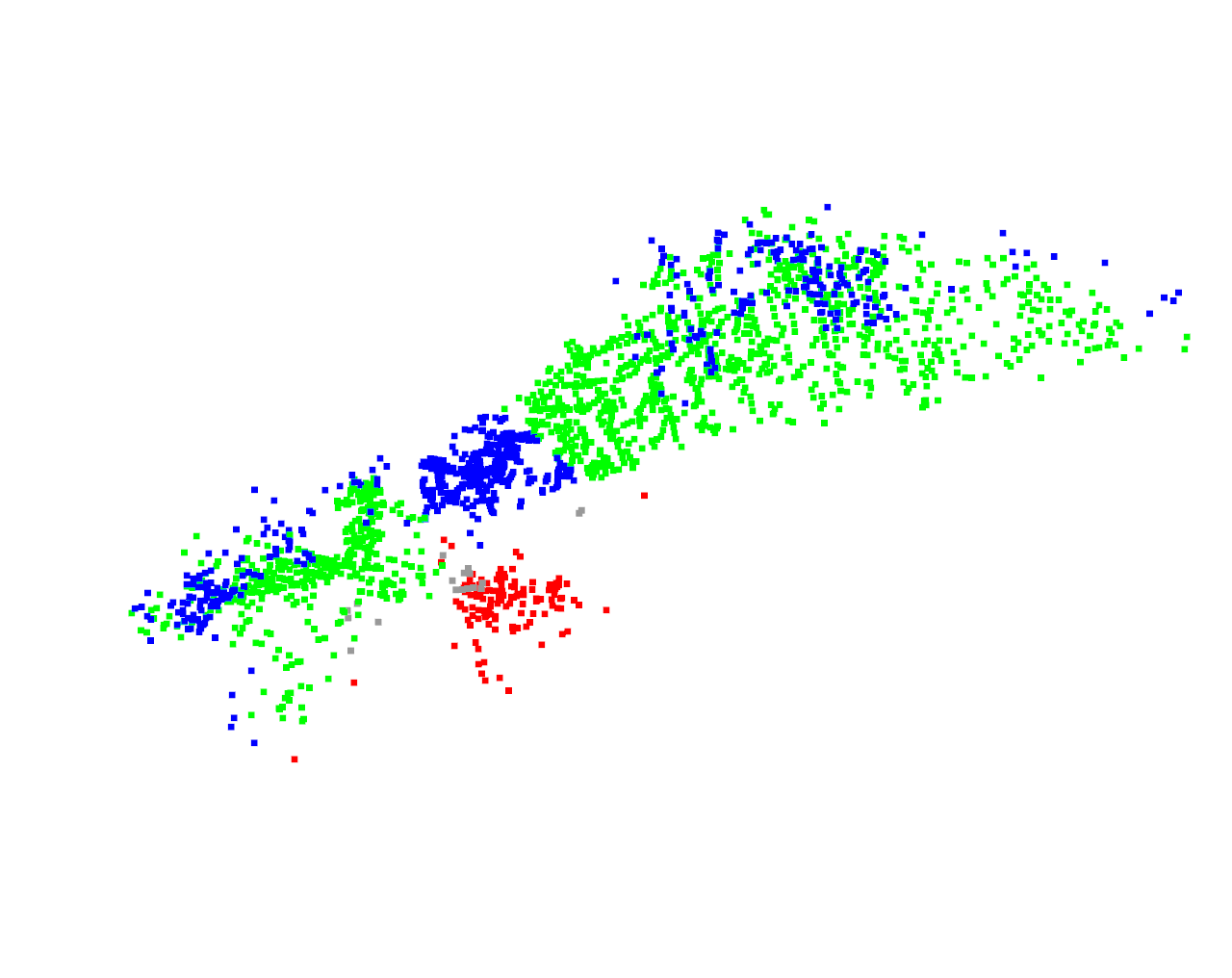}
        & \ablimg{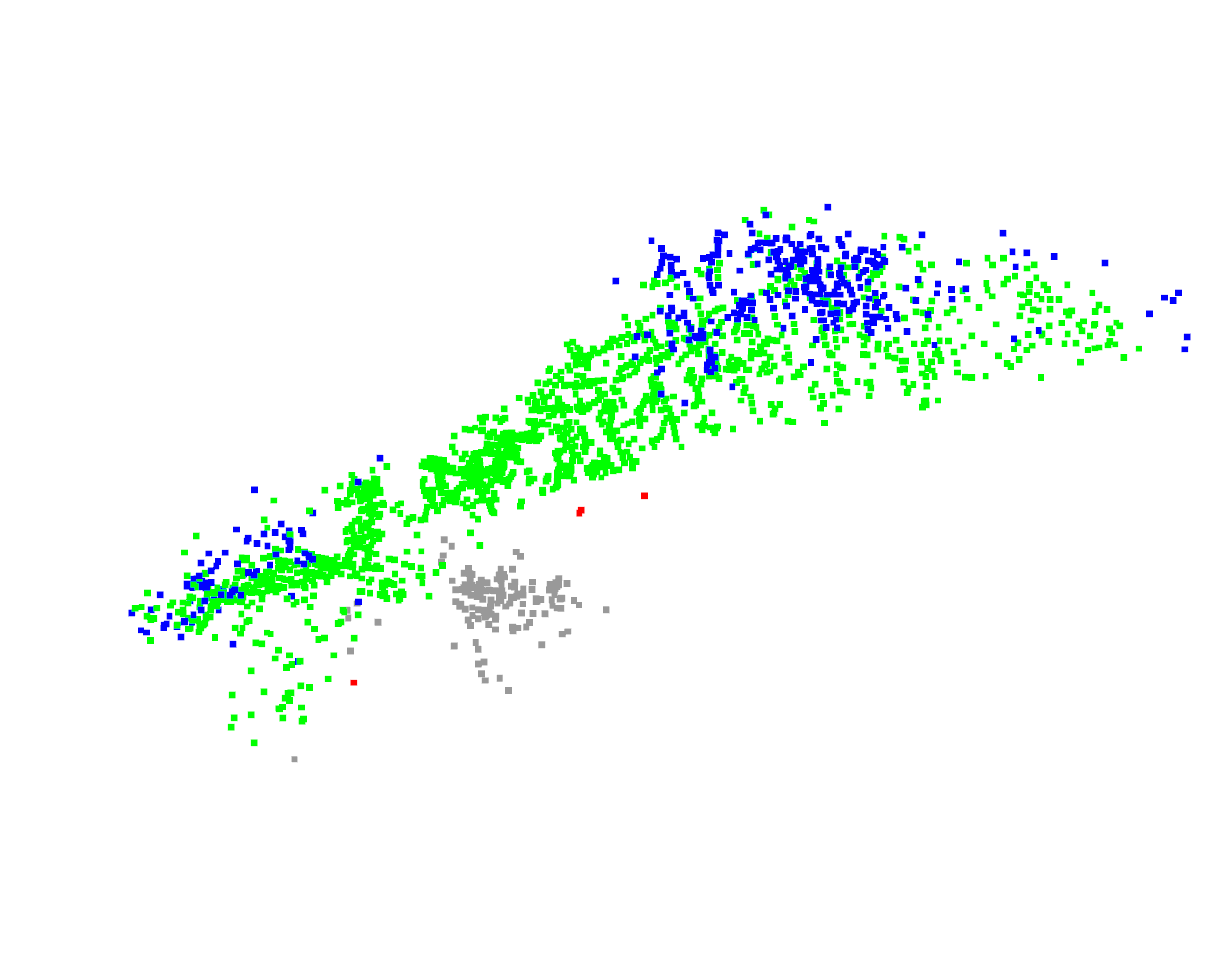}
        & \ablimg{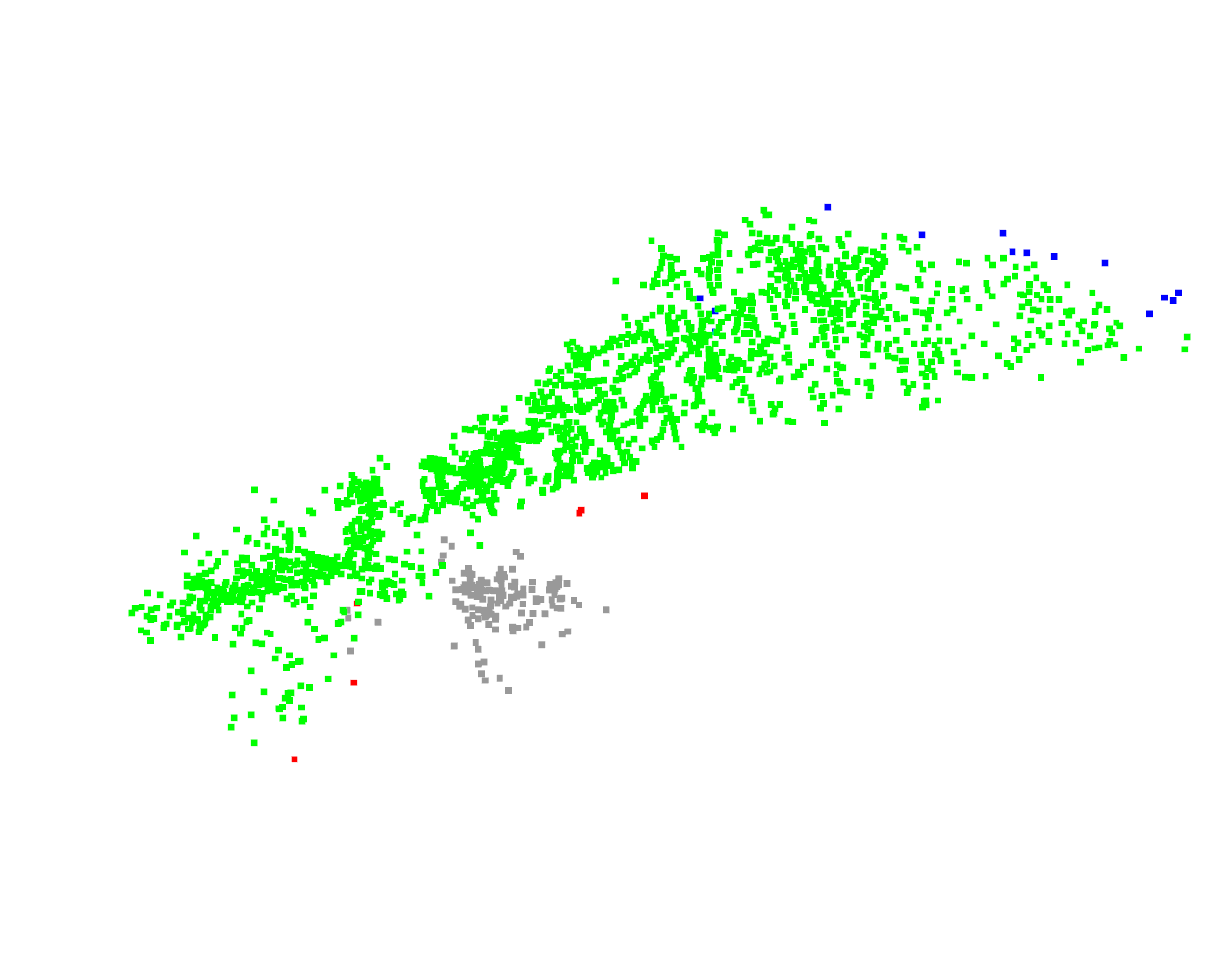} \\

        \caselabel{Case2}
        & \ablimg{fig/seg_tea_1_crop.jpg}
        & \ablimg{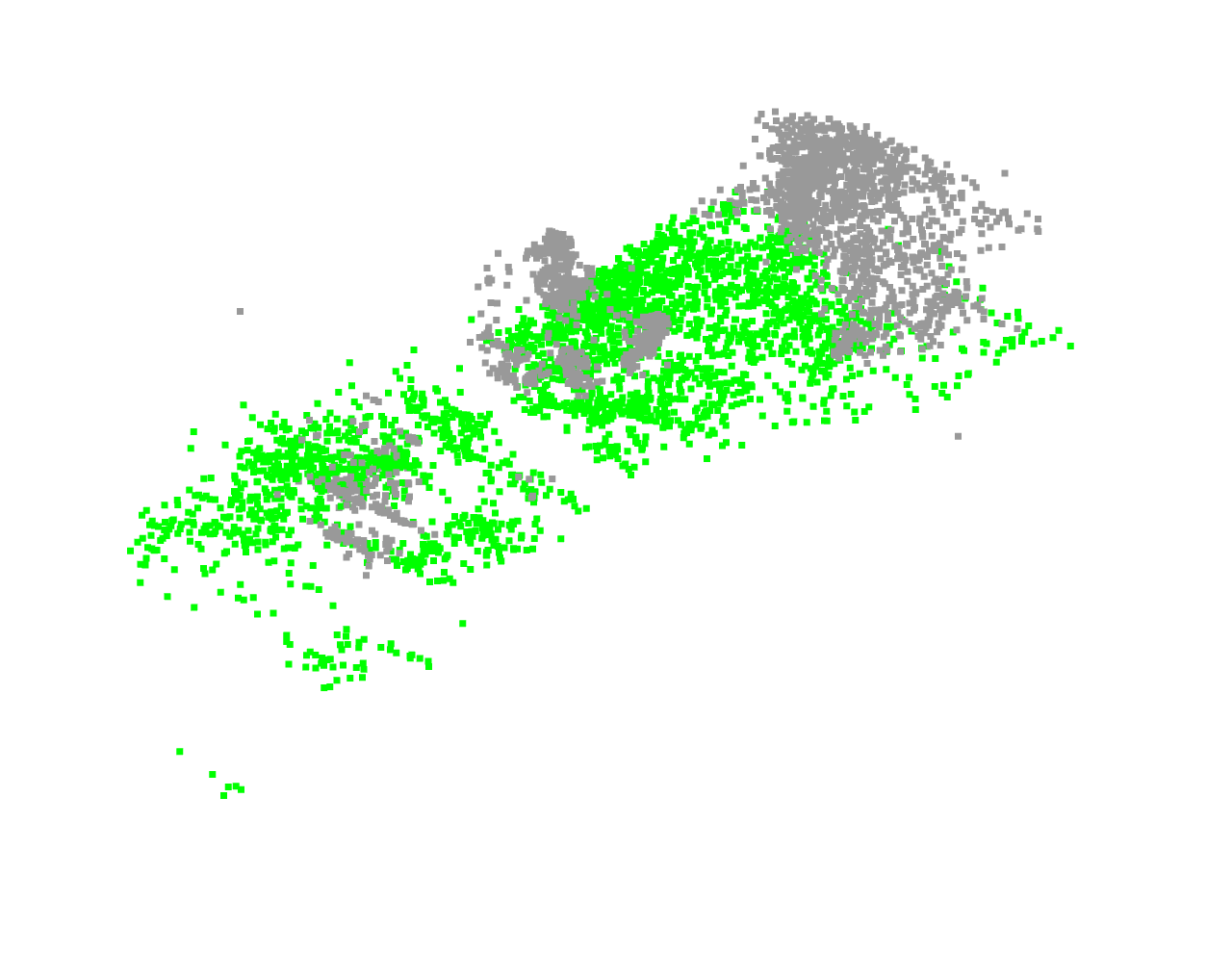} 
        & \ablimg{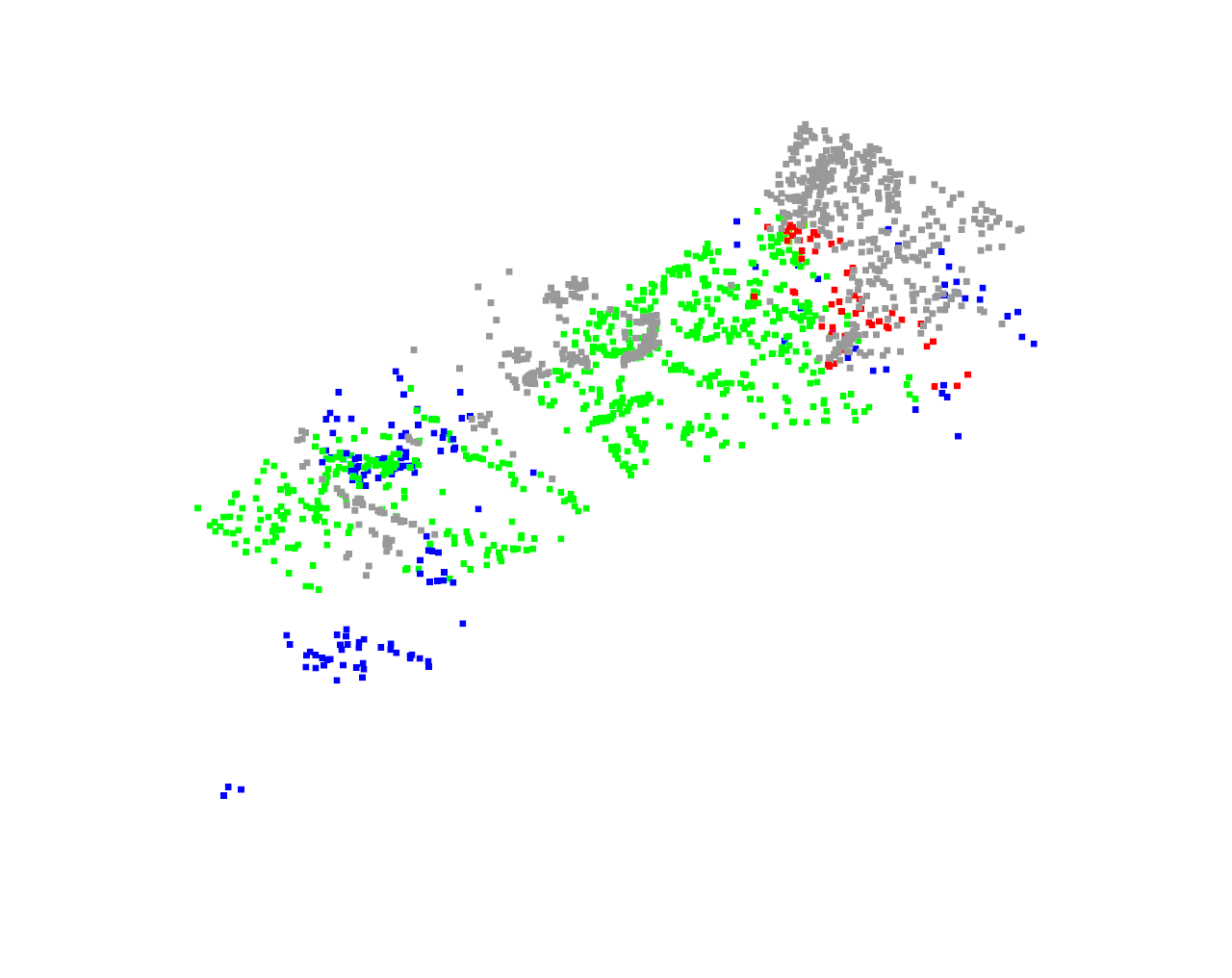}
        & \ablimg{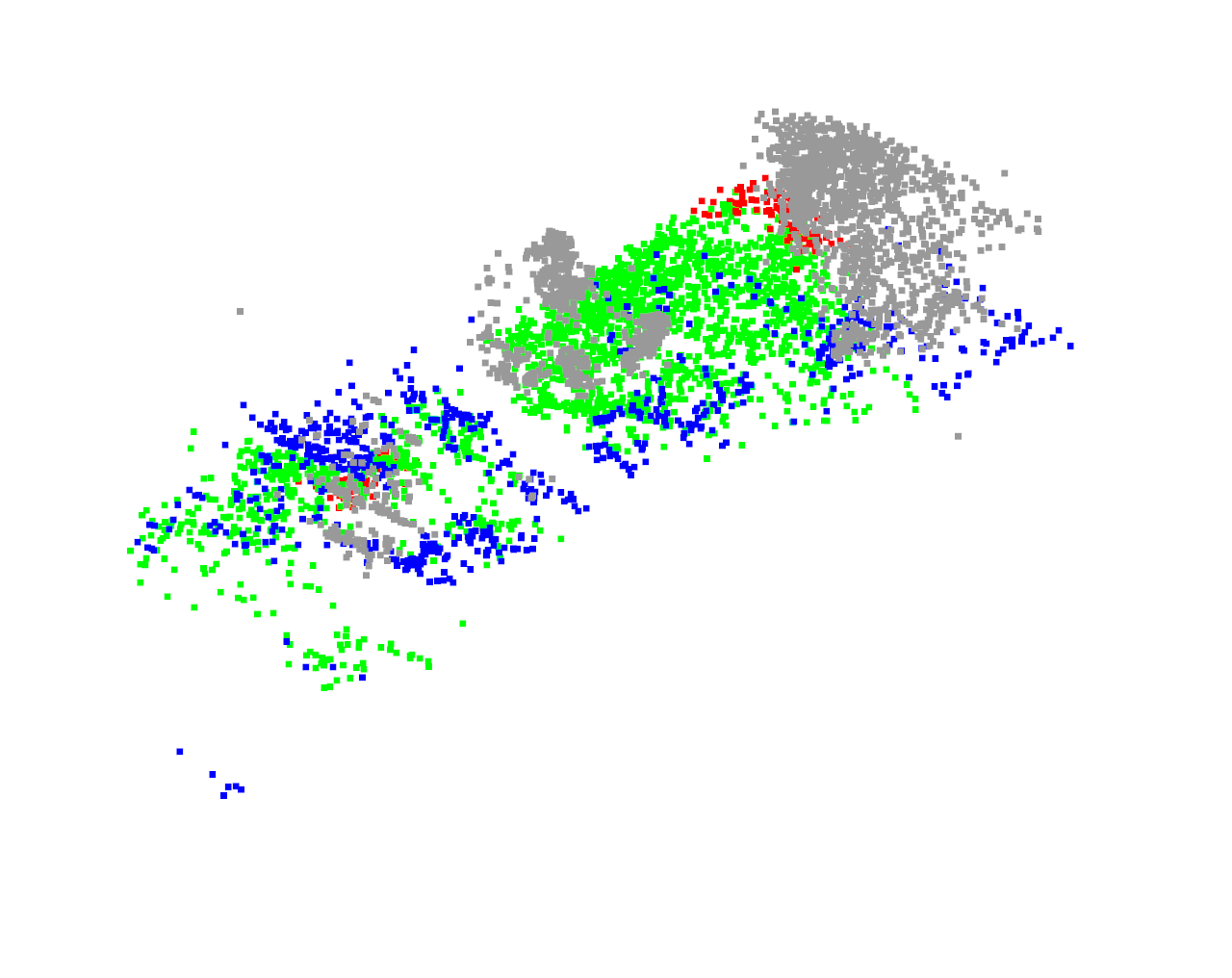}
        & \ablimg{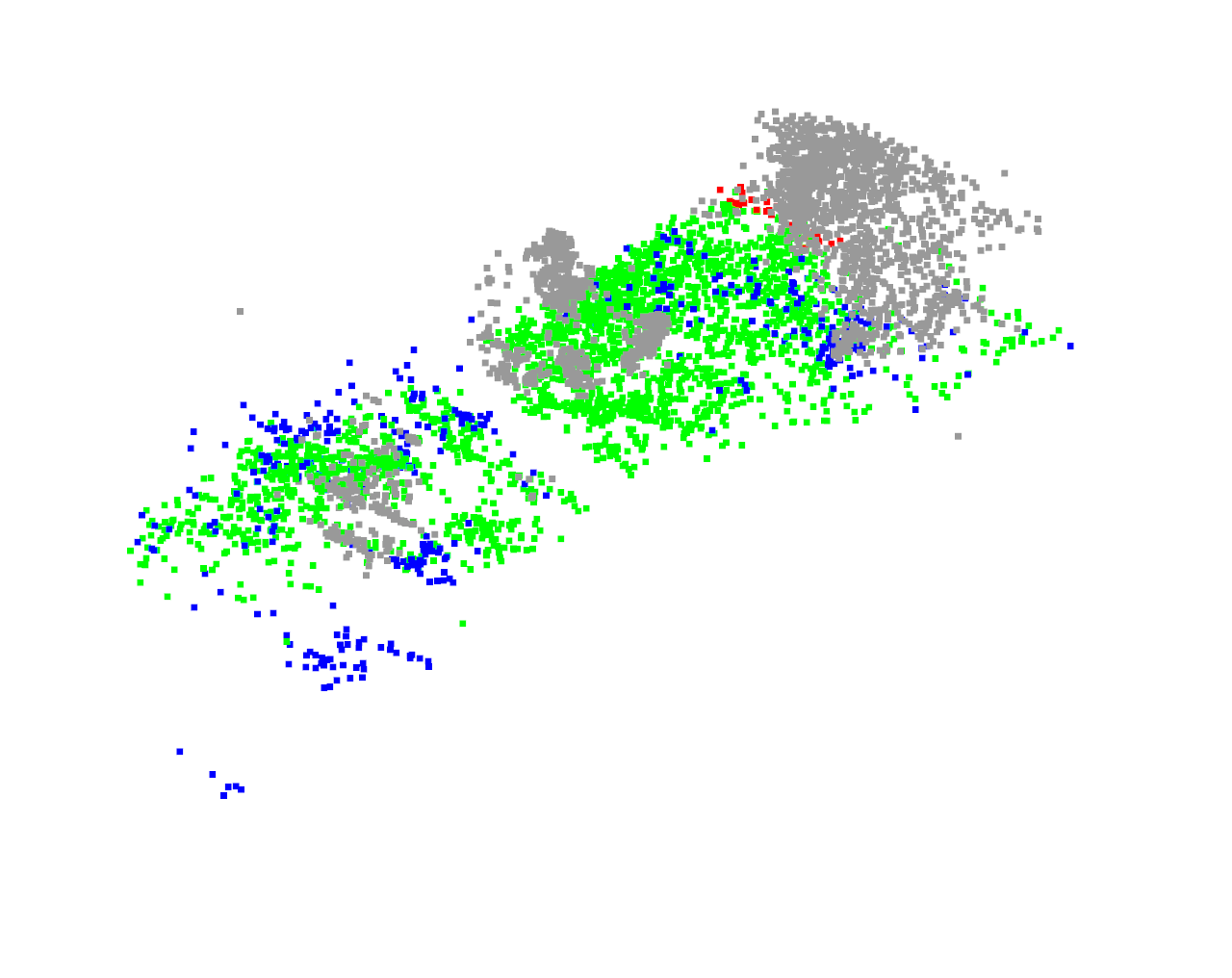}
        & \ablimg{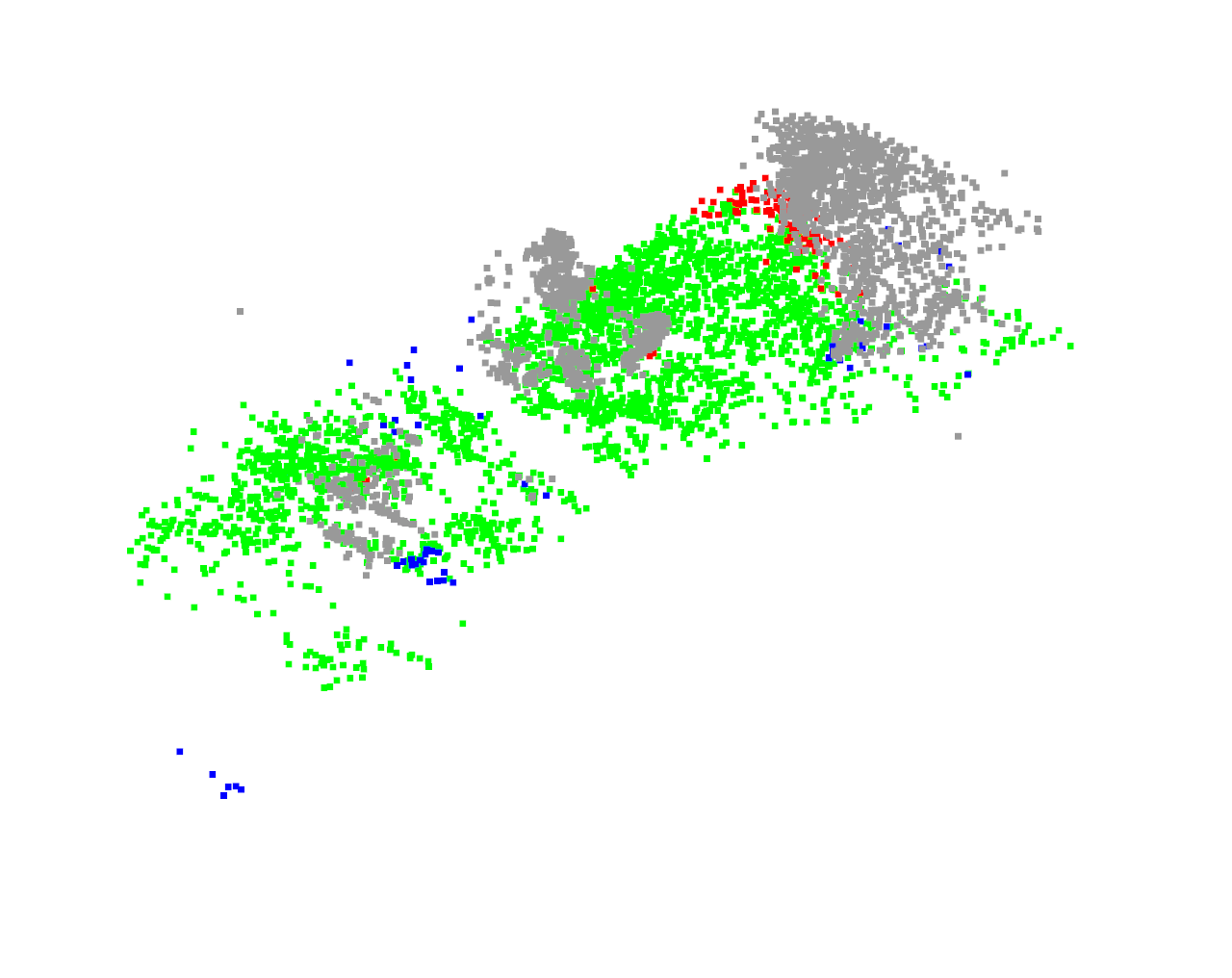} \\

        \bottomrule
    \end{tabular}

    \caption{\textbf{Qualitative ablation study of the proposed ground segmentation method.}
    Each row corresponds to one representative case, and each column shows the segmentation result obtained by progressively adding modules to the baseline. 
    \textbf{TI} and \textbf{PSI} denote temporal integration and prior-based seed initialization, respectively, while \textbf{Proposed} corresponds to the full model with ground segmentation refinement. 
    The visual comparison shows that the added components progressively reduce misclassification and improve the spatial continuity of the extracted ground regions.
    }

    \vspace{-0.3cm}
    \label{fig:ablation_qualitative}
\end{figure*}

To analyze the contribution of each component in the proposed ground segmentation pipeline, an ablation study is conducted by progressively adding temporal integration (TI), prior-based seed initialization (PSI), and ground segmentation refinement (Ref) to the baseline method. 
The no-PSI setting removes only the prior-constrained seed initialization, while the prior elevation consistency check and the terrain-model-based global recall remain active.
The overall quantitative results are reported in Table~\ref{tab:ablation}, while the scenario-wise results are presented in Table~\ref{tab:scenes_ablation}. 
Representative qualitative results are shown in Fig.~\ref{fig:ablation_qualitative}.

As shown in Table~\ref{tab:ablation}, adding temporal integration improves Recall, IoU, and F1 score compared with the baseline, indicating that short-window accumulation provides denser spatial support for sparse mmWave point clouds. 
Introducing prior-based seed initialization further improves segmentation reliability, especially in Precision and IoU, by constraining the initial ground hypotheses using the terrain prior. 
Adding refinement on top of TI improves all metrics, confirming its ability to recover missed ground points. 
However, its precision gain is smaller than that achieved by PSI, indicating that PSI is more effective in suppressing false positives during initialization.
Although the refinement step slightly reduces Precision compared with the combination of  TI and PSI, it substantially improves Recall, IoU, and F1 score, indicating a better balance between segmentation completeness and accuracy.

The scenario-wise results in Table~\ref{tab:scenes_ablation} further reveal the role of each component under different environmental conditions. 
Temporal integration provides consistent improvements in most scenarios by increasing the spatial support of radar observations. 
Prior-based seed initialization is particularly beneficial in the water scenario, where strong reflections and multi-path effects introduce many non-ground outliers; by incorporating terrain priors, the method substantially suppresses false ground hypotheses and improves segmentation reliability. 
Compared with the variant without PSI, the full model shows substantial gains in the water scenario, while slight decreases in some metrics may appear in less outlier-dominated scenes. 
This trade-off comes from the conservative initialization constraint introduced by PSI, which is designed to improve cross-scenario robustness under severe outliers and multi-path interference.
When added to the PSI-based pipeline, the refinement step further improves Recall, IoU, and F1 score across all scenarios, with clear gains in the tea plantation hill scenario where vegetation occlusion and local terrain variation make ground extraction incomplete after the first-pass segmentation.

\begin{table*}[t]
\centering
\caption{\textbf{Scenario-wise ablation study of the proposed ground segmentation method.} 
TI, PSI, and Ref denote temporal integration, prior-based seed initialization, and ground segmentation refinement, respectively. 
All metrics are reported in percentage. \textbf{Bold} values indicate the best performance, and \underline{underlined} values indicate the second-best performance.
}
\label{tab:scenes_ablation}

\renewcommand{\arraystretch}{1.15}
\setlength{\tabcolsep}{3.5pt}

\begin{tabular}{c c c c cccc cccc cccc cccc}
\toprule

\multirow{2}{*}{\textbf{Baseline}} 
& \multirow{2}{*}{\textbf{TI}} 
& \multirow{2}{*}{\textbf{PSI}} 
& \multirow{2}{*}{\textbf{Ref}}
& \multicolumn{4}{c}{\textbf{Flat Field}} 
& \multicolumn{4}{c}{\textbf{Slope}} 
& \multicolumn{4}{c}{\textbf{Water}} 
& \multicolumn{4}{c}{\textbf{Tea Plantation Hill}} \\

\cmidrule(lr){5-8} 
\cmidrule(lr){9-12} 
\cmidrule(lr){13-16} 
\cmidrule(lr){17-20}

& & & 
& \textbf{Prec.} & \textbf{Recall} & \textbf{IoU} & \textbf{F1}
& \textbf{Prec.} & \textbf{Recall} & \textbf{IoU} & \textbf{F1}
& \textbf{Prec.} & \textbf{Recall} & \textbf{IoU} & \textbf{F1}
& \textbf{Prec.} & \textbf{Recall} & \textbf{IoU} & \textbf{F1} \\

\midrule

\checkmark &  &  & 
& \underline{99.26} & 85.84 & 85.29 & 92.06
& 98.44 & 82.50 & 81.44 & 89.77
& 45.69 & 14.85 & 12.62 & 22.41
& 98.57 & 76.19 & 75.36 & 85.95 \\

\checkmark & \checkmark &  &  
& 98.89 & 88.18 & 87.31 & 93.23
& 97.68 & 84.42 & 82.77 & 90.57
& 47.52 & 16.05 & 13.63 & 23.99 
& 98.82 & 77.43 & 76.72 & 86.82 \\

\checkmark & \checkmark & \checkmark &       
& \textbf{99.89} & 88.79 & 88.71 & 94.02
& \textbf{99.32} & 85.45 & 84.95 & 91.86
& \textbf{99.96} & \underline{95.88} & \underline{95.85} & \underline{97.88}
& \textbf{98.93} & 77.80 & 77.15 & 87.10 \\

\checkmark & \checkmark &  & \checkmark     
& 98.97 & \textbf{91.88} & \textbf{91.01} & \textbf{95.29}
& 97.92 & \textbf{93.66} & \textbf{91.83} & \textbf{95.74}
& 54.79 & 21.58 & 18.32 & 30.96
& 98.49 & \textbf{87.77} & \textbf{86.59} & \textbf{92.82} \\

\checkmark & \checkmark & \checkmark & \checkmark   
& \textbf{99.89} & \underline{90.23} & \underline{90.14} & \underline{94.82}
& \underline{99.21} & \underline{91.70} & \underline{91.03} & \underline{95.31}
& \underline{99.88} & \textbf{97.73} & \textbf{97.62} & \textbf{98.79}
& \underline{98.88} & \underline{86.95} & \underline{86.10} & \underline{92.53}\\

\bottomrule
\end{tabular}
\end{table*}

The qualitative results in Fig.~\ref{fig:ablation_qualitative} are consistent with the quantitative trends. 
As more components are introduced, false negatives and false positives are gradually reduced, and the extracted ground regions become more spatially continuous. 
These results verify that the proposed components are complementary and jointly contribute to robust ground segmentation in complex agricultural environments.

\subsection{Terrain Modeling Accuracy}
\subsubsection{Quantitative Comparison}

To evaluate the accuracy of different terrain surface modeling strategies, three representative methods are compared, including $k$-nearest neighbor interpolation (KNN), second-order polynomial fitting (POLY), and cubic B-spline surface modeling (BSP). 
For each method, the terrain height is queried at the ground-truth sampling locations, and the RMSE between the estimated height and the reference elevation is computed. 
The evaluation is conducted across four representative scenarios and three flight altitudes for each scenario.

\begin{table*}[!t]
\centering
\caption{\textbf{RMSE comparison of terrain surface modeling across different scenarios and flight altitudes.} 
All values are reported in meters. Overall denotes the mean RMSE across the three flight altitudes within each scenario. \textbf{Bold} values indicate the best performance.}
\label{tab:rmse_comparison}

\small
\setlength{\tabcolsep}{3pt}
\renewcommand{\arraystretch}{1.1}

\begin{threeparttable}
\resizebox{\textwidth}{!}{%
\begin{tabular}{lcccc cccc cccc cccc}
\toprule

\multirow{2}{*}{\textbf{Method}} &
\multicolumn{4}{c}{\textbf{Flat Field}} &
\multicolumn{4}{c}{\textbf{Slope}} &
\multicolumn{4}{c}{\textbf{Water}} &
\multicolumn{4}{c}{\textbf{Tea Plantation Hill}} \\

\cmidrule(lr){2-5} \cmidrule(lr){6-9} \cmidrule(lr){10-13} \cmidrule(lr){14-17}

& \textbf{3m} & \textbf{5m} & \textbf{8m} & \textbf{Overall}
& \textbf{5m} & \textbf{8m} & \textbf{10m} & \textbf{Overall}
& \textbf{3m} & \textbf{6m} & \textbf{10m} & \textbf{Overall}
& \textbf{18m} & \textbf{20m} & \textbf{22m} & \textbf{Overall} \\

\midrule

KNN\tnote{1}  
& 0.321 & 0.314 & 0.466 & 0.367
& 0.454 & 0.527 & 0.572 & 0.518
& 0.437 & 0.492 & 0.552 & 0.494
& 0.493 & 0.597 & 0.594 & 0.561 \\

POLY\tnote{2}   
& 0.257 & 0.297 & 0.366 & 0.307
& 0.693 & 0.730 & 0.775 & 0.733
& 0.523 & 0.514 & 0.548 & 0.528
& 0.977 & 1.072 & 0.884 & 0.978 \\

\textbf{BSP\tnote{3}}      
& \textbf{0.239} & \textbf{0.261} & \textbf{0.232} & \textbf{0.244}
& \textbf{0.370} & \textbf{0.511} & \textbf{0.466} & \textbf{0.449}
& \textbf{0.332} & \textbf{0.391} & \textbf{0.427} & \textbf{0.383}
& \textbf{0.450} & \textbf{0.568} & \textbf{0.547} & \textbf{0.522} \\

\bottomrule
\end{tabular}%
}

\begin{tablenotes}[flushleft]
\footnotesize
\item[1] KNN: $k$-nearest neighbor interpolation ($k=4$).
\item[2] POLY: Second-order polynomial surface fitting.
\item[3] BSP: Cubic B-spline surface modeling (degree 3).
\item[] Hyperparameters of all methods are empirically tuned for their best performance.
\end{tablenotes}
\end{threeparttable}
\end{table*}

Table~\ref{tab:rmse_comparison} reports the RMSE comparison of different terrain modeling methods. 
The BSP-based terrain model achieves the lowest RMSE across all scenarios and flight altitudes, demonstrating better accuracy and robustness under different terrain conditions.
Compared with KNN, BSP provides smoother and more consistent terrain reconstruction by exploiting the structured control-point representation rather than relying only on local nearest-neighbor interpolation. 
Compared with POLY, BSP avoids the limitations of a single global polynomial model and better adapts to locally varying terrain geometry through its piecewise-polynomial representation.

The advantage of BSP is particularly evident in the slope and tea plantation hill scenarios, where the terrain contains stronger height variation and local surface changes. 
In these scenarios, POLY produces noticeably larger errors, indicating that a low-order global polynomial is insufficient to represent complex terrain variations in these scenarios.
In contrast, BSP maintains lower overall RMSE values of 0.449~m and 0.522~m in the slope and tea plantation hill scenarios, respectively. 
These results indicate that the proposed B-spline terrain model provides a more accurate and stable continuous terrain representation for agricultural UAV terrain perception.

\subsubsection{Runtime Analysis of Terrain Height Query}

\begin{table}[!t]
\centering
\caption{\textbf{Comparison of terrain height query time on the RK3588 platform.} 
All values are reported in milliseconds per query. Overall denotes the average over all evaluated sampling locations. \textbf{Bold} values indicate the best performance.}
\label{tab:query_time}

\small
\setlength{\tabcolsep}{8pt}
\renewcommand{\arraystretch}{1.1}

\begin{threeparttable}
\begin{tabular}{lccccc}
\toprule
\textbf{Method} & \textbf{Flat Field} & \textbf{Slope} & \textbf{Water} & \textbf{Tea Hill} & \textbf{Overall} \\
\midrule
KNN  & 1.061 & 1.428 & 0.451 & 1.624 & 1.141 \\
POLY & 0.358 & 0.269 & 0.214 & 0.245 & 0.272 \\
\textbf{BSP} & \textbf{0.075} & \textbf{0.075} & \textbf{0.060} & \textbf{0.073} & \textbf{0.071} \\
\bottomrule
\end{tabular}
\end{threeparttable}
\end{table}


Beyond reconstruction accuracy, online terrain-following flight also demands efficient terrain height querying. To assess this, the query time of different terrain representations was evaluated on the RK3588 embedded computing platform using the same sampling locations as those used in the accuracy evaluation. Table~\ref{tab:query_time} summarizes the average query latency across various scenarios.

As shown in Table~\ref{tab:query_time}, BSP achieves the lowest query time in all scenarios, with an overall latency of 0.071~ms per query. 
KNN requires repeated neighborhood search during query and therefore has the highest latency, while POLY provides faster evaluation but suffers from lower reconstruction accuracy in complex terrain. 
By contrast, the B-spline surface benefits from local support, where only a limited number of basis functions and control points are involved in each query. 
This enables efficient height evaluation while preserving the smoothness and local adaptability of the terrain surface.

These results show that the BSP-based terrain representation provides a favorable balance between reconstruction accuracy and query efficiency, making it suitable for online terrain-following applications on agricultural UAV platforms.

\subsection{Terrain-Following Consistency Analysis}








\begin{figure*}[htbp]
    \centering
    \setlength{\tabcolsep}{2pt}
    \renewcommand{\arraystretch}{1.0}

    \newlength{\fieldcolw}
    \setlength{\fieldcolw}{0.32\textwidth}

    \newlength{\sceneh}
    \newlength{\obliqueh}
    \newlength{\sideh}
    \setlength{\sceneh}{0.45\fieldcolw}   
    \setlength{\obliqueh}{0.8\fieldcolw}   
    \setlength{\sideh}{0.4\fieldcolw}     

    \newcommand{\fieldimg}[3]{%
        \includegraphics[width=#1,height=#2,keepaspectratio,valign=m]{#3}%
    }

    \newcommand{\fieldlabel}[1]{%
        \adjustbox{valign=m}{%
            \rotatebox[origin=c]{90}{%
                \makebox[2.8cm][c]{\textbf{#1}}%
            }%
        }%
    }

    \begin{tabular}{c c c c}
        \toprule
        & \textbf{Slope} 
        & \textbf{Water} 
        & \textbf{Tea Plantation Hill} \\
        \midrule

        \fieldlabel{Scene}
        & \fieldimg{\fieldcolw}{\sceneh}{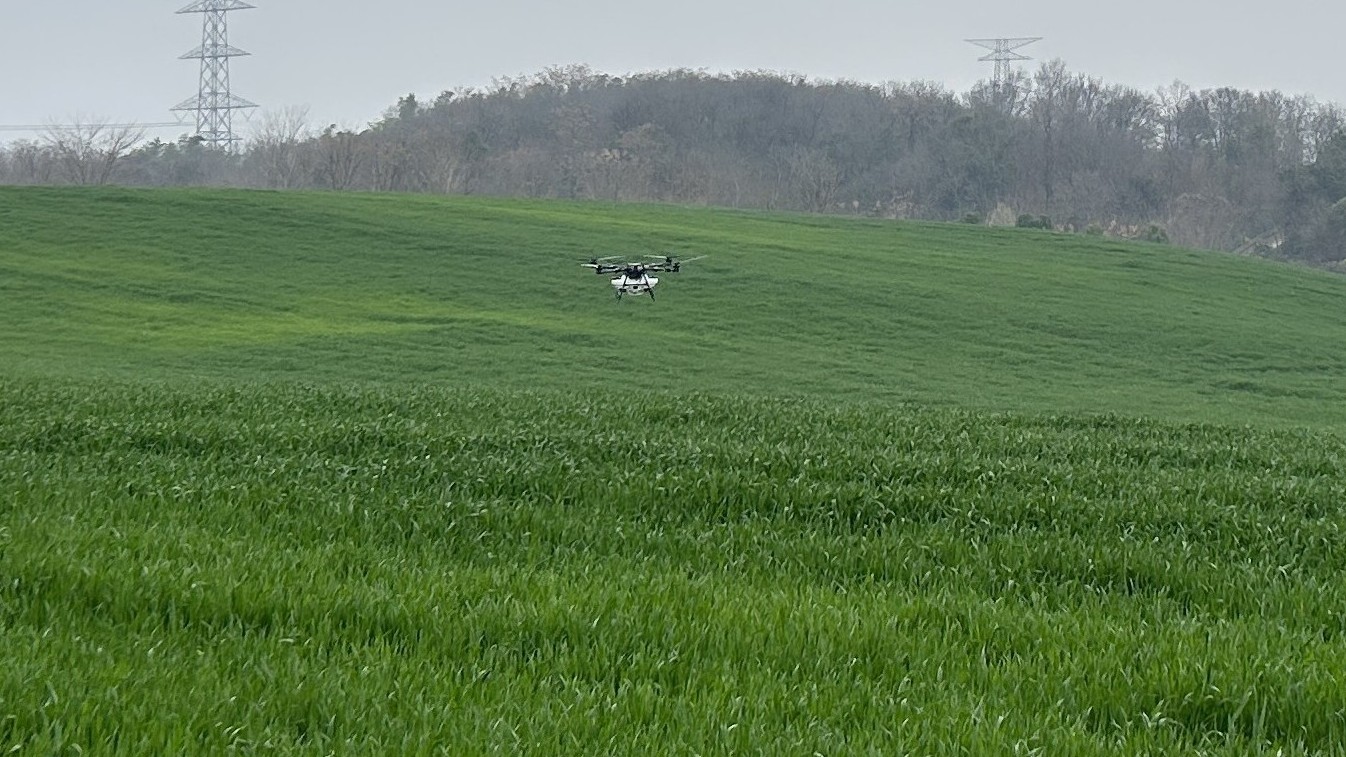}
        & \fieldimg{\fieldcolw}{\sceneh}{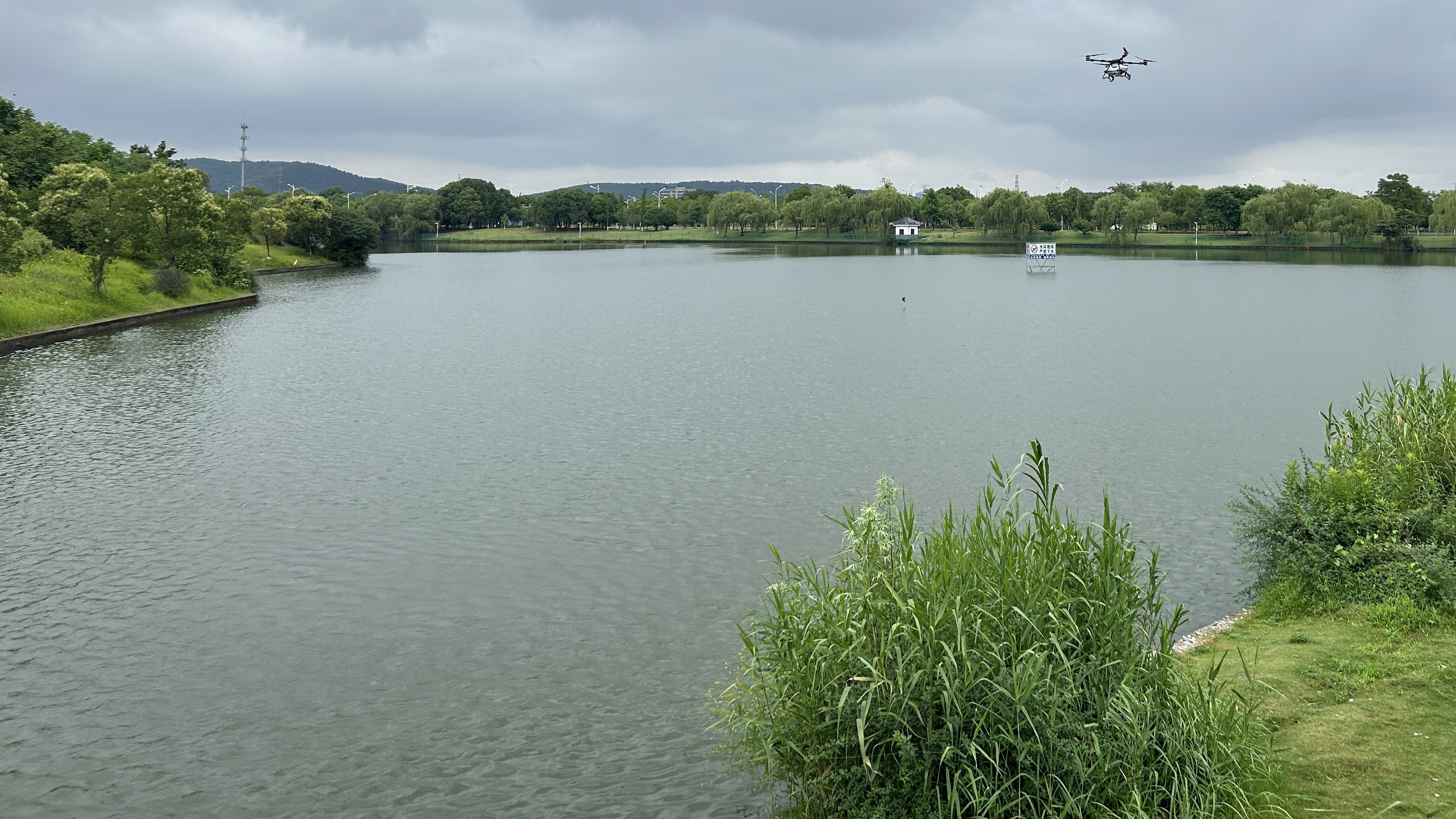}
        & \fieldimg{\fieldcolw}{\sceneh}{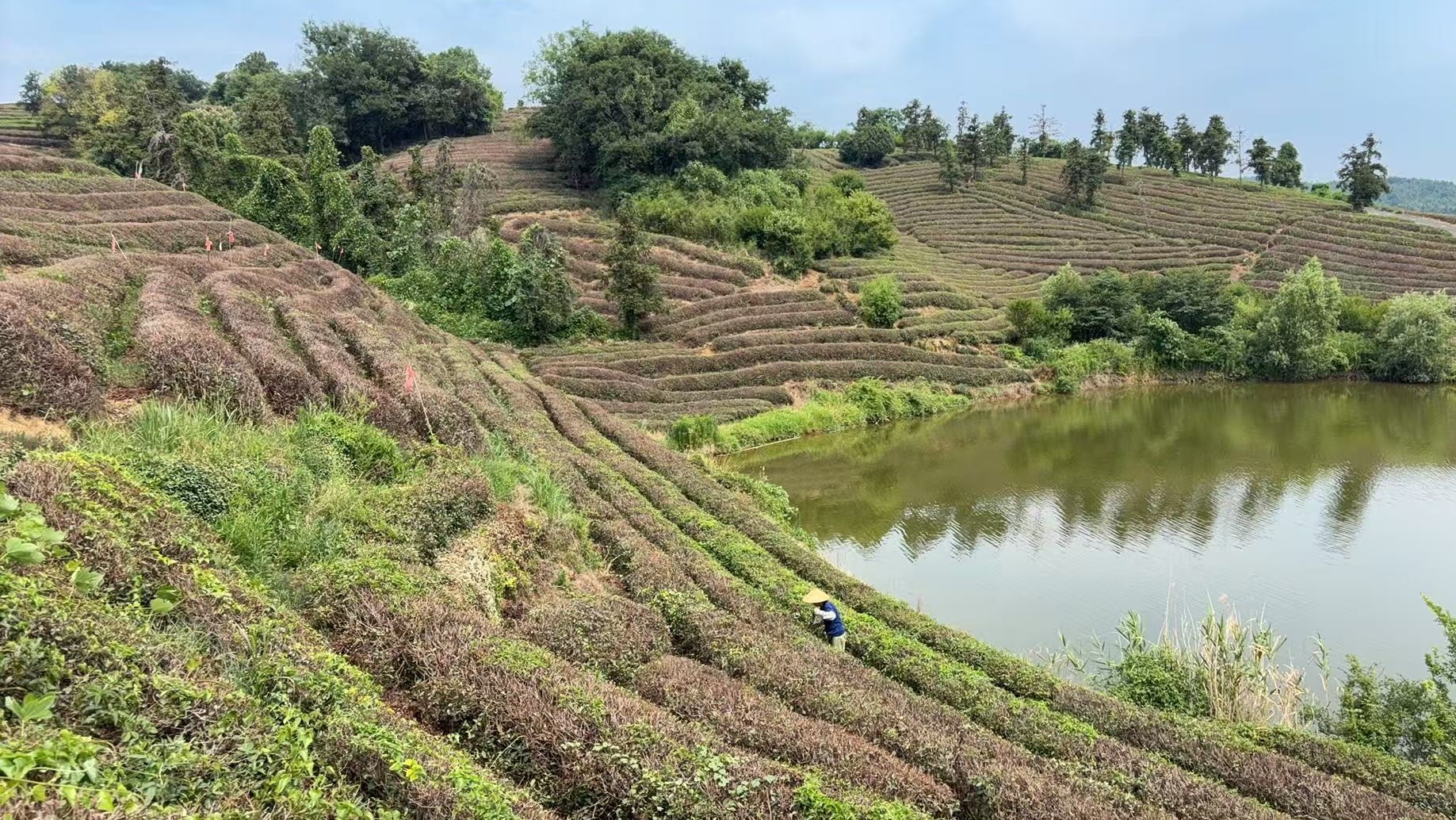} \\

        \fieldlabel{Oblique View}
        & \fieldimg{\fieldcolw}{\obliqueh}{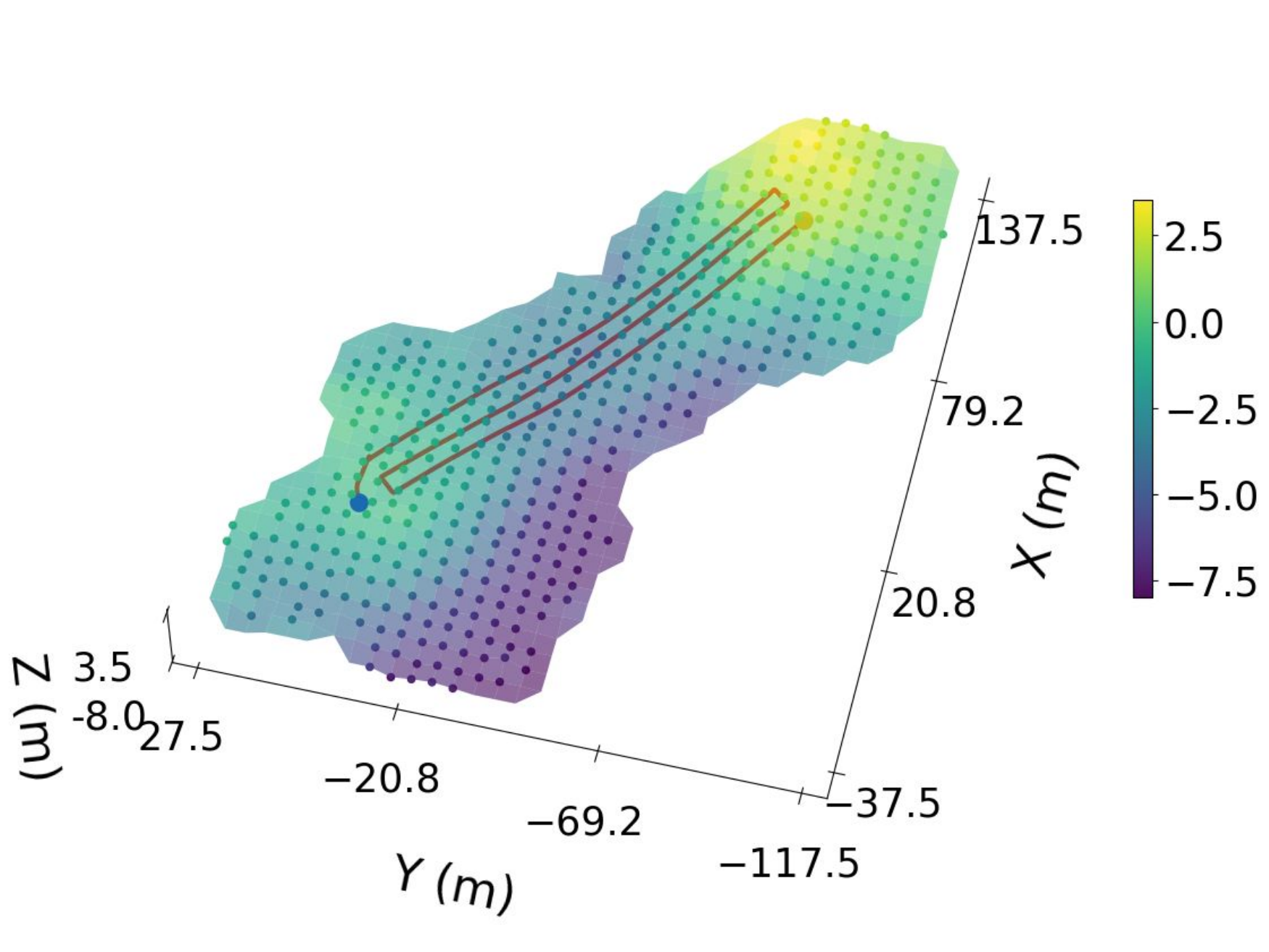}
        & \fieldimg{\fieldcolw}{\obliqueh}{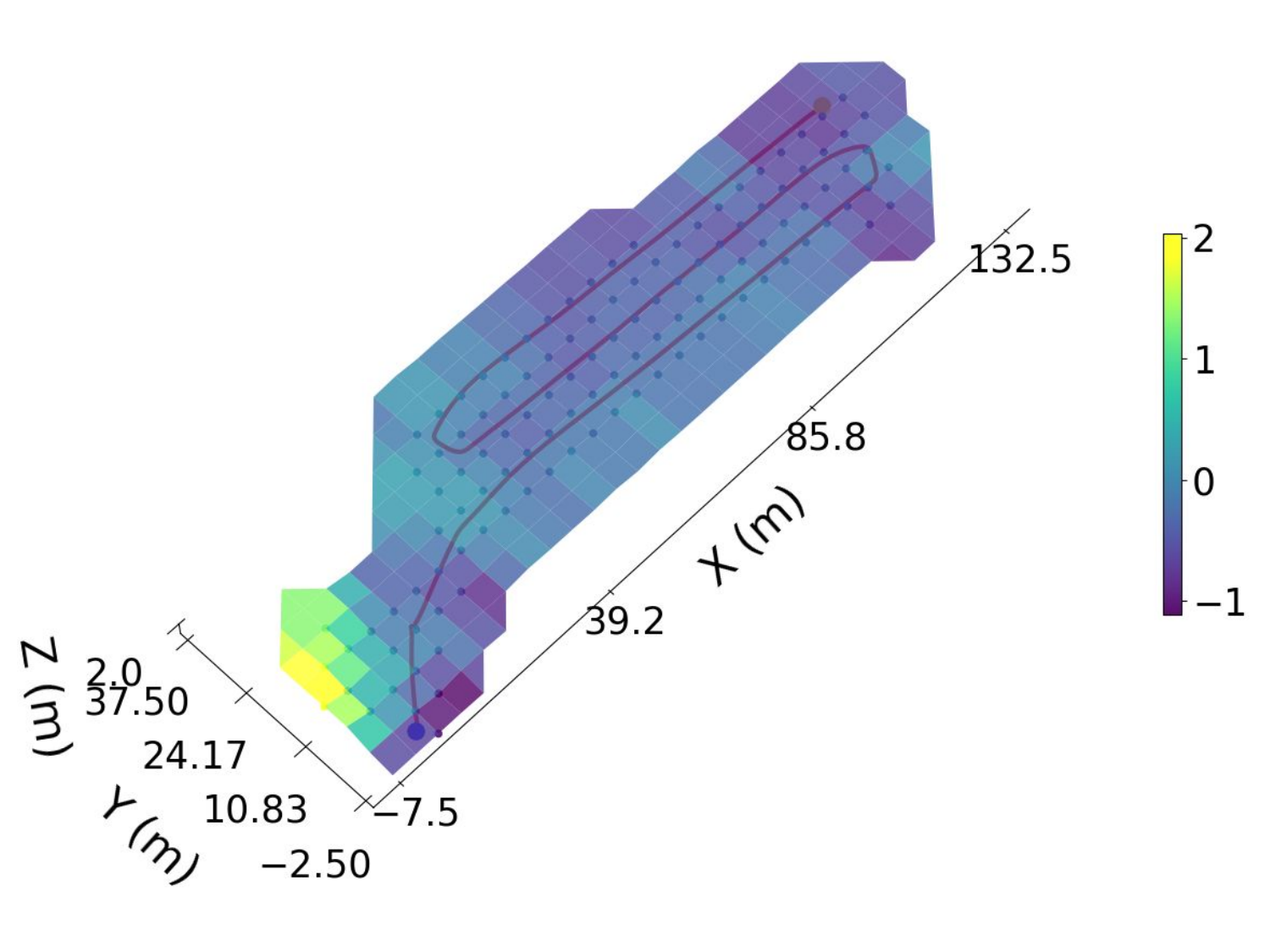}
        & \fieldimg{\fieldcolw}{\obliqueh}{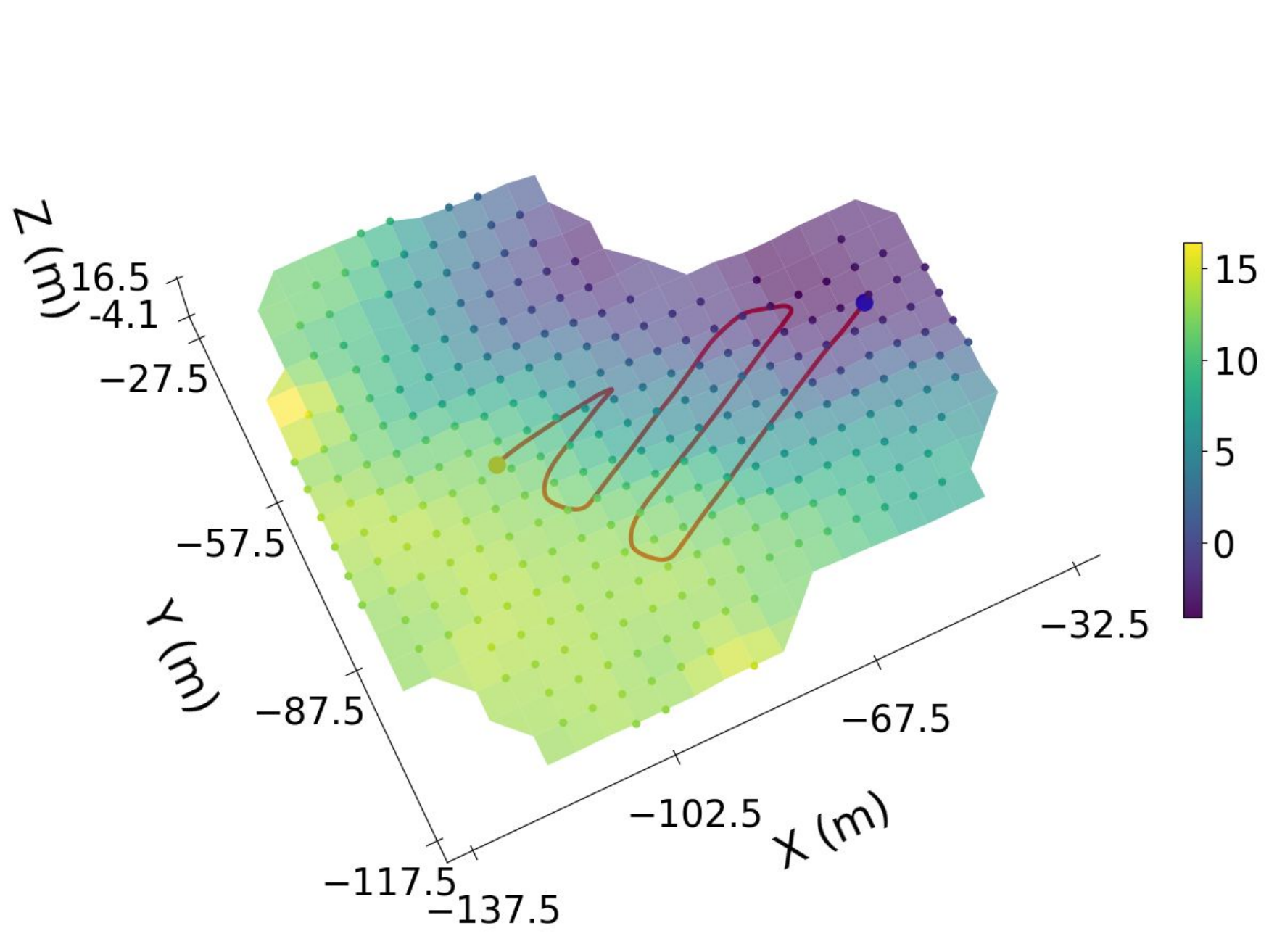} \\

        \fieldlabel{Side View}
        & \fieldimg{\fieldcolw}{\sideh}{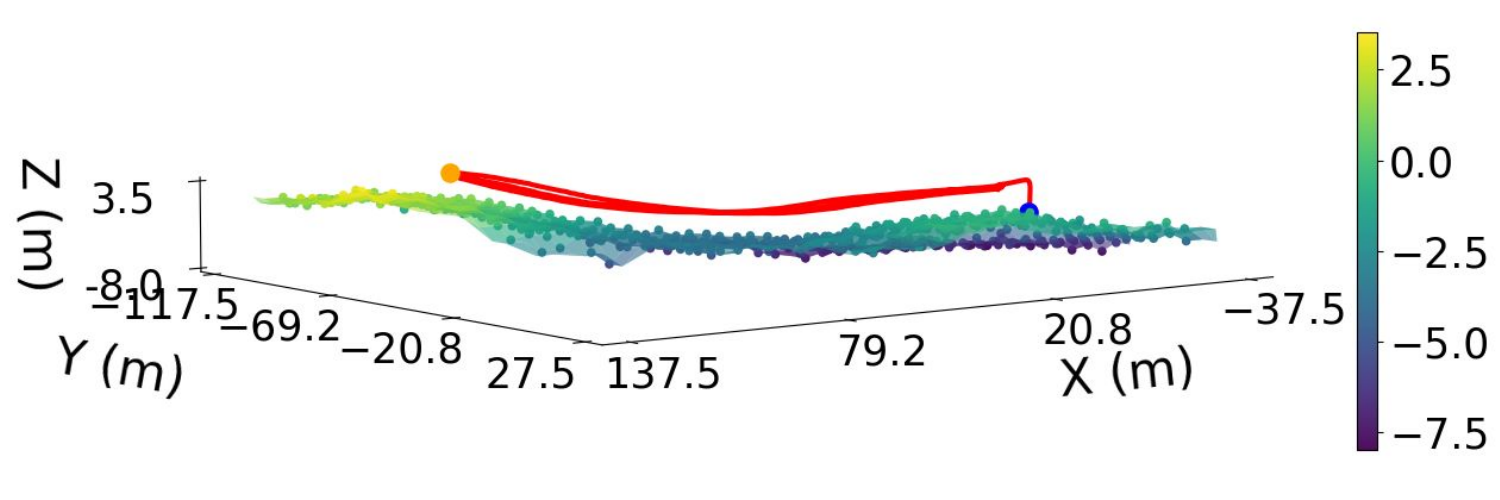}
        & \fieldimg{\fieldcolw}{\sideh}{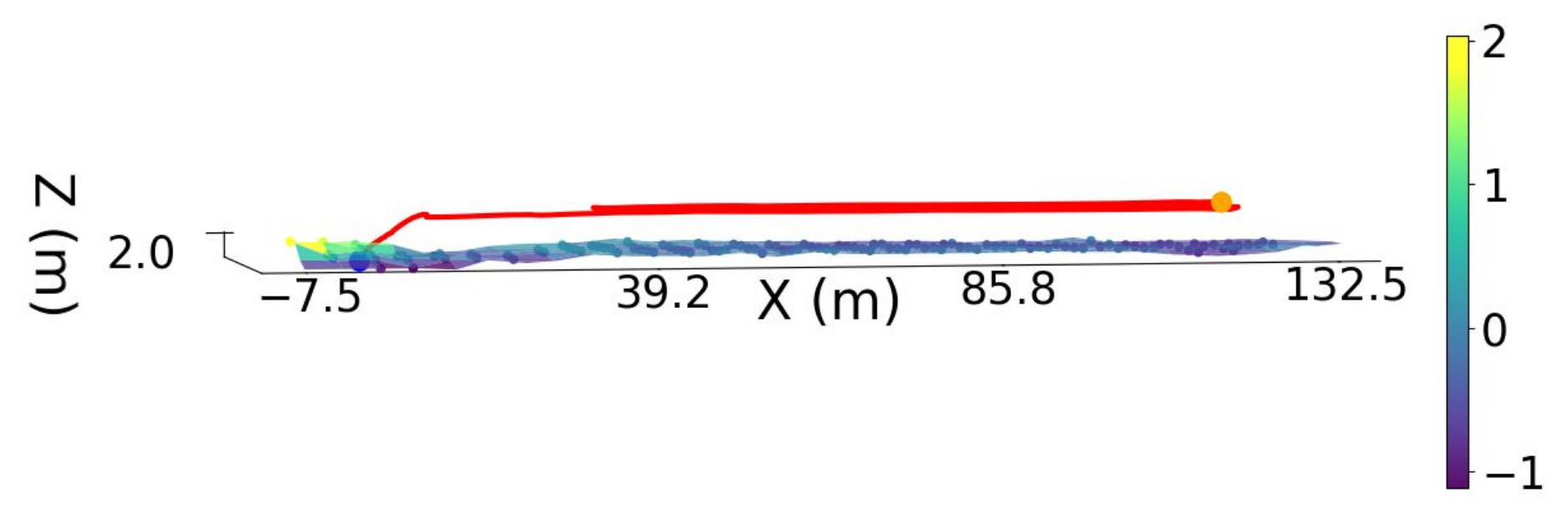}
        & \fieldimg{\fieldcolw}{\sideh}{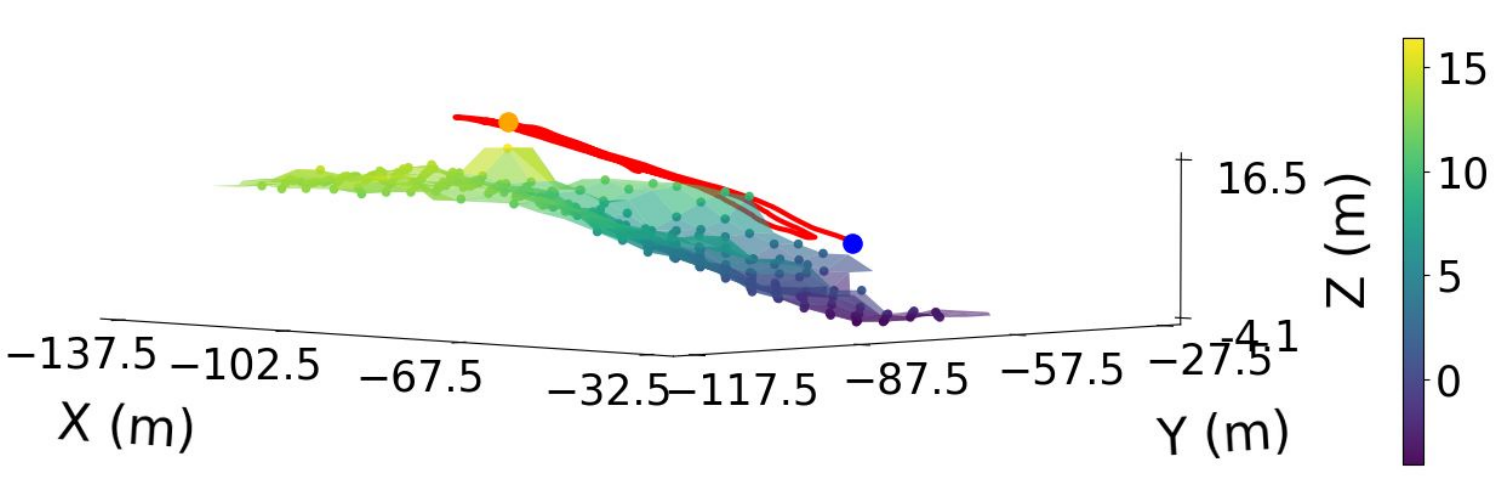} \\

        \bottomrule
    \end{tabular}

    \caption{\textbf{Terrain-following consistency analysis in representative agricultural scenarios.} 
    Each column corresponds to one experimental scenario, including slope, water, and tea plantation hill scenes. 
    From top to bottom, the rows show the reference scene image, the oblique view of the UAV trajectory, and the side view overlaid with the reconstructed terrain surface.
    }

    \vspace{-0.3cm}
    \label{fig:uav_field_results}
\end{figure*}

\begin{table*}[t]
\centering
\caption{\textbf{Closed-loop terrain-relative height consistency based on modality-specific terrain estimates.}
LiDAR and radar results are obtained from separate flights using the same UAV, prescribed path, and flight parameters. 
Residuals are calculated using each flight's own estimated terrain map, without an independent terrain reference. 
Mean, RMSE, and Std. are computed separately for each flight; Average denotes the arithmetic mean of each metric over three flights.
All values are in meters. A dash (--) indicates that terrain-reference generation failed and no valid metric was available.
}
\label{tab:terrain_following_consistency}

\renewcommand{\arraystretch}{1.1}
\setlength{\tabcolsep}{6pt}

\begin{tabular}{l c l cc cc cc}
\toprule
\multirow{2}{*}{\textbf{Scenario}}
& \multirow{2}{*}{\textbf{Desired Height}}
& \multirow{2}{*}{\textbf{Flight}}
& \multicolumn{2}{c}{\textbf{Mean}}
& \multicolumn{2}{c}{\textbf{RMSE}}
& \multicolumn{2}{c}{\textbf{Std.}} \\
\cmidrule(lr){4-5}
\cmidrule(lr){6-7}
\cmidrule(lr){8-9}
& & 
& \textbf{LiDAR} & \textbf{Radar}
& \textbf{LiDAR} & \textbf{Radar}
& \textbf{LiDAR} & \textbf{Radar} \\
\midrule

\multirow{4}{*}{Slope}
& \multirow{4}{*}{3.0}
& Flight 1
& 0.016 & 0.128
& 0.148 & 0.311
& 0.147 & 0.283 \\

& & Flight 2
& 0.024 & 0.147
& 0.164 & 0.331
& 0.163 & 0.297 \\

& & Flight 3
& 0.029 & 0.162
& 0.177 & 0.355
& 0.174 & 0.315 \\

\cmidrule(lr){3-9}
& & \textbf{Average}
& \textbf{0.023} & \textbf{0.146} 
& \textbf{0.163} & \textbf{0.332} 
& \textbf{0.161} & \textbf{0.298} \\

\midrule

\multirow{4}{*}{Water}
& \multirow{4}{*}{5.0}
& Flight 1
& -- & 0.062
& -- & 0.125
& -- & 0.109 \\

& & Flight 2
& -- & 0.068 
& -- & 0.140 
& -- & 0.123 \\

& & Flight 3
& -- & 0.077
& -- & 0.256
& -- & 0.244 \\

\cmidrule(lr){3-9}
& & \textbf{Average}
& -- & \textbf{0.069} 
& -- & \textbf{0.174} 
& -- & \textbf{0.159} \\

\midrule

\multirow{4}{*}{Tea Plantation Hill}
& \multirow{4}{*}{8.0}
& Flight 1 
& -0.026 & -0.033
& 0.256 & 0.398
& 0.254 & 0.397 \\

& & Flight 2
& -0.018 & -0.025
& 0.272 & 0.414
& 0.272 & 0.414 \\

& & Flight 3
& -0.013 & -0.020
& 0.290 & 0.390 
& 0.290 & 0.389 \\

\cmidrule(lr){3-9}
& & \textbf{Average}
& \textbf{-0.019} & \textbf{-0.026}
& \textbf{0.273} & \textbf{0.401} 
& \textbf{0.272} & \textbf{0.400} \\

\bottomrule
\end{tabular}
\end{table*}

\begin{figure*}[htbp]
    \centering
    \setlength{\tabcolsep}{1pt}
    \renewcommand{\arraystretch}{1.02}
    
    \newcommand{\fieldimg}[2][0.48\textwidth]{%
        \includegraphics[width=#1,valign=m]{#2}%
    }

    \newcommand{\fieldlabel}[1]{%
        \adjustbox{valign=m}{%
            \rotatebox[origin=c]{90}{%
                \makebox[3.2cm][c]{\textbf{#1}}%
            }%
        }%
    }
    
    \begin{tabular}{c c c}
        \toprule
        & \textbf{LiDAR} 
        & \textbf{Radar} \\
        \midrule
    
        \fieldlabel{Slope}
        & \fieldimg{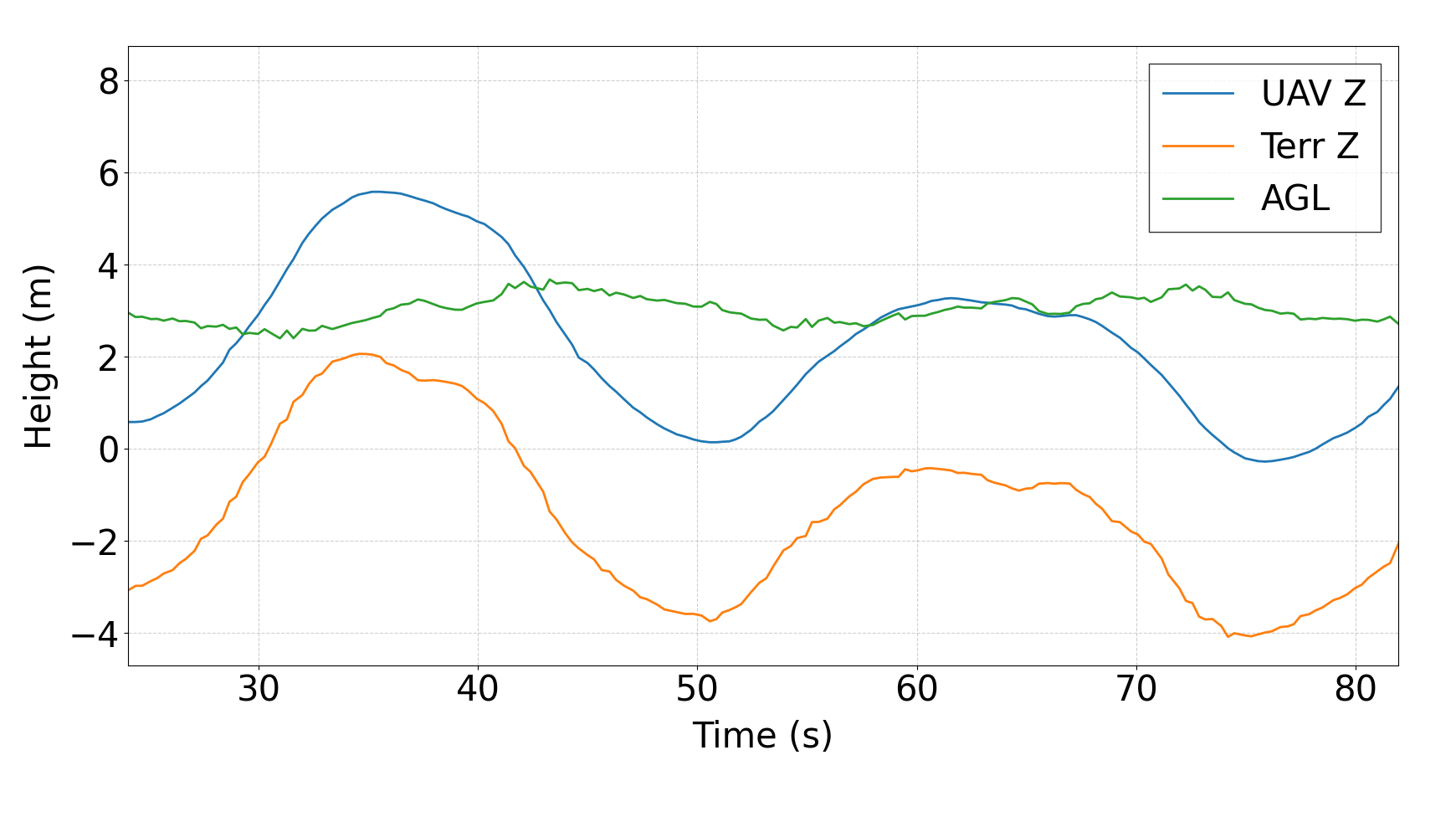}
        & \fieldimg{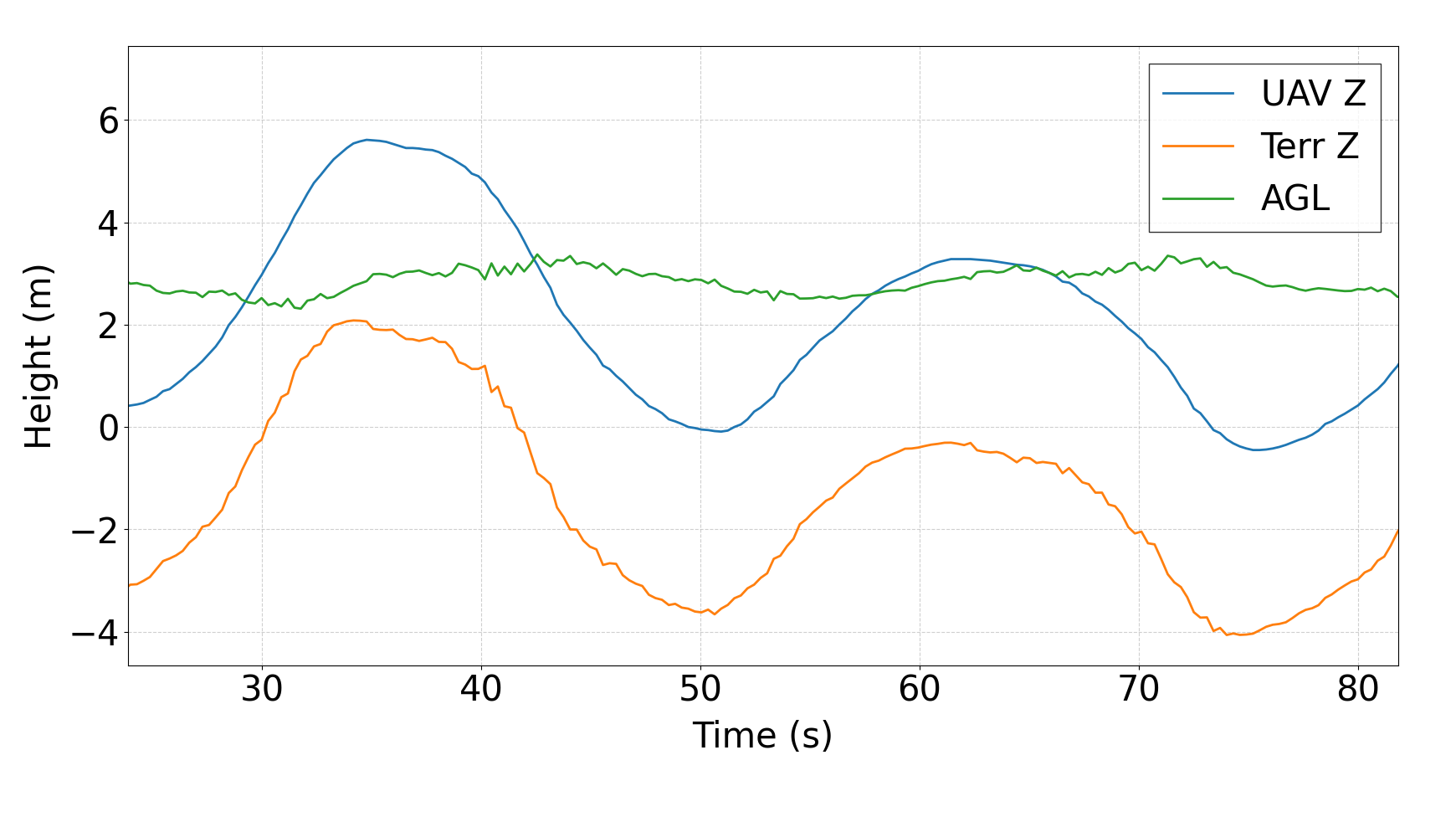} \\

        \fieldlabel{Water}
        & \fieldimg{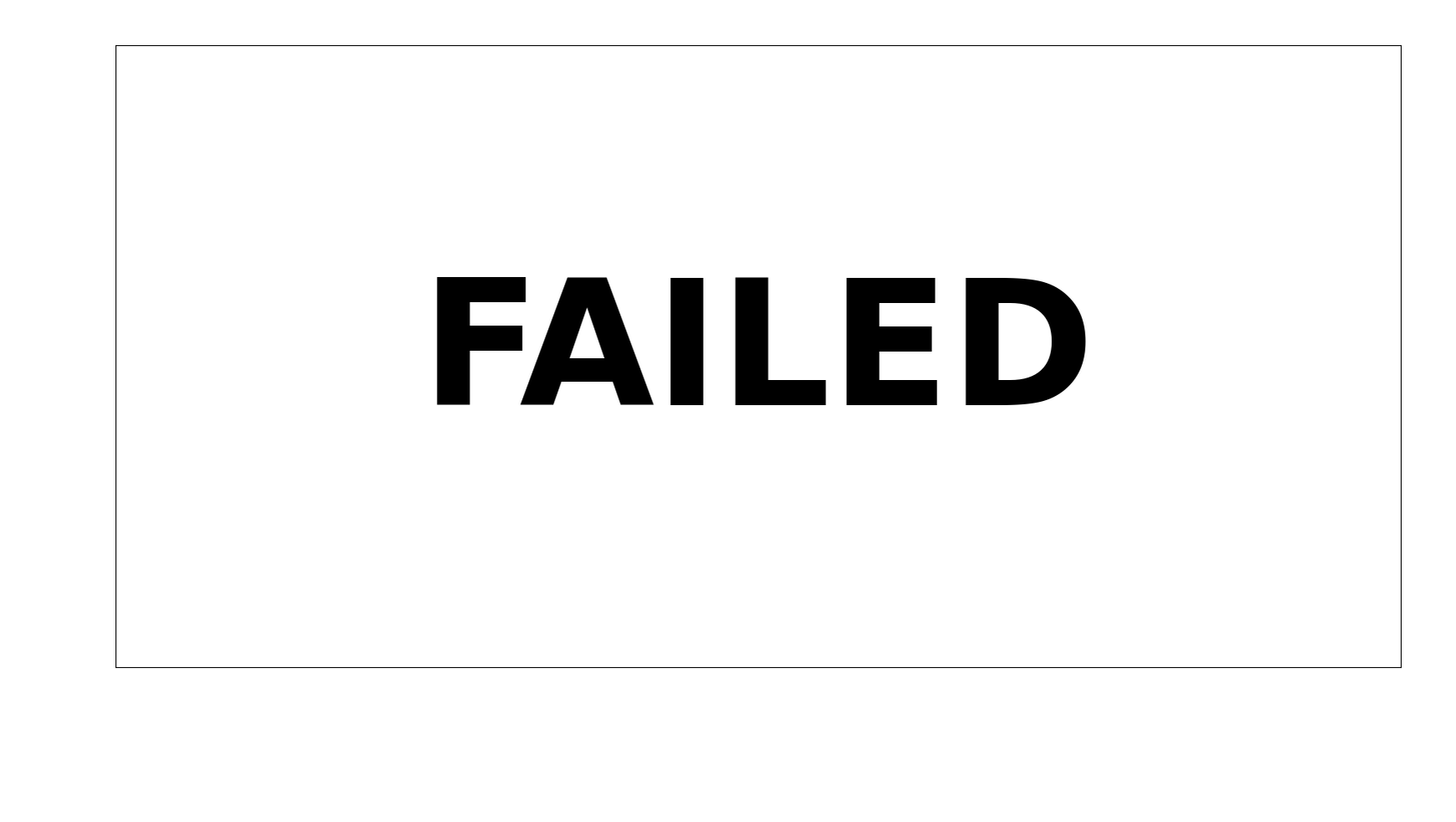}
        & \fieldimg{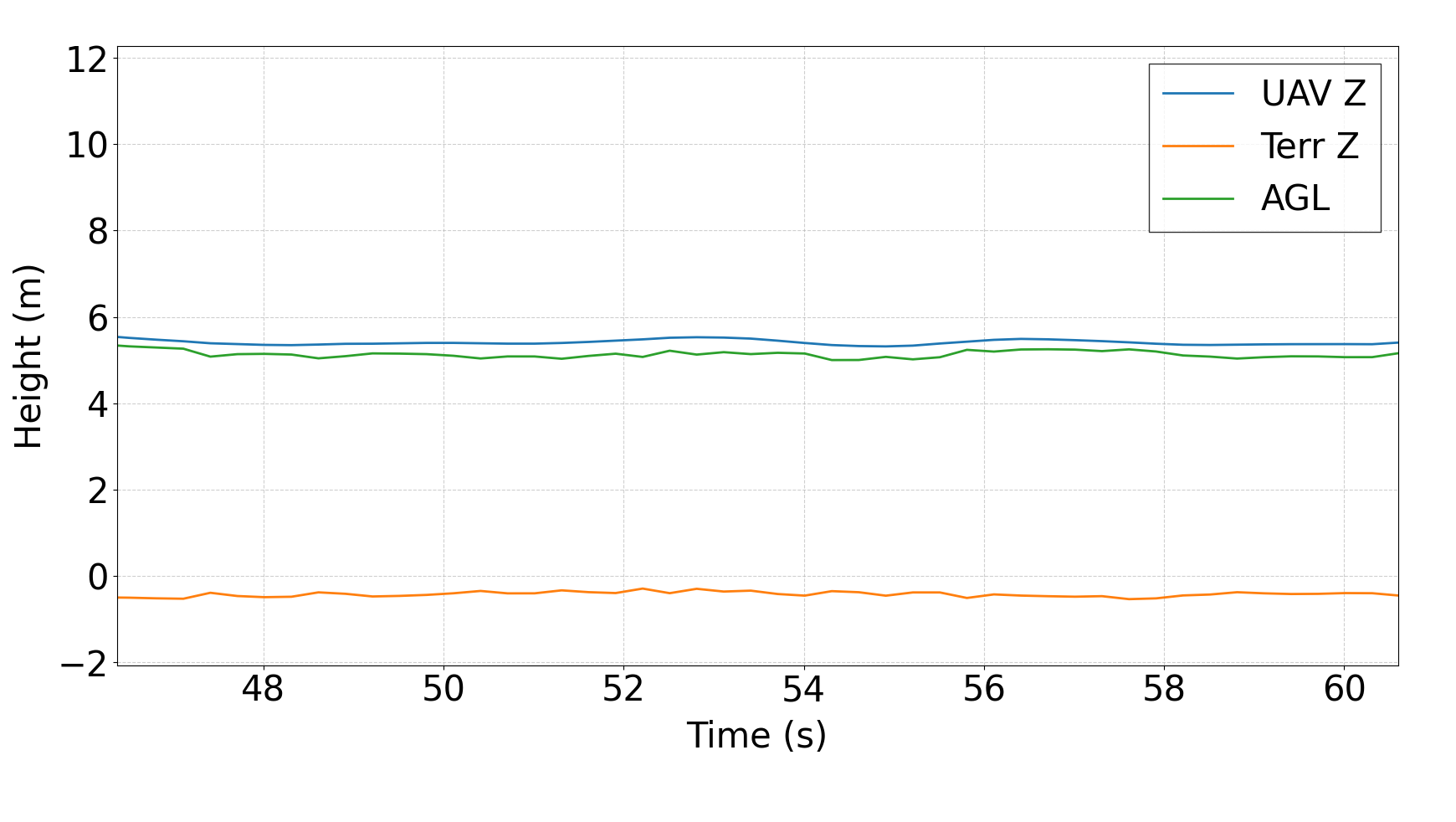} \\
    
        \fieldlabel{Tea Plantation Hill}
        & \fieldimg{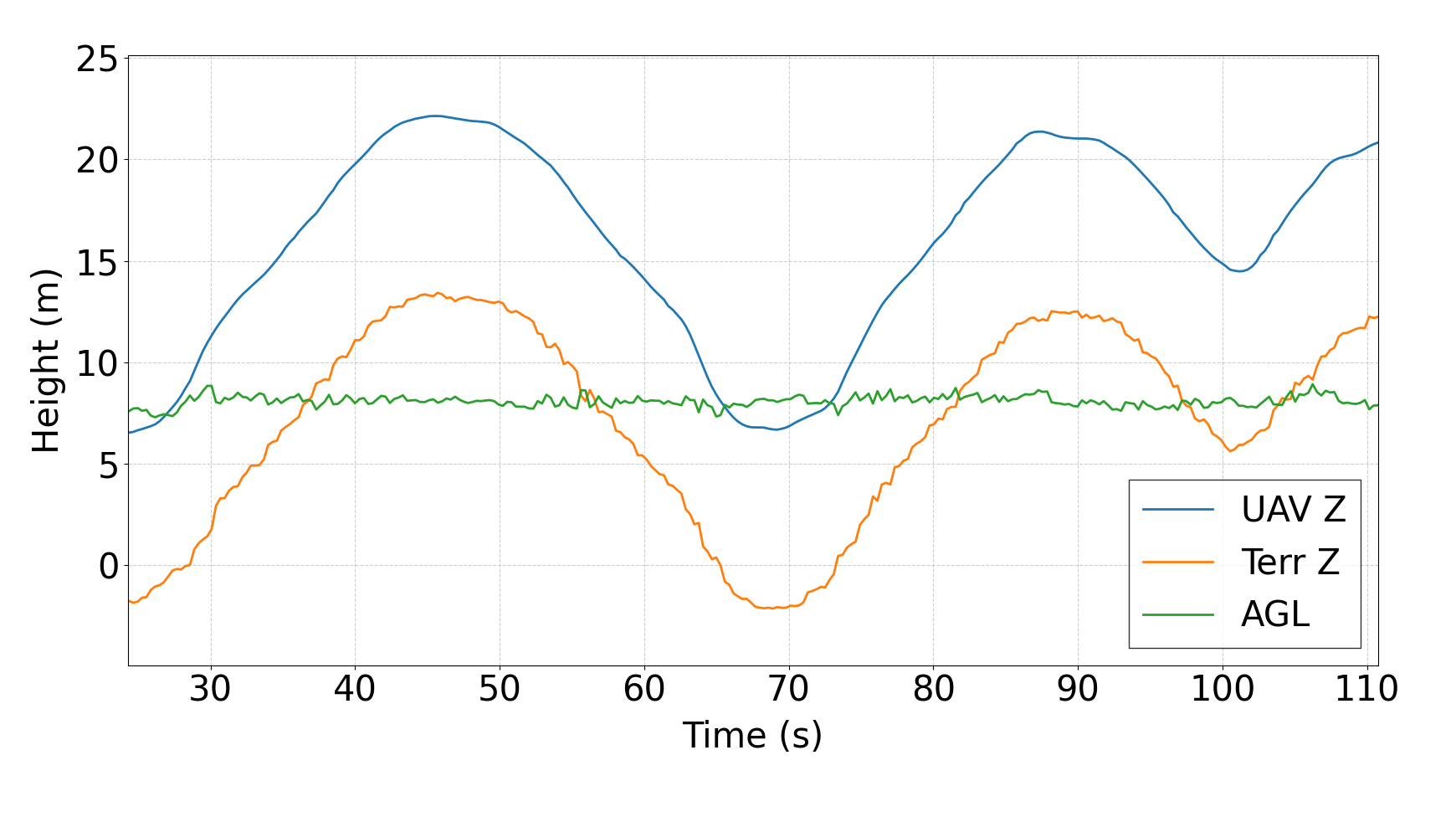}
        & \fieldimg{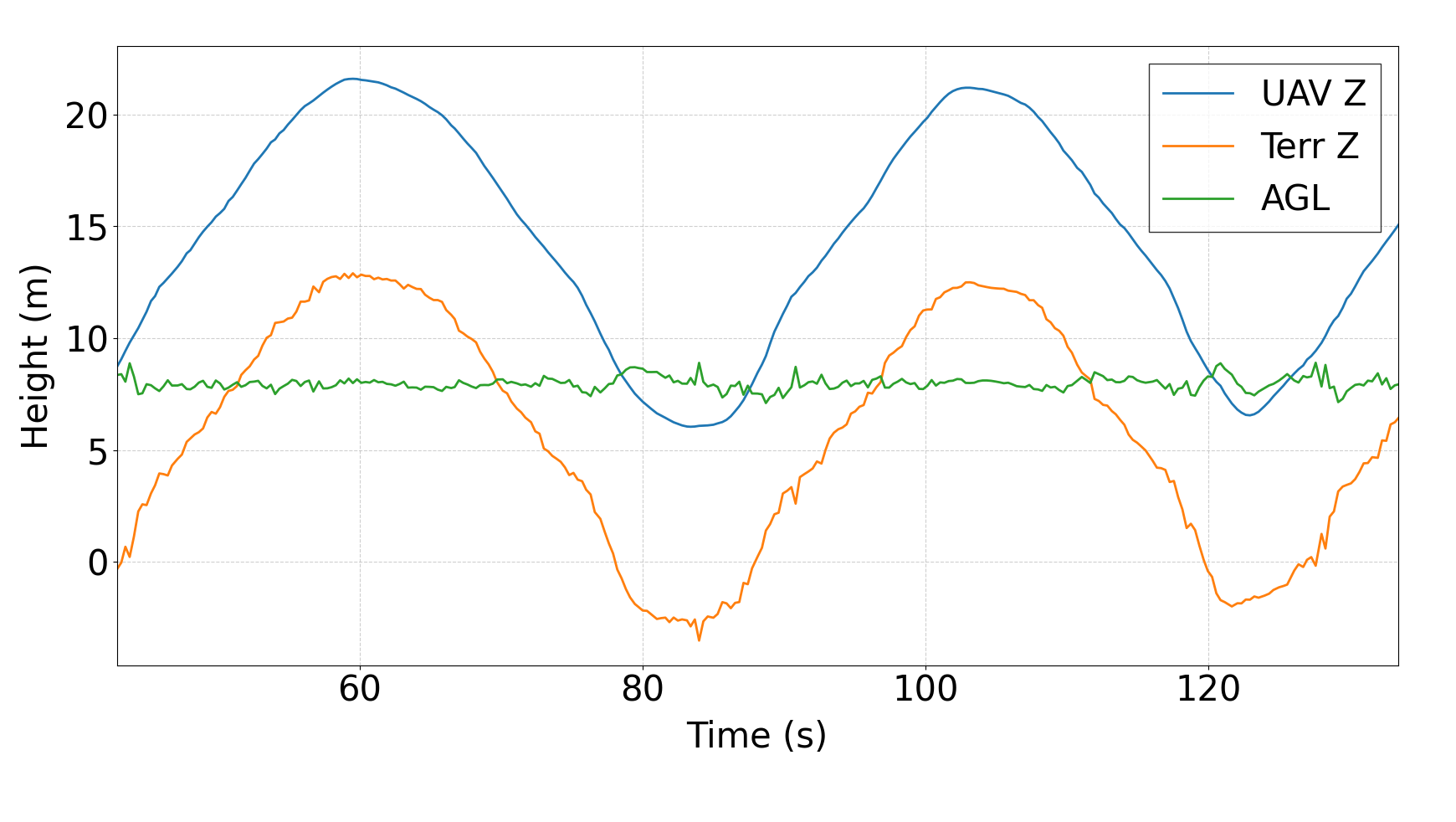} \\
    
        \bottomrule
    \end{tabular}
    
    \caption{
    \textbf{Terrain-following height profiles from separate LiDAR- and radar-based flights along the same prescribed path in each scenario.}
    Rows correspond to slope, water, and tea plantation hill scenarios; columns correspond to LiDAR and radar. Each terrain profile is estimated by the corresponding sensing modality during that flight, rather than obtained from an independent terrain reference.
    }

    \vspace{-0.3cm}
    \label{fig:uav_field_comp}

\end{figure*}

To demonstrate the downstream applicability of the proposed terrain estimation system, terrain-following experiments are conducted in slope, water, and tea plantation hill scenarios. 
These experiments evaluate closed-loop height consistency relative to the estimated terrain surface and the availability of terrain references under challenging sensing conditions. 
The same UAV, prescribed flight path, and flight parameters are used in separate flights with LiDAR and radar as the respective terrain-sensing inputs.
For each scenario, three independent flights are conducted for each sensing modality, resulting in a total of 18 real-flight sorties.

For terrain-following validation, real-flight logs from all three scenarios are analyzed. 
The desired terrain-relative height is set to 3~m, 5~m, and 8~m for the slope, water, and tea plantation hill scenarios, respectively, to accommodate their different terrain and canopy conditions.
Fig.~\ref{fig:uav_field_results} visualizes the spatial alignment between the UAV trajectory and the reconstructed terrain surface.
Fig.~\ref{fig:uav_field_comp} further compares the terrain-following height profiles obtained from LiDAR- and radar-based sensing.

For each sensing modality, the estimated terrain-relative height is computed by subtracting the terrain height queried from that flight's reconstructed map from the UAV altitude. 
The height residual is defined as
\begin{equation}
    e_m(t)=z_{\mathrm{UAV},m}(t)
    -\hat{z}_{\mathrm{terr},m}\bigl(x_m(t),y_m(t)\bigr)
    -h_{\mathrm{desired}},
\end{equation}
where $m$ denotes LiDAR or radar, $z_{\mathrm{UAV},m}(t)$ is provided by the onboard state estimator, and $\hat{z}_{\mathrm{terr},m}$ is queried at the UAV's horizontal position. 
The UAV altitude and terrain height are expressed in the same world coordinate frame. 
No independent terrain reference is used to calculate this residual.

The oblique and side-view visualizations show that the UAV trajectory remains spatially consistent with the reconstructed terrain surface in all three scenarios. 
The terrain control points and the reconstructed 2.5-D surface exhibit smooth and continuous variation, without obvious discontinuities or abrupt height changes. 
These results indicate that the proposed method can provide a spatially coherent terrain reference during real flight.

The height profiles further illustrate how the UAV altitude follows the reference generated from the corresponding estimated terrain surface, with deviations from the desired terrain-relative height.
In the slope scenario, the generated terrain reference is smooth and supports stable terrain-following behavior. 
In the water scenario, LiDAR-based terrain-following reference generation fails, while the proposed radar-based system still provides a valid height profile, showing the advantage of radar perception over weak or unreliable water-surface returns.
In the more challenging tea plantation hill scenario, the proposed method still provides a consistent terrain reference despite larger terrain variation and canopy interference.

Table~8 summarizes the height residuals relative to each system's reconstructed terrain surface. 
Mean, RMSE, and standard deviation are computed independently for each flight, and the Average row reports the arithmetic mean of the corresponding metrics over three flights. 
For radar, the average mean residuals are 0.146~m, 0.069~m, and -0.026~m in the slope, water, and tea plantation hill scenarios, respectively; the corresponding average RMSE values are 0.332~m, 0.174~m, and 0.401~m. 
LiDAR yields lower average RMSE values in the slope and tea plantation hill scenarios, at 0.163~m and 0.273~m, respectively, whereas no valid LiDAR-based terrain reference is obtained in the tested water scenario.

These metrics characterize closed-loop height consistency relative to the estimated terrain surface, rather than independently measured above-ground height accuracy. 
A systematic terrain-model bias may therefore remain undetected by this evaluation. 
Moreover, the LiDAR and radar results are obtained from separate flights, so differences in realized trajectories and disturbances may also influence the comparison. 
The LiDAR residual is consequently not interpreted as pure controller tracking error, and the radar-LiDAR RMSE difference does not isolate radar perception error. 
Terrain estimation accuracy against independently surveyed reference points is evaluated separately in Section~IV-D.

Overall, these experiments demonstrate the use of the proposed terrain estimation system as an online reference for downstream terrain-following. 
The comparison illustrates a practical trade-off: LiDAR provides lower height-consistency errors in the tested slope and tea plantation hill scenarios, while radar retains a usable terrain reference over the tested water surface where LiDAR-based reference generation fails. 
These results support the applicability of the proposed system under the evaluated agricultural conditions.

\subsection{Practical Limitations}

The proposed system relies on terrain priors that are progressively established after initialization. 
When the online terrain model is not yet available, PSI uses the default terrain height $z_{u,v}^{\mathrm{hist}}=0$, while the prior elevation consistency check is bypassed and global recall is not applied at locations without a valid terrain prior. 
The geometric and statistical validation stages remain active.
This zero initialization is consistent with the adopted local world frame, whose origin is anchored at the UAV takeoff position with the $z$-axis aligned vertically. 
Under the intended agricultural operating conditions, the UAV is initialized from an approximately level and open area, such that the local ground elevation remains close to this initial reference.

The transition to prior-based operation is governed by the accumulation of valid terrain observations rather than by the convergence of an iterative estimator. 
The preprocessing stage integrates the current radar frame with the preceding $K-1$ frames, with $K=3$ in our implementation. 
At the effective processing rate of 10~Hz, accumulating the required $K$ radar frames takes approximately 0.3~s from initialization.
Once valid ground patches are obtained, the incrementally constructed terrain model can provide queried terrain heights to replace the default PSI value. 
The local prior elevation consistency check additionally requires $N_h=3$ valid historical ground frames, corresponding to approximately another 0.3~s under continuous valid observations. 
Thus, the local prior consistency check becomes available on the order of 0.6~s after initialization under nominal conditions. 
This duration represents the nominal data-accumulation time rather than a guaranteed recovery time, since sparse or unreliable ground returns may delay the establishment of the terrain priors. 
Moreover, the incremental update rule allows early control-point estimates to be revised when subsequent observations differ sufficiently from the stored values, reducing the persistence of initialization errors in the terrain model.

The current initialization therefore assumes an approximately level and open takeoff area. 
Strongly sloped, highly uneven, or heavily obstructed takeoff locations may cause the zero-centered PSI prior to deviate from the local ground elevation, reducing the availability or reliability of the initial seed candidates. 
Before sufficient historical observations have been accumulated, the prior elevation consistency check is also unavailable, increasing the reliance on the remaining geometric and statistical criteria. 
Sparse or unreliable ground returns may further delay the construction of the online terrain model and the local historical prior. 
These effects are progressively reduced as valid observations accumulate and the terrain priors become available; nevertheless, initialization over arbitrary terrain remains a practical limitation of the current system.

\section{CONCLUSION AND FUTURE WORK}
\label{sec:conclusion}

This paper presented a rotating mmWave radar-based terrain perception system for agricultural UAV terrain-following applications. 
A complete pipeline was developed, including point cloud preprocessing, region-wise ground segmentation, ground refinement, and continuous terrain surface modeling. 
Based on the extracted ground observations and B-spline-based 2.5-D reconstruction, the proposed method provides a smooth and incrementally updatable terrain reference for online terrain-following flight.

Field experiments in multiple agricultural scenarios demonstrated that the proposed method consistently outperformed representative baselines in both ground segmentation and terrain modeling accuracy. 
In addition, real-flight terrain-following experiments demonstrated consistent closed-loop height behavior across the slope, water, and tea plantation hill scenarios. 
In particular, the radar-based system retained a usable terrain reference over the tested water surface, where LiDAR-based terrain-reference generation failed.

Future work will focus on improving the adaptability of the current method under more diverse environments and sensor configurations, extending initialization to more general takeoff conditions, exploring richer terrain representations beyond 2.5-D modeling, and further integrating the proposed terrain perception module with predictive terrain-following planning and control for fully autonomous agricultural UAV operation.

\section*{ACKNOWLEDGMENT}

The authors thank TopXGun Robotics for supporting field experiments and system deployment.

\bibliographystyle{IEEEtran}
\bibliography{ref}

@article{cheng2023precision,
  title={Precision agriculture management based on a surrogate model assisted multiobjective algorithmic framework},
  author={Cheng, Du and Yao, Yifei and Liu, Renyun and Li, Xiaoning and Guan, Boxu and Yu, Fanhua},
  journal={Scientific Reports},
  volume={13},
  number={1},
  pages={1142},
  year={2023},
  publisher={Nature Publishing Group UK London}
}

@article{radoglou2020compilation,
  title={A compilation of UAV applications for precision agriculture},
  author={Radoglou-Grammatikis, Panagiotis and Sarigiannidis, Panagiotis and Lagkas, Thomas and Moscholios, Ioannis},
  journal={Computer Networks},
  volume={172},
  pages={107148},
  year={2020},
  publisher={Elsevier}
}

@inproceedings{he2025multi,
  title={A Multi-Sensor Fusion Approach for Rapid Orthoimage Generation in Large-Scale UAV Mapping},
  author={He, Jialei and Zhan, Zhihao and Tu, Zhituo and Zhu, Xiang and Yuan, Jie},
  booktitle={2025 IEEE/RSJ International Conference on Intelligent Robots and Systems (IROS)},
  pages={6808--6815},
  year={2025},
  organization={IEEE}
}

@article{wang2024uav,
  title={UAV multispectral remote sensing for agriculture: A comparative study of radiometric correction methods under varying illumination conditions},
  author={Wang, Yuxiang and Kootstra, Gert and Yang, Zengling and Khan, Haris Ahmad},
  journal={Biosystems Engineering},
  volume={248},
  pages={240--254},
  year={2024},
  publisher={Elsevier}
}

@article{farhan2024comprehensive,
  title={A comprehensive review of LiDAR applications in crop management for precision agriculture},
  author={Farhan, Sheikh Muhammad and Yin, Jianjun and Chen, Zhijian and Memon, Muhammad Sohail},
  journal={Sensors},
  volume={24},
  number={16},
  pages={5409},
  year={2024},
  publisher={MDPI}
}

@article{corradi2022radar,
  title={Radar perception for autonomous unmanned aerial vehicles: A survey},
  author={Corradi, Federico and Fioranelli, Francesco},
  journal={System Engineering for constrained embedded systems},
  pages={14--20},
  year={2022}
}

@inproceedings{doer2022gnss,
  title={Gnss aided radar inertial odometry for uas flights in challenging conditions},
  author={Doer, Christopher and Atman, Jamal and Trnmmer, Gert F},
  booktitle={2022 IEEE Aerospace Conference (AERO)},
  pages={1--10},
  year={2022},
  organization={IEEE}
}

@article{zhan2025agrilira4d,
  title={AgriLiRa4D: A Multi-Sensor UAV Dataset for Robust SLAM in Challenging Agricultural Fields},
  author={Zhan, Zhihao and Ming, Yuhang and Li, Shaobin and Yuan, Jie},
  journal={arXiv preprint arXiv:2512.01753},
  year={2025}
}

@article{tarolli2020agriculture,
  title={Agriculture in hilly and mountainous landscapes: threats, monitoring and sustainable management},
  author={Tarolli, Paolo and Straffelini, Eugenio},
  journal={Geography and sustainability},
  volume={1},
  number={1},
  pages={70--76},
  year={2020},
  publisher={Elsevier}
}

@inproceedings{zermas2017fast,
  title={Fast segmentation of 3D point clouds: A paradigm on LiDAR data for autonomous vehicle applications},
  author={Zermas, Dimitris and Izzat, Izzat and Papanikolopoulos, Nikolaos},
  booktitle={2017 IEEE International Conference on Robotics and Automation (ICRA)},
  pages={5067--5073},
  year={2017},
  organization={IEEE}
}

@article{lim2021patchwork,
  title={Patchwork: Concentric zone-based region-wise ground segmentation with ground likelihood estimation using a 3D LiDAR sensor},
  author={Lim, Hyungtae and Oh, Minho and Myung, Hyun},
  journal={IEEE Robotics and Automation Letters},
  volume={6},
  number={4},
  pages={6458--6465},
  year={2021},
  publisher={IEEE}
}

@inproceedings{lee2022patchwork++,
  title={Patchwork++: Fast and robust ground segmentation solving partial under-segmentation using 3D point cloud},
  author={Lee, Seungjae and Lim, Hyungtae and Myung, Hyun},
  booktitle={2022 IEEE/RSJ International Conference on Intelligent Robots and Systems (IROS)},
  pages={13276--13283},
  year={2022},
  organization={IEEE}
}

@INPROCEEDINGS{khan2012visual,
  author={Khan, Yasir Niaz and Masselli, Andreas and Zell, Andreas},
  booktitle={2012 IEEE International Conference on Robotics and Automation}, 
  title={Visual terrain classification by flying robots}, 
  year={2012},
  volume={},
  number={},
  pages={498-503},
  doi={10.1109/ICRA.2012.6224988}}

@INPROCEEDINGS{forster2015continuous,
  author={Forster, Christian and Faessler, Matthias and Fontana, Flavio and Werlberger, Manuel and Scaramuzza, Davide},
  booktitle={2015 IEEE International Conference on Robotics and Automation (ICRA)}, 
  title={Continuous on-board monocular-vision-based elevation mapping applied to autonomous landing of micro aerial vehicles}, 
  year={2015},
  volume={},
  number={},
  pages={111-118},
  doi={10.1109/ICRA.2015.7138988}}

@Article{campos2016height,
AUTHOR = {Campos, Igor S. G. and Nascimento, Erickson R. and Freitas, Gustavo M. and Chaimowicz, Luiz},
TITLE = {A Height Estimation Approach for Terrain Following Flights from Monocular Vision},
JOURNAL = {Sensors},
VOLUME = {16},
YEAR = {2016},
NUMBER = {12},
ARTICLE-NUMBER = {2071},
URL = {https://www.mdpi.com/1424-8220/16/12/2071},
PubMedID = {27929424},
ISSN = {1424-8220},
DOI = {10.3390/s16122071}
}

@article{garratt2008vision,
author = {Garratt, Matthew A. and Chahl, Javaan S.},
title = {Vision-based terrain following for an unmanned rotorcraft},
journal = {Journal of Field Robotics},
volume = {25},
number = {4-5},
pages = {284-301},
doi = {https://doi.org/10.1002/rob.20239},
url = {https://onlinelibrary.wiley.com/doi/abs/10.1002/rob.20239},
eprint = {https://onlinelibrary.wiley.com/doi/pdf/10.1002/rob.20239},
year = {2008}
}

@INPROCEEDINGS{campos2015terrain,
  author={Campos, Igor S.G. and Nascimento, Erickson R. and Chaimowicz, Luiz},
  booktitle={2015 12th Latin American Robotics Symposium and 2015 3rd Brazilian Symposium on Robotics (LARS-SBR)}, 
  title={Terrain Classification from UAV Flights Using Monocular Vision}, 
  year={2015},
  volume={},
  number={},
  pages={271-276},
  doi={10.1109/LARS-SBR.2015.49}}

@ARTICLE{xue2025traversability,
  author={Xue, Hanzhang and Xiao, Liang and Hu, Xiaochang and Xie, Rui and Fu, Hao and Nie, Yiming and Dai, Bin},
  journal={IEEE Transactions on Intelligent Transportation Systems}, 
  title={Contrastive Label Disambiguation for Self-Supervised Terrain Traversability Learning in Off-Road Environments}, 
  year={2025},
  volume={26},
  number={12},
  pages={22830-22842},
  doi={10.1109/TITS.2025.3611408}}

@ARTICLE{zurn2021terrainclassification,
  author={Zürn, Jannik and Burgard, Wolfram and Valada, Abhinav},
  journal={IEEE Transactions on Robotics}, 
  title={Self-Supervised Visual Terrain Classification From Unsupervised Acoustic Feature Learning}, 
  year={2021},
  volume={37},
  number={2},
  pages={466-481},
  doi={10.1109/TRO.2020.3031214}}

@ARTICLE{chung2024elevation,
  author={Chung, Chanyoung and Georgakis, Georgios and Spieler, Patrick and Padgett, Curtis and Agha, Ali and Khattak, Shehryar},
  journal={IEEE Robotics and Automation Letters}, 
  title={Pixel to Elevation: Learning to Predict Elevation Maps at Long Range Using Images for Autonomous Offroad Navigation}, 
  year={2024},
  volume={9},
  number={7},
  pages={6170-6177},
  doi={10.1109/LRA.2024.3396089}}

@ARTICLE{chen2024terrainparameter,
  author={Chen, Jiaqi and Frey, Jonas and Zhou, Ruyi and Miki, Takahiro and Martius, Georg and Hutter, Marco},
  journal={IEEE Robotics and Automation Letters}, 
  title={Identifying Terrain Physical Parameters From Vision - Towards Physical-Parameter-Aware Locomotion and Navigation}, 
  year={2024},
  volume={9},
  number={11},
  pages={9279-9286},
  doi={10.1109/LRA.2024.3455788}}

@ARTICLE{cai2024riskaware,
  author={Cai, Xiaoyi and Ancha, Siddharth and Sharma, Lakshay and Osteen, Philip R. and Bucher, Bernadette and Phillips, Stephen and Wang, Jiuguang and Everett, Michael and Roy, Nicholas and How, Jonathan P.},
  journal={IEEE Transactions on Robotics}, 
  title={EVORA: Deep Evidential Traversability Learning for Risk-Aware Off-Road Autonomy}, 
  year={2024},
  volume={40},
  number={},
  pages={3756-3777},
  doi={10.1109/TRO.2024.3431828}}

@Article{du2022uavlidar,
AUTHOR = {Du, Mengmeng and Li, Hanyuan and Roshanianfard, Ali},
TITLE = {Design and Experimental Study on an Innovative UAV-LiDAR Topographic Mapping System for Precision Land Levelling},
JOURNAL = {Drones},
VOLUME = {6},
YEAR = {2022},
NUMBER = {12},
ARTICLE-NUMBER = {403},
URL = {https://www.mdpi.com/2504-446X/6/12/403},
ISSN = {2504-446X},
DOI = {10.3390/drones6120403}
}

@article{macdonell2023consumer,
author = {MacDonell, Craig John and Williams, Richard David and Maniatis, Georgios and Roberts, Kenny and Naylor, Mark},
title = {Consumer-grade UAV solid-state LiDAR accurately quantifies topography in a vegetated fluvial environment},
journal = {Earth Surface Processes and Landforms},
volume = {48},
number = {11},
pages = {2211-2229},
doi = {https://doi.org/10.1002/esp.5608},
url = {https://onlinelibrary.wiley.com/doi/abs/10.1002/esp.5608},
eprint = {https://onlinelibrary.wiley.com/doi/pdf/10.1002/esp.5608},
year = {2023}
}

@Article{trepekli2022uav,
author={Trepekli, Katerina
and Balstr{\o}m, Thomas
and Friborg, Thomas
and Fog, Bjarne
and Allotey, Albert N.
and Kofie, Richard Y.
and M{\o}ller-Jensen, Lasse},
title={UAV-borne, LiDAR-based elevation modelling: a method for improving local-scale urban flood risk assessment},
journal={Natural Hazards},
year={2022},
month={Aug},
day={01},
volume={113},
number={1},
pages={423-451},
issn={1573-0840},
doi={10.1007/s11069-022-05308-9},
url={https://doi.org/10.1007/s11069-022-05308-9}
}

@Article{lin2019evaluation,
AUTHOR = {Lin, Yi-Chun and Cheng, Yi-Ting and Zhou, Tian and Ravi, Radhika and Hasheminasab, Seyyed Meghdad and Flatt, John Evan and Troy, Cary and Habib, Ayman},
TITLE = {Evaluation of UAV LiDAR for Mapping Coastal Environments},
JOURNAL = {Remote Sensing},
VOLUME = {11},
YEAR = {2019},
NUMBER = {24},
ARTICLE-NUMBER = {2893},
URL = {https://www.mdpi.com/2072-4292/11/24/2893},
ISSN = {2072-4292},
DOI = {10.3390/rs11242893}
}

@Article{bartminski2023effectiveness,
AUTHOR = {Bartmiński, Piotr and Siłuch, Marcin and Kociuba, Waldemar},
TITLE = {The Effectiveness of a UAV-Based LiDAR Survey to Develop Digital Terrain Models and Topographic Texture Analyses},
JOURNAL = {Sensors},
VOLUME = {23},
YEAR = {2023},
NUMBER = {14},
ARTICLE-NUMBER = {6415},
URL = {https://www.mdpi.com/1424-8220/23/14/6415},
PubMedID = {37514709},
ISSN = {1424-8220},
DOI = {10.3390/s23146415}
}

@Article{oniga2024enhancing,
AUTHOR = {Oniga, Valeria-Ersilia and Loghin, Ana-Maria and Macovei, Mihaela and Lazar, Anca-Alina and Boroianu, Bogdan and Sestras, Paul},
TITLE = {Enhancing LiDAR-UAS Derived Digital Terrain Models with Hierarchic Robust and Volume-Based Filtering Approaches for Precision Topographic Mapping},
JOURNAL = {Remote Sensing},
VOLUME = {16},
YEAR = {2024},
NUMBER = {1},
ARTICLE-NUMBER = {78},
URL = {https://www.mdpi.com/2072-4292/16/1/78},
ISSN = {2072-4292},
DOI = {10.3390/rs16010078}
}

@article{choi2023acquisition,
title = {Acquisition of high-resolution topographic information in forest environments using integrated UAV-LiDAR system: System development and field demonstration},
journal = {Heliyon},
volume = {9},
number = {9},
pages = {e20225},
year = {2023},
issn = {2405-8440},
doi = {https://doi.org/10.1016/j.heliyon.2023.e20225},
url = {https://www.sciencedirect.com/science/article/pii/S2405844023074339},
author = {Shin-Kyu Choi and Ryan Angeles Ramirez and Tae-Hyuk Kwon}
}

@ARTICLE{steinke2024groundgrid,
  author={Steinke, Nicolai and Goehring, Daniel and Rojas, Raúl},
  journal={IEEE Robotics and Automation Letters}, 
  title={GroundGrid: LiDAR Point Cloud Ground Segmentation and Terrain Estimation}, 
  year={2024},
  volume={9},
  number={1},
  pages={420-426},
  doi={10.1109/LRA.2023.3333233}}

@article{delpino2023probabilistic,
author = {Xue, Hanzhang and Fu, Hao and Xiao, Liang and Fan, Yiming and Zhao, Dawei and Dai, Bin},
title = {Traversability analysis for autonomous driving in complex environment: A LiDAR-based terrain modeling approach},
journal = {Journal of Field Robotics},
volume = {40},
number = {7},
pages = {1779-1803},
doi = {https://doi.org/10.1002/rob.22209},
url = {https://onlinelibrary.wiley.com/doi/abs/10.1002/rob.22209},
eprint = {https://onlinelibrary.wiley.com/doi/pdf/10.1002/rob.22209},
year = {2023}
}

@ARTICLE{xue2023traversability,
  author={Pino, Iván del and Santamaria-Navarro, Angel and Garrell Zulueta, Anaís and Torres, Fernando and Andrade-Cetto, Juan},
  journal={IEEE Transactions on Intelligent Vehicles}, 
  title={Probabilistic Graph-Based Real-Time Ground Segmentation for Urban Robotics}, 
  year={2024},
  volume={9},
  number={5},
  pages={4989-5002},
  doi={10.1109/TIV.2024.3383599}}

@ARTICLE{harlow2024newwave,
  author={Harlow, Kyle and Jang, Hyesu and Barfoot, Timothy D. and Kim, Ayoung and Heckman, Christoffer},
  journal={IEEE Transactions on Robotics}, 
  title={A New Wave in Robotics: Survey on Recent MmWave Radar Applications in Robotics}, 
  year={2024},
  volume={40},
  number={},
  pages={4544-4560},
  doi={10.1109/TRO.2024.3463504}}

@ARTICLE{zhang2024radardiffusion,
  author={Zhang, Ruibin and Xue, Donglai and Wang, Yuhan and Geng, Ruixu and Gao, Fei},
  journal={IEEE Robotics and Automation Letters}, 
  title={Towards Dense and Accurate Radar Perception via Efficient Cross-Modal Diffusion Model}, 
  year={2024},
  volume={9},
  number={9},
  pages={7429-7436},
  doi={10.1109/LRA.2024.3426389}}

@ARTICLE{huang2021cmclr,
  author={Huang, Jui-Te and Lu, Chen-Lung and Chang, Po-Kai and Huang, Ching-I and Hsu, Chao-Chun and Ewe, Zu Lin and Huang, Po-Jui and Wang, Hsueh-Cheng},
  journal={IEEE Robotics and Automation Letters}, 
  title={Cross-Modal Contrastive Learning of Representations for Navigation Using Lightweight, Low-Cost Millimeter Wave Radar for Adverse Environmental Conditions}, 
  year={2021},
  volume={6},
  number={2},
  pages={3333-3340},
  doi={10.1109/LRA.2021.3062011}}

@ARTICLE{barra2023microdrone,
  author={Barra, Jeremy and Creuzet, Thierry and Lesecq, Suzanne and Scorletti, Gerard and Blanco, Eric and Zarudniev, Mykhailo},
  journal={IEEE Sensors Journal}, 
  title={Micro-Drone Ego-Velocity and Height Estimation in GPS-Denied Environments Using an FMCW MIMO Radar}, 
  year={2023},
  volume={23},
  number={3},
  pages={2684-2692},
  doi={10.1109/JSEN.2022.3229421}}

@Article{awan2024altimetry,
AUTHOR = {Awan, Maaz Ali and Dalveren, Yaser and Kara, Ali and Derawi, Mohammad},
TITLE = {Advancing mmWave Altimetry for Unmanned Aerial Systems: A Signal Processing Framework for Optimized Waveform Design},
JOURNAL = {Drones},
VOLUME = {8},
YEAR = {2024},
NUMBER = {9},
ARTICLE-NUMBER = {440},
URL = {https://www.mdpi.com/2504-446X/8/9/440},
ISSN = {2504-446X},
DOI = {10.3390/drones8090440}
}

@inproceedings{wang2025landing,
author = {Wang, Haoyang and Xu, Jingao and Luo, Xinyu and Chen, Xuecheng and Zhang, Ting and Duan, Ruiyang and Liu, Yunhao and Chen, Xinlei},
title = {Ultra-High-Frequency Harmony: mmWave Radar and Event Camera Orchestrate Accurate Drone Landing},
year = {2025},
isbn = {9798400714795},
publisher = {Association for Computing Machinery},
address = {New York, NY, USA},
url = {https://doi.org/10.1145/3715014.3722048},
doi = {10.1145/3715014.3722048},
booktitle = {Proceedings of the 23rd ACM Conference on Embedded Networked Sensor Systems},
pages = {15–29},
numpages = {15},
location = {UC Irvine Student Center., Irvine, CA, USA},
series = {SenSys '25}
}

@article{reina2011radar,
author = {Reina, Giulio and Underwood, James and Brooker, Graham and Durrant-Whyte, Hugh},
title = {Radar-based perception for autonomous outdoor vehicles},
journal = {Journal of Field Robotics},
volume = {28},
number = {6},
pages = {894-913},
doi = {https://doi.org/10.1002/rob.20393},
url = {https://onlinelibrary.wiley.com/doi/abs/10.1002/rob.20393},
eprint = {https://onlinelibrary.wiley.com/doi/pdf/10.1002/rob.20393},
year = {2011}
}

@INPROCEEDINGS{herraez2025groundaware,
  author={Herraez, Daniel Casado and Kaschner, Franz and Zeller, Matthias and Muhle, Dominik and Behley, Jens and Heidingsfeld, Michael and Cremers, Daniel and Stachniss, Cyrill},
  booktitle={2025 IEEE International Conference on Robotics and Automation (ICRA)}, 
  title={Ground-Aware Automotive Radar Odometry}, 
  year={2025},
  volume={},
  number={},
  pages={13007-13013},
  doi={10.1109/ICRA55743.2025.11128161}}

@INPROCEEDINGS{yang2025rio,
  author={Yang, Wooseong and Jang, Hyesu and Kim, Ayoung},
  booktitle={2025 IEEE International Conference on Robotics and Automation (ICRA)}, 
  title={Ground-Optimized 4D Radar-Inertial Odometry Via Continuous Velocity Integration Using Gaussian Process}, 
  year={2025},
  volume={},
  number={},
  pages={3815-3821},
  doi={10.1109/ICRA55743.2025.11127572}}

@article{fischler1981random,
  title={Random sample consensus: a paradigm for model fitting with applications to image analysis and automated cartography},
  author={Fischler, Martin A and Bolles, Robert C},
  journal={Communications of the ACM},
  volume={24},
  number={6},
  pages={381--395},
  year={1981},
  publisher={ACM New York, NY, USA}
}

@inproceedings{nurunnabi2014diagnostics,
  title={Diagnostics based principal component analysis for robust plane fitting in laser data},
  author={Nurunnabi, Abdul and Belton, David and West, Geoff},
  booktitle={16th Int'l Conf. Computer and Information Technology},
  pages={484--489},
  year={2014},
  organization={IEEE}
}

@article{fan2018road,
  title={Road surface 3D reconstruction based on dense subpixel disparity map estimation},
  author={Fan, Rui and Ai, Xiao and Dahnoun, Naim},
  journal={IEEE Transactions on Image Processing},
  volume={27},
  number={6},
  pages={3025--3035},
  year={2018},
  publisher={IEEE}
}

@article{fan2019pothole,
  title={Pothole detection based on disparity transformation and road surface modeling},
  author={Fan, Rui and Ozgunalp, Umar and Hosking, Brett and Liu, Ming and Pitas, Ioannis},
  journal={IEEE Transactions on Image Processing},
  volume={29},
  pages={897--908},
  year={2019},
  publisher={IEEE}
}

@article{nguyen2025semi,
  title={A Semi-Autonomous Robotic System for In Situ Soil Sampling, Analysis, and Mapping in Precision Agriculture},
  author={Nguyen, Thien Hoang and Muller, Erik and Rubin, Michael R and Wang, Xiaofei and Sibona, Fiorella and McBratney, Alex and Sukkarieh, Salah},
  journal={IEEE Transactions on Field Robotics},
  volume={3},
  pages={22--39},
  year={2025},
  publisher={IEEE}
}

@article{wembe2026predictive,
  title={A Predictive Control Strategy to Offset-Point Tracking for Agricultural Mobile Robots},
  author={Wembe, Stephane Ngnepiepaye and Rousseau, Vincent and Laconte, Johann and Lenain, Roland},
  journal={IEEE Transactions on Field Robotics},
  year={2026},
  publisher={IEEE}
}

@article{zang2019impact,
  title={The Impact of Adverse Weather Conditions on Autonomous Vehicles: How Rain, Snow, Fog, and Hail Affect the Performance of a Self-Driving Car},
  author={Zang, Shizhe and Ding, Ming and Smith, David and Tyler, Paul and Rakotoarivelo, Thierry and Kaafar, Mohamed Ali},
  journal={IEEE Vehicular Technology Magazine},
  volume={14},
  number={2},
  pages={103--111},
  year={2019},
  doi={10.1109/MVT.2019.2892497}
}

@inproceedings{bijelic2020seeing,
  title={Seeing Through Fog Without Seeing Fog: Deep Multimodal Sensor Fusion in Unseen Adverse Weather},
  author={Bijelic, Mario and Gruber, Tobias and Mannan, Fahim and Kraus, Florian and Ritter, Werner and Dietmayer, Klaus and Heide, Felix},
  booktitle={Proceedings of the IEEE/CVF Conference on Computer Vision and Pattern Recognition (CVPR)},
  pages={11682--11692},
  year={2020}
}

@inproceedings{qian2021robust,
  title={Robust Multimodal Vehicle Detection in Foggy Weather Using Complementary Lidar and Radar Signals},
  author={Qian, Kun and Zhu, Shilin and Zhang, Xinyu and Li, Li Erran},
  booktitle={Proceedings of the IEEE/CVF Conference on Computer Vision and Pattern Recognition (CVPR)},
  pages={444--453},
  year={2021}
}

@article{yang2025ralibev,
  title={RaLiBEV: Radar and LiDAR BEV Fusion Learning for Anchor Box Free Object Detection Systems},
  author={Yang, Yanlong and Liu, Jianan and Huang, Tao and Han, Qing-Long and Ma, Gang and Zhu, Bing},
  journal={IEEE Transactions on Circuits and Systems for Video Technology},
  volume={35},
  number={5},
  pages={4130--4143},
  year={2025},
  doi={10.1109/TCSVT.2024.3521375}
}

@inproceedings{zhang2014loam,
  title={LOAM: Lidar Odometry and Mapping in Real-time},
  author={Zhang, Ji and Singh, Sanjiv},
  booktitle={Robotics: Science and Systems},
  year={2014},
  doi={10.15607/RSS.2014.X.007}
}

@article{li2025ua,
  title={UA-MPC: Uncertainty-Aware Model Predictive Control for Motorized LiDAR Odometry},
  author={Li, Jianping and Xu, Xinhang and Liu, Jinxin and Cao, Kun and Yuan, Shenghai and Xie, Lihua},
  journal={IEEE Robotics and Automation Letters},
  volume={10},
  number={4},
  pages={3652--3659},
  year={2025}
}

@article{chen2026spinning,
  title={Accurate Calibration and Robust LiDAR-Inertial Odometry for Spinning Actuated LiDAR Systems},
  author={Chen, Zijie and Liu, Xiaowei and Xu, Yong and Yuan, Shenghai and Li, Jianping and Xie, Lihua},
  journal={IEEE Robotics and Automation Letters},
  volume={PP},
  number={99},
  pages={1--8},
  year={2026}
}

@inproceedings{barnes2020oxford,
  title={The Oxford Radar RobotCar Dataset: A Radar Extension to the Oxford RobotCar Dataset},
  author={Barnes, Dan and Gadd, Matthew and Murcutt, Paul and Newman, Paul and Posner, Ingmar},
  booktitle={2020 IEEE International Conference on Robotics and Automation (ICRA)},
  pages={6433--6438},
  year={2020},
  doi={10.1109/ICRA40945.2020.9196884}
}

@Article{fu2023rotating,
AUTHOR = {Meiqi Fu and Hua Zhang and Shuang Wang and Yuhang Shui},
TITLE = {Rotating 3D laser mapping system for multi-rotor drones},
JOURNAL = {Intelligence \& Robotics},
VOLUME = {3},
YEAR = {2023},
NUMBER = {4},
ARTICLE-NUMBER = {632-46},
URL = {https://www.oaepublish.com/articles/ir.2023.35},
ISSN = {2770-3541},
DOI = {10.20517/ir.2023.35}
}

@article{chen2023self,
author = {Nan Chen  and Fanze Kong  and Wei Xu  and Yixi Cai  and Haotian Li  and Dongjiao He  and Youming Qin  and Fu Zhang },
title = {A self-rotating, single-actuated UAV with extended sensor field of view for autonomous navigation},
journal = {Science Robotics},
volume = {8},
number = {76},
pages = {eade4538},
year = {2023},
doi = {10.1126/scirobotics.ade4538},
URL = {https://www.science.org/doi/abs/10.1126/scirobotics.ade4538},
eprint = {https://www.science.org/doi/pdf/10.1126/scirobotics.ade4538}}

\end{document}